\documentclass{article}

\PassOptionsToPackage{numbers, compress}{natbib}

\usepackage[preprint]{neurips_2026}

\usepackage[utf8]{inputenc} 
\usepackage[T1]{fontenc}    
\usepackage{hyperref}       
\usepackage{url}            
\usepackage{booktabs}       
\usepackage{longtable}       
\usepackage{amsfonts}       
\usepackage{nicefrac}       
\usepackage{microtype}      
\usepackage{xcolor}         
 \usepackage{multirow}      
\usepackage{xspace}         
\usepackage{svg}            
\usepackage{caption} 
\usepackage{pifont}         
\usepackage{makecell}         
\usepackage[table]{xcolor} 
\usepackage{graphicx}
\usepackage{titletoc}
\usepackage{enumitem}

\newcommand{\name}{SMMBench\xspace}
\newcommand{\cmark}{\textcolor{green!50!black}{\ding{51}}}
\newcommand{\xmark}{\textcolor{red!75!black}{\ding{55}}}

\title{SMMBench: A Benchmark for Source-Distributed Multimodal Agent Memory}

\author{%
  Huacan Chai$^{1}$, Yukai Wang$^{1}$, Yingxuan Yang$^{1}$, Dan Peng$^{2}$, Yuanyi Song$^{1}$, Zhihui Fu$^{2}$\\
  \textbf{Weiwen Liu}$^{1,*}$,\textbf{Jianghao Lin}$^{1,*}$,\textbf{Jun Wang}$^{2,*}$,\textbf{Weinan Zhang}$^{1,}$\thanks{Corresponding Authors}\\
  $^{1}$Shanghai Jiao Tong University, China;\\ 
  $^{2}$OPPO, China; \\
  \texttt{\{fatcat, wwliu, linjianghao, wnzhang\}@sjtu.edu.cn, wangjun7@oppo.com}
}

\begin{document}

\maketitle

\begin{abstract}
Existing benchmarks for multimodal memory reasoning largely evaluate systems within pre-assembled contexts, but under-evaluate whether agents can use evidence distributed across independently originated sources. We argue that \emph{source-distributed memory composition} is an important and under-examined bottleneck in multimodal agent memory, especially when relevant evidence is fragmented across heterogeneous artifacts such as conversations, profiles, screenshots, tables, images, and documents. To address this gap, we introduce \emph{Source-distributed Multimodal Memory Benchmark} (\name), which measures whether agents can retrieve, align, and compose multimodal evidence scattered across multiple sources rather than reason within a single curated context. \name evaluates four core capabilities: (1) cross-source multimodal reasoning; (2) conflict resolution; (3) preference reasoning; (4) memory-grounded action prediction. The benchmark contains $1,877$ samples grounded in $264$ sources. Experiments on representative memory-style and retrieval-based baselines show that current systems still struggle on these capabilities, positioning source-distributed multimodal memory as an important and still under-evaluated challenge for multimodal agents. 
Our data are available at \url{https://huggingface.co/datasets/HuacanChai/SMMBench}.
\end{abstract}

\section{Introduction}
Multimodal agents are increasingly expected to act as persistent assistants in productivity, desktop, and enterprise settings~\citep{zhou2026externalization,nie2026holos,guo2026skillprobesecurityauditingemerging,yang2025agentnetdecentralizedevolutionarycoordination}, where most real-world tasks are inherently \emph{cross-source}: the information needed to answer questions or execute actions is typically accumulated over time across chats, tables, documents, and other artifacts, rather than packaged in a single context~\citep{liu2026positionrealbarrierllm,xu2026multihaystack,abdallah2026mmbrightmultitaskmultimodalbenchmark}. This setting reveals a challenge in agent memory: the difficulty is often not merely reading a long input, but using evidence that is distributed across \emph{independent sources} created at different times and for different purposes, rather than reasoning over one pre-assembled context prepared for the final query~\citep{zhu2025evolutionaryperspectivesevaluationllmbased}.

\begin{figure*}[t]
\centering
  \centering
  \includegraphics[width=\linewidth]{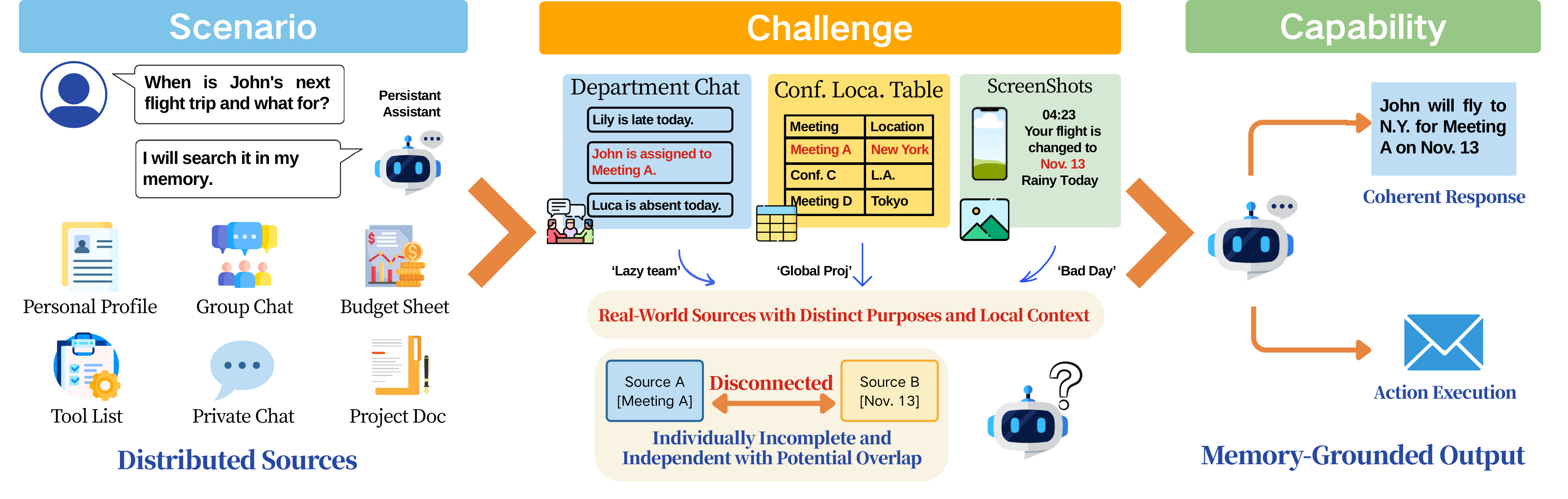}
  \vspace{-0.5cm}
  \caption{In real-world tasks, the necessary evidence is often distributed across multiple sources with distinct purposes and local contexts, while remaining potentially overlapping entities. Because no single source is sufficient, agents must retrieve and compose fragmented evidence across sources, making source-distributed memory a key bottleneck for memory-grounded responses and actions.}
  \label{fig:intro1}
  \vspace{-0.5cm}
\end{figure*}

\textbf{We argue that \textit{source-distributed memory composition} is an under-evaluated bottleneck in multimodal agent memory.} \emph{Source-distributed} means that the evidence is fragmented across multiple independently originated sources, such as separate group or private chats, profiles, tables, and documents, each with its own main purpose and local context. This brings challenges that are qualitatively different from reasoning over a single curated context. First, relevant evidence is distributed across multiple sources, and \textit{no single source is sufficient to determine the final answer on its own}. For example, as illustrated in Figure~\ref{fig:intro1}, an agent may need to connect evidence from a department chat, a meeting-location table, and a phone screenshot to infer that `John will fly to New York for Meeting A on Nov.~13'; no single source states this answer directly. 
Second, necessary evidence is often distributed across independently originated sources with different purposes and local contexts. Because these sources are created independently for different purposes rather than jointly organized for the query, \textit{their local contexts compartmentalize partial clues and make them harder to connect}. This creates a distinct memory bottleneck: the agent must identify sources and bridge across their contextual boundaries to compose the answer.
Third, information from different sources may conflict with one another, \textit{requiring the agent to update evidence and resolve conflicts} by reasoning over their different authority levels or time states. Generally, \textbf{the key challenge is not merely remembering isolated facts, but composing distributed evidence into answers or actions}.


Prior benchmarks have made important progress on multimodal long-context and memory settings, but most of them still evaluate reasoning within a single pre-assembled context. Multimodal long-context benchmarks such as MILEBench~\citep{song2024milebench} and Mementos~\citep{wang2024mementos} evaluate whether MLLMs can retrieve, compare, and reason over long text-image contexts or visual streams. Mem-Gallery~\citep{bei2026memgallery} further moves toward conversational memory over coherent multimodal interaction traces. However, these benchmarks primarily evaluate evidence use within a coherent context or unified retrieval corpus. They therefore leave under-evaluated whether agent memory systems can compose multimodal evidence distributed across independently originated sources.



To evaluate this gap, we introduce \emph{\textbf{S}ource-distributed \textbf{M}ultimodal \textbf{M}emory \textbf{Bench}} (\textbf{\name}), a benchmark for multimodal agent memory in which relevant evidence is intentionally distributed across multiple independently originated sources rather than provided as one pre-assembled context. The benchmark covers representative artifact types that arise in real-world persistent assistant scenarios, including conversations, profiles, tables, images, and documents, and organizes evaluation around four core capabilities: cross-source reasoning, conflict resolution, preference reasoning, and memory-grounded action prediction. It provides fine-grained evidence annotations together with both open-book and retrieval-based evaluation settings, enabling analysis of not only end-task accuracy but also how systems use distributed memory under different access conditions. Overall, the benchmark contains $1877$ evaluation samples across $5$ task types and $264$ sources. Experimental results show that even the strongest evaluated systems still perform poorly in this setting, highlighting source-distributed memory as an important and still under-evaluated challenge for multimodal agents. Our contributions are as follows:

\begin{itemize}[leftmargin=1.2em]
  \item \textbf{Problem Identification.} We identify \emph{source-distributed memory composition} as an under-evaluated bottleneck in multimodal agent memory, where the key challenge is composing evidence across independently originated sources rather than reasoning within one prepared context.
  \item \textbf{Challenge Characterization.} We clarify how source-distributed memory differs from standard long-context reasoning by characterizing the distinct challenges introduced by independent sources, including source-level incompleteness, cross-source context bridging, and conflict resolution under different authority levels or time states.
  \item \textbf{Benchmark Construction.} We introduce \emph{\textbf{S}ource-distributed \textbf{M}ultimodal \textbf{M}emory \textbf{Bench}}, a benchmark that operationalizes this challenge through source objects such as conversations, profiles, screenshots, tables, images, and documents, and evaluates cross-source reasoning, conflict resolution, preference reasoning, and memory-grounded action prediction.
  \item \textbf{Empirical Findings.} We provide fine-grained evidence annotations and evaluation experiments under open-book and retrieval-based settings. Experiments on $1877$ samples, $5$ task types, and $264$ sources show that current representative methods remain far from effective.
\end{itemize}

\begin{table*}[t]
\centering
\setlength{\tabcolsep}{2.2pt}
\renewcommand{\arraystretch}{0.85}
\caption{Comparison with representative memory benchmarks. 
\cmark: Satisfies; \xmark: Does not satisfy.}
\vspace{-0.2cm}
\label{tab:benchmark_comparison}
\definecolor{smmbenchbg}{HTML}{FEEDDE}
\begin{tabular}{lccccc|cccc}
\toprule
\textbf{Benchmark} 
& \textbf{MM} 
& \textbf{Ev.M.} 
& \textbf{Src. M.} 
& \textbf{M.Ev.} 
& \textbf{Indep. Src.} 
& \textbf{X.S.} 
& \textbf{C.R.} 
& \textbf{P.R.} 
& \textbf{A.P.} \\
\midrule
LongMemEval~\citep{wu2024longmemeval}         & \xmark & T & C & \cmark & \xmark & \xmark & \xmark & \cmark & \xmark \\
MemoryAgentBench~\citep{hu2025evaluating}     & \xmark & T & C/D & \cmark & \xmark & \xmark & \cmark & \cmark & \xmark \\
LoCCO~\citep{jia-etal-2025-evaluating}        & \xmark & T & C & \xmark & \xmark & \xmark & \xmark & \xmark & \xmark \\
LoCoMo~\citep{maharana2024evaluating}         & \cmark & T/I & C & \cmark & \xmark & \cmark & \xmark & \cmark & \xmark \\
Mementos~\citep{wang2024mementos}             & \cmark & I & I & \cmark & \xmark & \cmark & \xmark & \xmark & \xmark \\
MMDU~\citep{liu2024mmdu}                      & \cmark & T/I & C & \xmark & \xmark & \cmark & \xmark & \xmark & \xmark \\
MMRC~\citep{xue2025mmrc}                      & \cmark & T/I & C & \xmark & \xmark & \cmark & \cmark & \xmark & \xmark \\
MultiHaystack~\citep{xu2026multihaystack}     & \cmark & D/I/V & D/I & \xmark & \cmark & \xmark & \xmark & \xmark & \xmark \\
Mem-Gallery~\citep{bei2026memgallery}         & \cmark & T/I & C & \cmark & \xmark & \cmark & \cmark & \xmark & \xmark \\
\midrule
\rowcolor{smmbenchbg}
\textbf{\name}                                & \cmark & T/I/D/Tab. & C/D/I & \cmark & \cmark & \cmark & \cmark & \cmark & \cmark \\
\bottomrule
\end{tabular}%
\vspace{-0.2em}
\begin{flushleft}
\fontsize{8pt}{6.4pt}\selectfont
\textbf{MM}: contains multimodal inputs;
\textbf{Ev.M.} \& \textbf{Src.M.}: modality types of evidence and sources, T for text, I for image, D for document, Tab. for table, C for conversation;
\textbf{M.E.}: problems cannot be solved without combining multiple pieces of evidence;
\textbf{Indep. Src.}: evidence distributed across heterogeneous and independent sources;
The following columns show the evaluated capabilities.
\textbf{X.S.}: cross-source reasoning;
\textbf{C.R.}: conflict resolution;
\textbf{P.R.}: preference reasoning;
\textbf{A.P.}: action prediction.
\end{flushleft}
\vspace{-0.3cm}
\end{table*}

\section{Related Work}
\vspace{-0.2cm}
\subsection{Multimodal Agent Memory Benchmark}
Recent benchmarks on multimodal agent memory have largely centered on the performance bottlenecks induced by long input contexts. MILEBench~\citep{song2024milebench} evaluates the long-context understanding ability of MLLMs, while Mementos~\citep{wang2024mementos} focuses on reasoning over long image sequences. Mem-Gallery~\citep{bei2026memgallery} moves closer to the agent memory setting by emphasizing memory maintenance in multi-session conversations, yet its conversational trajectories are still largely coherent rather than distributed across independently originated sources.
However, these benchmarks are still insufficient for evaluating the source-distributed setting targeted by \name. They mainly assume a \emph{coherent context}, such as a long conversation, image stream, or unified interaction history. By contrast, \name evaluates whether a system can identify and compose answer-critical evidence scattered across \emph{independently originated sources} with different purposes and local contexts. This source-level fragmentation is not reducible to ordinary long-context reasoning.

\vspace{-0.2cm}
\subsection{Multimodal Agent RAG Benchmark}
Multimodal RAG benchmarks have similarly developed. M$^2$RAG~\citep{Liu2025BenchmarkingRG} evaluates how effectively MLLMs retrieve and use multimodal documents for open-domain tasks such as captioning. MultiHaystack~\citep{xu2026multihaystack} emphasizes multimodal evidence under noisy retrieval settings. 
Nevertheless, most multimodal RAG benchmarks assume a \emph{coherent corpus retrieval} setting, where evidence is retrieved from a shared repository and the main challenge is locating relevant items under scale or noise. They therefore test retrieval of relevant multimodal evidence, but not composition across independently originated sources with separate contextual boundaries. By contrast, \name evaluates whether systems can identify relevant sources, recover partial clues from each, and compose them into a coherent answer or executable action.


\section{\name Benchmark}
\vspace{-0.2cm}
\begin{figure*}[t]
\centering
  \centering
  \includegraphics[width=\linewidth]{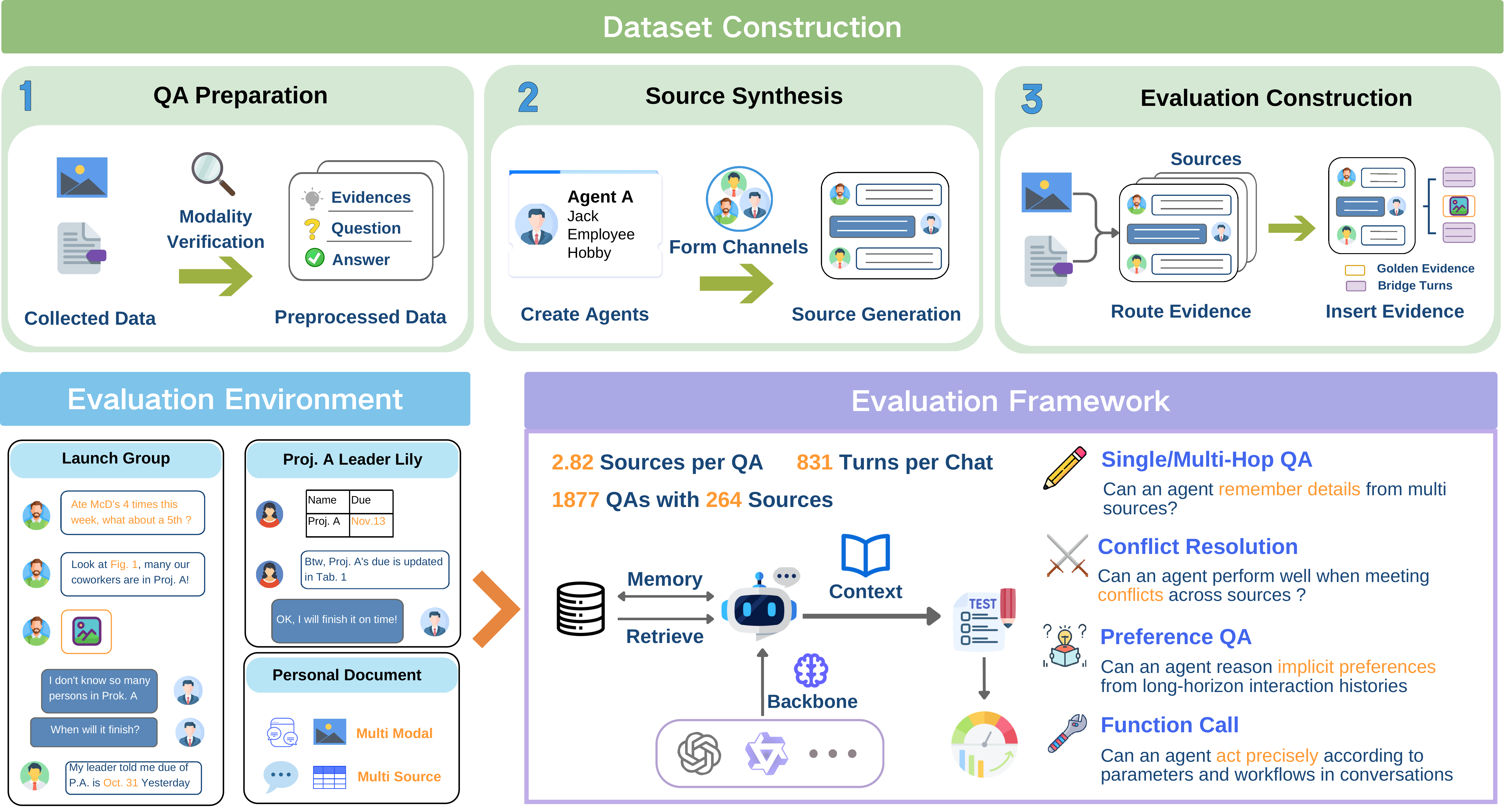}
  \caption{Overview of \name. \textbf{Top}: Dataset construction pipeline. \textbf{Bottom left}: Agents interact with heterogeneous memory sources, where answer-critical evidence is distributed across independent sources. \textbf{Bottom right}: Given the constructed environments, agents retrieve from memory and are evaluated on multiple task types, including single-/multi-hop QA, conflict resolution, preference reasoning, and function calling.}
  \label{fig:pipeline}
  \vspace{-0.2cm}
\end{figure*}

\subsection{Problem Formulation}
\label{section:problem_formulation}
We formulate \name as a memory-grounded question answering and action prediction benchmark over \emph{source-distributed} multimodal evidence. In \name, a \emph{source} is an independently originated memory object with its own local context and organizational boundary, such as a group or private chat, a profile page, an image, a table, or a document, rather than an artificial retrieval chunk obtained by splitting a larger object. Formally, each evaluation sample contains a set of sources
\begin{equation}
\mathcal{S} = \{S_1, S_2, \dots, S_m\},
\end{equation}
where each source $S_i$ consists of one or more evidence-bearing items
\begin{equation}
S_i = \{o_{i,1}, o_{i,2}, \dots, o_{i,n_i}\}, \quad o_{i,j} = \langle x_{i,j}, s_i, \tau_{i,j} \rangle,
\end{equation}
with content $x_{i,j}$, shared source identity $s_i$, and timestamp or local temporal position $\tau_{i,j}$. The content $x_{i,j}$ may include text, images, tables, document pages, or other multimodal evidence.

A sample is considered \emph{source-distributed} only if its answer-critical evidence satisfies two conditions: (1) the required evidence comes from at least two distinct sources, and (2) no single source alone is sufficient to determine the gold answer. Let $\mathcal{E}^*(q)$ be the minimal evidence set required for question $q$, and let $s(e)$ denote the source of evidence item $e$. We require
\begin{equation}
|\{s(e) \mid e \in \mathcal{E}^*(q)\}| \geq 2.
\end{equation}
Thus, each source provides only partial information, and the final answer must be obtained by composing evidence across source boundaries rather than by reading one locally complete source.

Given the source set $\mathcal{S}$, the agent incrementally observes items from these sources and maintains an external memory state $M_t$. For conversational sources, the observations follow their turn order; for non-conversational sources such as documents or images, the observations correspond to their associated source items. We denote the overall observation stream as
\begin{equation}
\mathcal{O} = \{o_1, o_2, \ldots, o_T\},
\end{equation}
where each observation retains its source identity. As each observation arrives, the memory is updated by the memory update operator $\Phi$:
\begin{equation}
M_{t+1} = \Phi(M_t, o_t).
\end{equation}
After ingesting the full sources, the agent receives a question $q$, retrieves relevant memory units, and generates the final answer:
\begin{equation}
M_{\mathrm{ret}} = R(M_T, q), \quad y = G(q, M_{\mathrm{ret}}).
\end{equation}
Under this formulation, success requires more than recalling isolated facts from a long input. A successful system must (1) preserve source-aware memory over heterogeneous source objects, (2) retrieve evidence spanning the right source boundaries, and (3) compose or reconcile these pieces into a coherent final answer or action. Therefore, \name evaluates \emph{source-distributed memory composition} rather than only long-context multimodal recall.

\subsection{Benchmark Construction}
\label{section:benchmark_construction}

We construct \name through a three-stage pipeline: QA preparation, conversational source synthesis, and source-aware evidence insertion. This pipeline turns curated multimodal QA instances into memory-grounded evaluation samples whose answer-critical evidence is distributed across multiple sources.
For the detailed building process, please refer to Appendix~\ref{appendix:dataset_construction_details}.

\vspace{-0.2cm}
\paragraph{QA Preparation}
Collected from diverse public multimodal benchmarks, we convert raw samples into a question-answer pair along with a unified set of evidence units, which will serve as the inputs to the following stages. Detailed preparation and verification procedures are deferred to Appendix~\ref{appendix:open_source_dataset_preprocess}.

\vspace{-0.2cm}
\paragraph{Conversational Source Synthesis}
Next, we construct a multi-source conversational environment to host the prepared evidence later. Concretely, we instantiate agents with predefined profiles and organize them into several conversational sources, including both group and private chats, with some participants recurring across sources. Each conversational source has its own participants and continuous communicative topics, obtaining multiple parallel yet related interaction streams rather than a single monolithic conversation. We then simulate ordinary multi-turn conversations within each source using high-level topical cues abstracted from the sampled evidence, so that the resulting conversations remain relevant to the later grounding stage without exposing the evidence itself.

\vspace{-0.2cm}
\paragraph{Source-Aware Evidence Insertion}
Finally, we ground the prepared evidence into the generated multi-source environment. A key design choice is to place complementary evidence units into different sources, rather than merely spreading them across distant positions within the same source, emphasizing the challenge of identifying relevant sources and composing information across sources. 
Specifically, an LLM-based source-aware inserter is used to route each evidence unit to a suitable source and local position according to the source context and the evidence content, while preserving the readability and continuity of the host source after insertion. For evidence that is complementary, updated, or conflicting, we further check cross-source placement consistency and preserve their temporal dependencies, so that newer information appears in a later and coherent position relative to the earlier evidence it supplements or revises. Therefore, samples that require agents to absorb and compose evidence from distributed sources can be obtained.

\begin{figure*}[t]
  \centering
  \begin{minipage}[t]{0.3\textwidth}
    \vspace{0pt}
    \centering
    \captionsetup{type=figure,font=small,margin=4pt}
    \captionbox{Illustration of categories in \name.\label{fig:category_illustration}}
      {\includegraphics[width=01\linewidth]{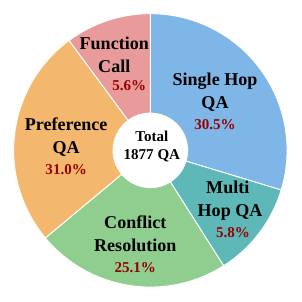}}
  \end{minipage}
  \hfill
  \begin{minipage}[t]{0.64\textwidth}
    \vspace{0pt}
    \centering
    \captionsetup{type=table,font=small,margin=4pt}
    \captionbox{Statistics of \name. `\textbf{Avg.}' means `Average', `\textbf{Avg. Evidence Int.}' means `Average evidence interval turns in conversational sources'. `\textbf{Avg. Sources}' means the average number of input sources to solve one QA pair.\label{tab:benchmark_statistics}}
      {%
      \small
      \setlength{\tabcolsep}{6pt}%
      \renewcommand{\arraystretch}{1.15}%
      \begin{tabular}{l c@{\hspace{1.8em}}l c}
      \toprule
      \multicolumn{2}{c}{\textbf{Source Statistics}} &
      \multicolumn{2}{c}{\textbf{QA Evidence Statistics}} \\
      \cmidrule(r){1-2} \cmidrule(l){3-4}
      \textbf{Metric per Source} & \textbf{Value} & \textbf{Metric per QA} & \textbf{Value} \\
      \midrule
      \# \textbf{Sources}            & 264    & \textbf{Avg. Evidence}         & 4.54 \\
      \textbf{Avg. Turns}                    & 831    & \textbf{Avg. Sources}  & 2.82 \\
      \textbf{Avg. Images}                   & 14.99  & \textbf{Avg. Texts}    & 2.31 \\
      \textbf{Avg. Docs }                    & 2.82   & \textbf{Avg. Images}   & 0.95 \\
      \textbf{Avg. Evidence Int.}                & 45.43  & \textbf{Avg. Docs}     & 1.24 \\
      \bottomrule
      \end{tabular}%
      }
  \end{minipage}
  \vspace{-0.2cm}
\end{figure*}

\subsection{Benchmark Analysis}

\paragraph{Task Introduction}
To better reflect real-world agent applications and the four core capabilities discussed, we organize the benchmark into $5$ task types: (1) \& (2) \textsc{Single-Hop QA} (\textbf{S.H.}) \&  \textsc{Multi-Hop QA} (\textbf{M.H.}), where the agent must retrieve multiple pieces of multimodal evidence from different sources and perform a \textit{single-} or \textit{multi-}step reasoning process to answer the question; (3) \textsc{Conflict Resolution} (\textbf{C.R.}), where the environment contains outdated evidence and the agent must identify the newer evidence from another source to override it; (4) \textsc{Preference Reasoning} (\textbf{P.R.}), where the agent must infer user preferences by integrating implicit personalized cues across multiple sources; and (5) \textsc{Function Call} (\textbf{F.C.}), which is designed to closely resemble realistic agent use cases and requires the agent to learn workflows from different sources, remember detailed parameters in multimodal content, and produce the exact tool invocation, serving as precise action-prediction tasks.


\vspace{-0.2cm}
\paragraph{Benchmark Statistics and Framework}
Figure~\ref{fig:category_illustration} and Table~\ref{tab:benchmark_statistics} summarize the overall composition of \name. \name has three notable properties. 
(1) \textbf{Source-distributed and complementary evidence.} The benchmark is explicitly multi-source: each QA instance requires evidence from 2.82 sources on average, with 4.54 supporting evidence items per sample, showing that answers typically depend on evidence composition across multiple sources rather than local retrieval within a single context. Additional experiments in the Appendix~\ref{appendix:modality_experiment} confirm that these evidence pieces are complementary rather than redundant. 
(2) \textbf{Rich multimodal evidence.} \name contains complementary textual and non-textual information, with each QA instance involving 2.31 text evidence items, 0.95 image evidence items, and 1.24 document evidence items on average, indicating that many cases require joint use of different modalities. 
(3) \textbf{Sparse long-horizon evidence placement.} The conversational sources are long and the relevant clues are intentionally sparse: each source contains 831 turns on average, while relevant evidence appears only every 45.43 turns on average. This design reduces shortcut solving from locally clustered clues and makes memory retrieval depend on sustained tracking over long interaction histories.

Each sample is evaluated as a memory-updated process following Section~\ref{section:benchmark_construction}. During memory construction, conversational turns from different sources are merged by timestamp and fed into the target agent memory sequentially; non-conversational sources are also inserted according to their designated temporal positions in the same global sequence. Once memory construction is complete, the benchmark question is used as a query to trigger memory retrieval, and the recalled items are concatenated with the question as the final context for the backbone LLM. The exact memory update and retrieval mechanisms are left to each agent memory system. For \textbf{F.C.} tasks, candidate tools are provided and the agent is required to generate the exact function invocation; for all other tasks, responses are evaluated in a multiple-choice format.
\vspace{-5pt}
\section{Experiment}
We conduct extensive experiments on \name to answer the following research questions.
\begin{enumerate}[itemsep=0.15em, topsep=0.25em]
  \item[\textbf{RQ1}]: \textbf{How do representative memory/retrieval baselines perform on our benchmark across different task categories? }
  \item[\textbf{RQ2}]: \textbf{Does distributed supporting evidence across multiple sources make memory reasoning more difficult? }
  \item[\textbf{RQ3}]: \textbf{How does the number of required sources affect benchmark difficulty?}
  \item[\textbf{RQ4}]: \textbf{How does the number of retrieved items affect the overall performance?}
  \item[\textbf{RQ5}]: \textbf{What are the main failure modes of current systems on source-distributed multimodal memory tasks?}
\end{enumerate}
\vspace{-5pt}
\subsection{Baselines and Evaluation Setup}
We evaluate representative baselines from two families: \emph{memory-style} methods and \emph{RAG-style} methods. The memory-style baselines include \textsc{Short-Term Mem.} ~\citep{zhang2025memengine}, \textsc{Reflexion Mem.} ~\citep{shinn2023reflexion}, \textsc{Gen.Agen Mem.} ~\citep{park2023generative}, \textsc{Self Controlled Mem.} ~\citep{wang2023enhancing}, \textsc{MIRIX} ~\citep{wang2025mirix}, \textsc{MemGPT} ~\citep{packer2023memgpt}, \textsc{MemVerse} ~\citep{liu2025memverse}, \textsc{Mem0} ~\citep{chhikara2025mem0}, and \textsc{OmniSimpleMem} ~\citep{omnisimplemem2026}. The RAG-style baselines include \textsc{Native RAG} ~\citep{zhang2025memengine}, \textsc{HMRAG} ~\citep{liu2025hm}, \textsc{UniversalRAG} ~\citep{yeo2025universalrag}, and \textsc{VRAG} ~\citep{wang2025vrag}. Among these methods, \textsc{MemGPT}, \textsc{MIRIX}, \textsc{UniversalRAG}, and \textsc{VRAG} provide native multimodal memory/retrieval, while other baselines are evaluated in a \emph{Text+Caption} setting, where non-textual evidence is converted into textual captions before storage and retrieval. In addition, \textsc{MemGPT} and \textsc{MIRIX} are also evaluated in a textual-memory setting for direct comparison across modality-access settings. We also include two reference baselines. The \textsc{Random Baseline} represents chance-level performance on the multiple-choice tasks and serves as a lower anchor for basic answerability, while the \textsc{Golden Evidence Baseline} directly provides supporting evidence to the backbone model, serving as a high-performance reference. 

For evaluation, we use gpt-4.1~\citep{openai2025gpt41} to generate captions for non-text evidence in the Text+Caption setting. For metrics, we report multiple-choice accuracy for \textsc{Single-Hop QA}, \textsc{Multi-Hop QA}, \textsc{Conflict Resolution}, and \textsc{Preference Reasoning}, and exact match for \textsc{Function Call}, together with the overall average. For all baselines, we use qwen3-vl-235b-instruct~\citep{Yang2025Qwen3TR} as the backbone model. For baselines that include retrieval/recall pipelines, the default number of retrieved items is set to 20. Detailed evaluation prompt templates can be seen in the Appendix~\ref{appendix:prompts}. Additional details and configurations are listed in the Appendix~\ref{appendix:additional_experiment}.

\begin{table*}[t]
  \centering
  \definecolor{rankone}{HTML}{bbd0e8}
  \definecolor{ranktwo}{HTML}{ffe6cc}
  \definecolor{rankthree}{HTML}{d5e8d4}
  \small
  \setlength{\tabcolsep}{5pt}
  \renewcommand{\arraystretch}{1.2}
  \caption{Main results on \name. Within each metric column, the top three values among benchmarked baselines with complete entries, excluding these two reference rows, are highlighted: \raisebox{-0.12ex}{\textcolor{rankone}{\rule{.65em}{.65em}}} (highest), \raisebox{-0.12ex}{\textcolor{ranktwo}{\rule{.65em}{.65em}}} (second), and \raisebox{-0.12ex}{\textcolor{rankthree}{\rule{.65em}{.65em}}} (third). \textbf{S.H.}, \textbf{M.H.}, \textbf{C.R.}, \textbf{P.R.}, and \textbf{F.C.} denote \textsc{Single-Hop QA}, \textsc{Multi-Hop QA}, \textsc{Conflict Resolution}, \textsc{Preference Reasoning}, and \textsc{Function Call}. All scores are averaged over 3 runs and the \textbf{Overall} reports the unweighted average score.}
  \label{tab:main_results}
  \begin{tabular}{lllcccccc}
  \hline
  \specialrule{1.1pt}{0pt}{0pt}
  \textbf{Modal} & \textbf{Method} & \textbf{Baseline} & \textbf{S.H.} & \textbf{M.H.} & \textbf{C.R.} & \textbf{P.R.} & \textbf{F.C.} & \textbf{Overall} \\
  \hline
  \specialrule{1.1pt}{0pt}{0pt}
  \multicolumn{3}{c}{Random Baseline} & 0.2500 & 0.2500 & 0.2500 & 0.2500 & 0.0000 & 0.2000 \\
  \specialrule{1.1pt}{0pt}{0pt}
  \multicolumn{3}{c}{Golden Evidence Baseline} & 0.8753 & 0.7768 & 0.8365 & 0.9699 & 0.2778 & 0.7473 \\
  \specialrule{1.1pt}{0pt}{0pt}
  \multirow{10}{*}{\makecell[c]{Text +\\ Caption}}
  & \multirow{8}{*}{Memory}
  & Short-Term Mem.        & 0.2990 & 0.2292 & 0.2246 & 0.2702 & 0.0093 & 0.2064 \\
  & & Reflexion Mem.       & 0.5385 & 0.4861 & 0.3962 & 0.3064 & 0.0278 & 0.3510 \\
  & & Gen. Agent Mem.      & 0.3164 & 0.2708 & 0.2754 & 0.2238 & 0.0185 & 0.2210 \\
  & & Self Controlled Mem. & 0.3129 & 0.3403 & 0.2881 & 0.2203 & 0.0185 & 0.2360 \\
  & & MIRIX                & 0.3601 & 0.2569 & 0.3072 & 0.2960 & 0.0278 & 0.2496 \\
  & & MemGPT               & 0.5734 & 0.4722 & 0.3326 & \cellcolor{rankone}0.4389 & 0.0370 & 0.3708 \\
  & & MemVerse             & 0.3549 & 0.3264 & 0.2691 & 0.2754 & 0.0093 & 0.2470 \\
  & & Mem0                 & \cellcolor{rankthree}0.7430 & \cellcolor{rankthree}0.5903 & 0.3665 & 0.2909 & 0.0398 & \cellcolor{rankthree}0.4061 \\
  \cline{2-9}
  & \multirow{2}{*}{RAG}
  & Native RAG            & \cellcolor{ranktwo}0.7797 & \cellcolor{ranktwo}0.6667 & \cellcolor{ranktwo}0.4068 & \cellcolor{ranktwo}0.3683 & \cellcolor{rankthree}0.0741 & \cellcolor{ranktwo}0.4591 \\
  & & HMRAG               & \cellcolor{rankone}0.8129 & \cellcolor{rankone}0.7153 & \cellcolor{rankone}0.5191 & \cellcolor{rankthree}0.3081 & \cellcolor{rankone}0.1111 & \cellcolor{rankone}0.4933 \\
  \hline
  \specialrule{1.1pt}{0pt}{0pt}
  \multirow{5}{*}{MM}
  & \multirow{3}{*}{Memory}
  & MemGPT               & 0.3497 & 0.5139 & 0.2691 & 0.1997 & 0.0241 & 0.2713 \\
  & & MIRIX              & 0.5507 & 0.3472 & \cellcolor{rankthree}0.4025 & 0.2031 & \cellcolor{ranktwo}0.0769 & 0.3161 \\
  & & OmniSimpleMem     & 0.2311 & 0.0375 & 0.2153 & 0.1840 & 0.0185 & 0.1373 \\
  \cline{2-9}
  & \multirow{2}{*}{RAG}
  & UniversalRAG         & 0.3322 & 0.3403 & 0.2606 & 0.1566 & 0.0370 & 0.2253 \\
  & & VRAG               & 0.4913 & 0.4514 & 0.3919 & 0.1842 & 0.0463 & 0.3130 \\
  \hline
  \specialrule{1.1pt}{0pt}{0pt}
  \end{tabular}
  \renewcommand{\arraystretch}{1.0}
  \vspace{-0.2cm}
\end{table*}

\subsection{RQ1: Main Results}
\vspace{-0.2cm}
The following insights can be drawn from Table~\ref{tab:main_results}. Additional analysis is provided in Appendix~\ref{appendix:main_results}.

\vspace{-0.2cm}
\paragraph{Existing methods still struggle on source-distributed, multimodal memory tasks.}
The reference baselines show that the benchmark is solvable but still far from performing strongly on it. Strong baselines clearly outperform the \textsc{Random Baseline}, yet there remains a clear gap to the \textsc{Golden Evidence Baseline}. Even baselines with relatively mature memory systems like \textsc{MIRIX} or \textsc{MemGPT} still struggle to retrieve, align, and compose the needed evidence under realistic multi-source access, indicating that current methods remain insufficient for source-distributed multimodal memory settings. Performance is also uneven across task families: systems that are relatively strong on some QA-style tasks often remain clearly weaker on preference reasoning or function calling, and no single method is consistently strong across the board.
\vspace{-0.2cm}
\paragraph{Native multimodal access alone does not resolve the challenge of source-distributed memory.}
Notably, even without observing native multimodal content, some memory systems such as \textsc{Reflexion Mem.} still perform reasonably well. A clean comparison comes from paired results within the same memory systems. \textsc{MIRIX} improves from 0.2496 to 0.3161 overall when moving from Text+Caption to native MM setting, whereas \textsc{MemGPT} drops from 0.3708 to 0.2713. This suggests that the challenge in the source-distributed setting is not solely determined by access to native multimodal content, but also by the ability to compose distributed evidence across sources.
\vspace{-0.2cm}
\paragraph{Source-distributed difficulty is amplified on precision-sensitive tasks.}
\textsc{Function Call} remains the most challenging setting in our benchmark. The best \textbf{F.C.} score is only 0.1111, and even the \textsc{Golden Evidence Baseline} reaches only 0.2778, far below its scores on the other tasks. This pattern suggests that the difficulty of source-distributed memory is especially amplified when the task requires precise argument recovery from dispersed multimodal evidence. In other words, source-distributed settings do not only make evidence harder to find, but also make it harder to accurately align and assemble fine-grained information into executable actions. The same gap is also visible within individual systems: \textsc{MemGPT} and \textsc{MIRIX} both achieve much stronger QA performance than \textbf{F.C.}. Therefore, as a stress-test subset, \textsc{Function Call} can evaluate whether remembered information can be converted into downstream actions under source-distributed conditions.

\subsection{RQ2: Impact of Source-Concentrated V.S. Source-Distributed Evidence}
\label{section:rq2_source_distribution_experiment}
\vspace{-0.2cm}

\begin{figure*}[t]
\centering
\begin{minipage}[b]{0.48\textwidth}
    \vspace{0pt}
  \centering
  \includegraphics[width=\linewidth]{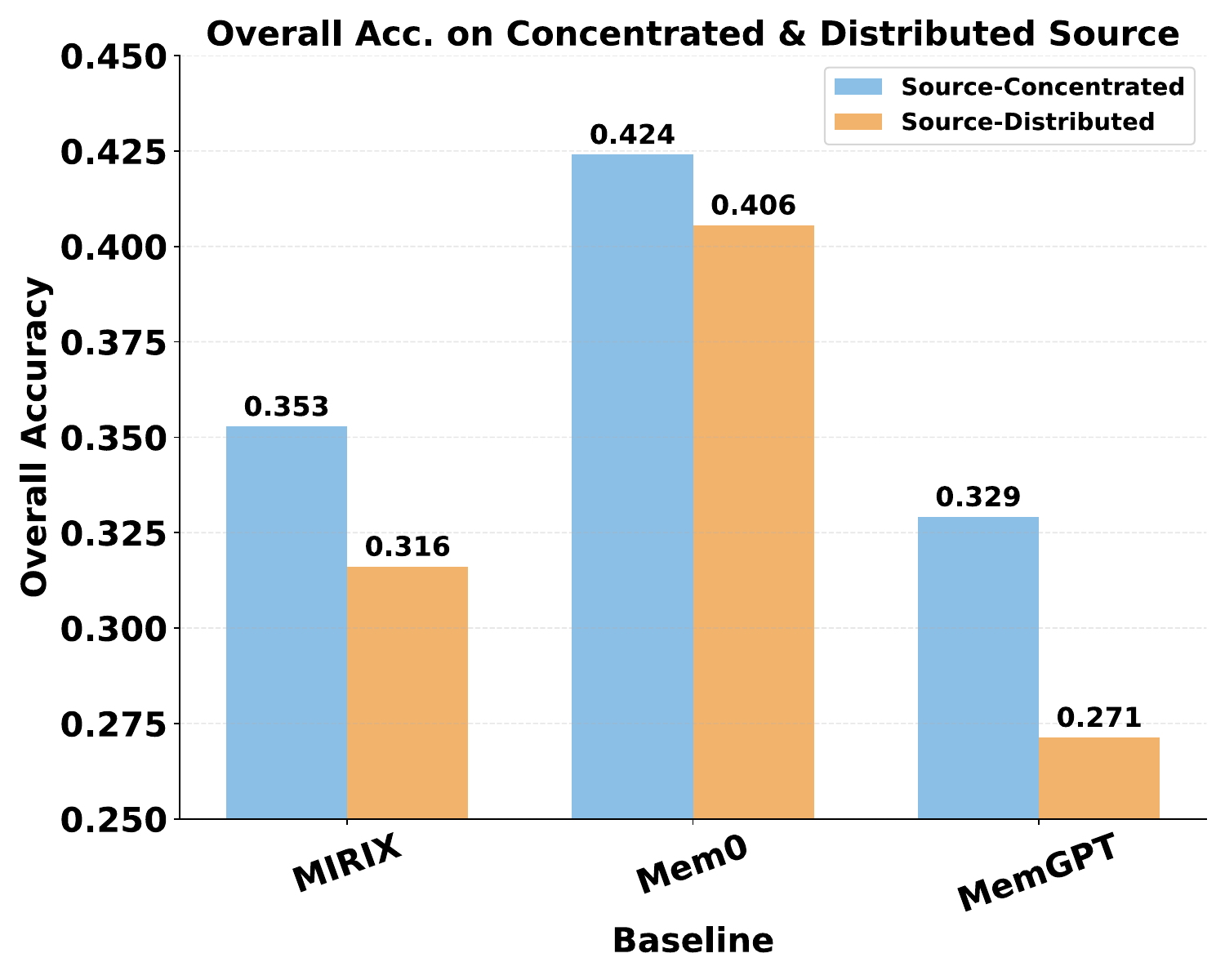}
  \vspace{-0.5cm}
  \caption{Comparison between Source-Concentrated and Source-Distributed Settings}
  \label{fig:rq2_source_dispersion}
\end{minipage}
\hfill
\begin{minipage}[b]{0.48\textwidth}
    \vspace{0pt}
  \centering
  \includegraphics[width=\linewidth]{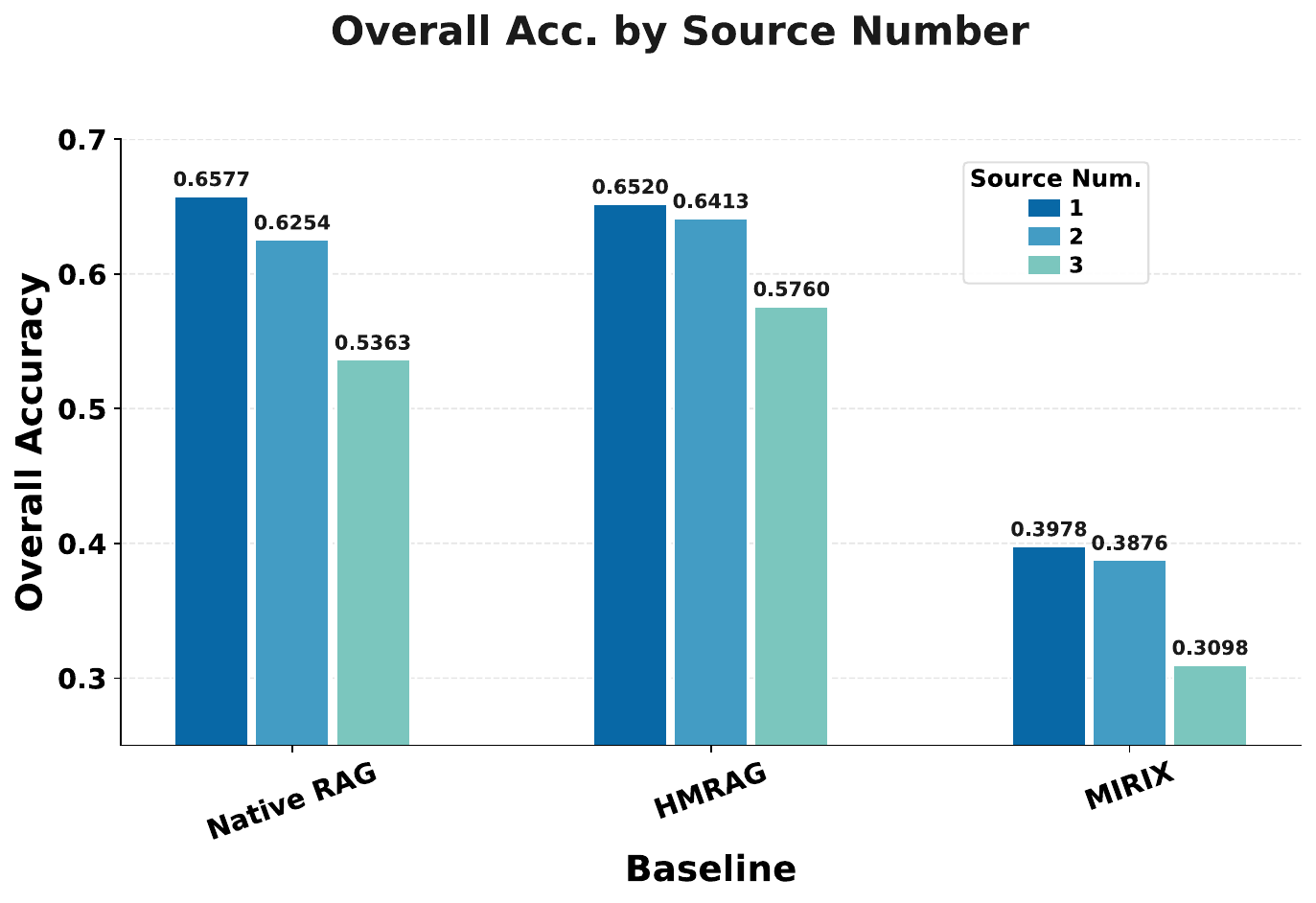}
  \vspace{-0.5cm}
  \caption{Experiment on the Different Source Numbers}
  \label{fig:rq3_source_num_accuracy}
\end{minipage}
\vspace{-0.5cm}
\end{figure*}

We compare \emph{source-concentrated} and \emph{source-distributed} evidence settings to evaluate whether dispersing the same supporting evidence across multiple sources increases the difficulty of memory-grounded reasoning. To keep the comparison fair, we fix the QA pairs and their gold evidence content, and only reorganize where that gold evidence is placed. Concretely, for each QA sample, we swap the chunks containing its gold evidence with chunks from other sources so that all supporting evidence is concentrated into one source in the \emph{source-concentrated} condition, while preserving its original placement across multiple sources in the \emph{source-distributed} condition; we then lightly polish the local scaffold for fluency without changing the evidence content. This design changes only whether evidence is concentrated in one source or distributed across sources, while keeping context budget matched across settings.

According to the overall scores in Figure~\ref{fig:rq2_source_dispersion}, the performance of all three representative baselines consistently drops from the \emph{source-concentrated} setting to the \emph{source-distributed} setting. This result supports the view that the primary challenge comes from the \emph{source-distributed} design itself. In the source-concentrated setting, the required evidence remains accessible within a single local context. By contrast, in the source-distributed setting, models must locate the relevant independent sources, connect their separate local contexts, and compose evidence that is individually incomplete but jointly sufficient. This highlights that \textbf{the difficulty is not reading more content, but coordinating reasoning across independently originated sources}. Appendix~\ref{appendix:source_distribution_experiment} task-level breakdowns follow the same pattern.


\vspace{-0.2cm}
\subsection{RQ3: Impact of the Number of Sources}
\vspace{-0.2cm}
To better investigate the effect of source distribution, we analyze the performance of representative baselines on samples with different numbers of required sources. Evaluation samples with different source numbers are interleaved within the same benchmark environments, and are therefore evaluated against source pools of similar overall length and distracting content, while differing primarily in how the critical evidence is distributed across those sources. Since the task coverage is not identical across evidence number buckets, we restrict this analysis to the overlapping task subset (\textbf{S.H.}, \textbf{M.H.}, and \textbf{P.R.}) and report the unweighted average scores over shared tasks. The results are shown in Figure~\ref{fig:rq3_source_num_accuracy}.


Although the absolute performance levels differ across baselines due to their different algorithmic architectures, the direction of change is fully consistent: all methods perform worse as the number of required sources increases. This consistency suggests that \textbf{dispersion of sources itself is an important difficulty factor in multimodal agent memory}, rather than very long context alone. As the evidence required by a question becomes more distributed across more independent sources, the agent must retrieve, preserve, and integrate increasingly dispersed evidence, which makes the task substantially harder. In our benchmark, the challenge is not merely reading more multimodal content, but identifying which sources jointly matter, recovering partial clues scattered across them, and composing these fragments into a coherent final answer.

\vspace{-0.2cm}
\subsection{RQ4: Impact of Retrieval Budget}

We vary the retrieval budget by changing the number of retrieved entries $K$ to examine the trade-off between evidence coverage and retrieval noise. Table~\ref{tab:retrieval_recall_precision} shows a clear first-order pattern across representative retrieval-based baselines: as $K$ increases, Recall@$K$ generally rises, while Precision@$K$ declines. This trend indicates that larger retrieval pools make it easier to surface relevant evidence from sources, but also introduce more irrelevant or weakly relevant candidates into the retrieved set. 
In summary, \textbf{larger retrieval budgets improve evidence coverage but also increase retrieval noise, making downstream evidence filtering and composition more difficult}.

\begin{table*}[t]
  \centering
  \small
  \vspace{-0.5cm}
  \setlength{\tabcolsep}{5pt}
  \renewcommand{\arraystretch}{1.15}
  \caption{Retrieval Recall@$K$ and Precision@$K$ by baseline and retrieval budget~$K$.}
  \label{tab:retrieval_recall_precision}
  \begin{tabular}{lcccccccc}
    \hline
    \multirow{2}{*}{\textbf{Baseline}}
      & \multicolumn{4}{c}{\textbf{Recall}}
      & \multicolumn{4}{c}{\textbf{Precision}} \\
    \cline{2-5}\cline{6-9}
    & \textbf{$K{=}10$} & \textbf{$K{=}20$} & \textbf{$K{=}50$} & \textbf{$K{=}100$}
    & \textbf{$K{=}10$} & \textbf{$K{=}20$} & \textbf{$K{=}50$} & \textbf{$K{=}100$} \\
    \hline
    Reflexion Mem.   & 0.2832 & 0.2835 & 0.2835 & 0.2801 & 0.0746 & 0.0374 & 0.0149 & 0.0073 \\
    Native RAG       & 0.4336 & 0.6355 & 0.7313 & 0.7903 & 0.1143 & 0.0837 & 0.0389 & 0.0207 \\
    HMRAG            & 0.2677 & 0.4099 & 0.5159 & 0.5244 & 0.0705 & 0.0540 & 0.0272 & 0.0138 \\
    VRAG             & 0.0981 & 0.2181 & 0.2489 & 0.2485 & 0.0260 & 0.0287 & 0.0131 & 0.0066 \\
    \hline
  \end{tabular}
  \vspace{-0.2cm} 
  \renewcommand{\arraystretch}{1.0}
\end{table*}

This trade-off is especially important under source-distributed settings, where the supporting evidence is scattered across multiple independent sources rather than concentrated in a single local context. With a small retrieval budget, systems can easily miss crucial clues because the evidence needed for one question may be fragmented across different chats, files, documents, or other sources. Increasing $K$ helps improve coverage over these dispersed sources and therefore reduces the risk of source-level omission. However, it also introduces more irrelevant or weakly related candidates, making downstream source selection and evidence composition harder. From the perspective of a source-distributed memory environment, \textbf{larger retrieval budgets are therefore not unconditionally better}: they improve the chance of covering the necessary sources, but at the cost of a noisier candidate pool that places greater demands on subsequent filtering, grounding, and cross-source composition. End-task results in the Appendix~\ref{appendix:retrieval_experiment} are broadly consistent with this insight, suggesting that larger $K$ often helps in source-distributed evaluation, although the gains depend on how effectively each system can identify and combine the truly relevant sources.

\subsection{RQ5: Failure Analysis}

\begin{figure*}[ht]
\centering
\vspace{-0.5cm}
\begin{minipage}[t]{0.4\textwidth}
    \vspace{0pt}
  \centering
  \includegraphics[width=\linewidth]{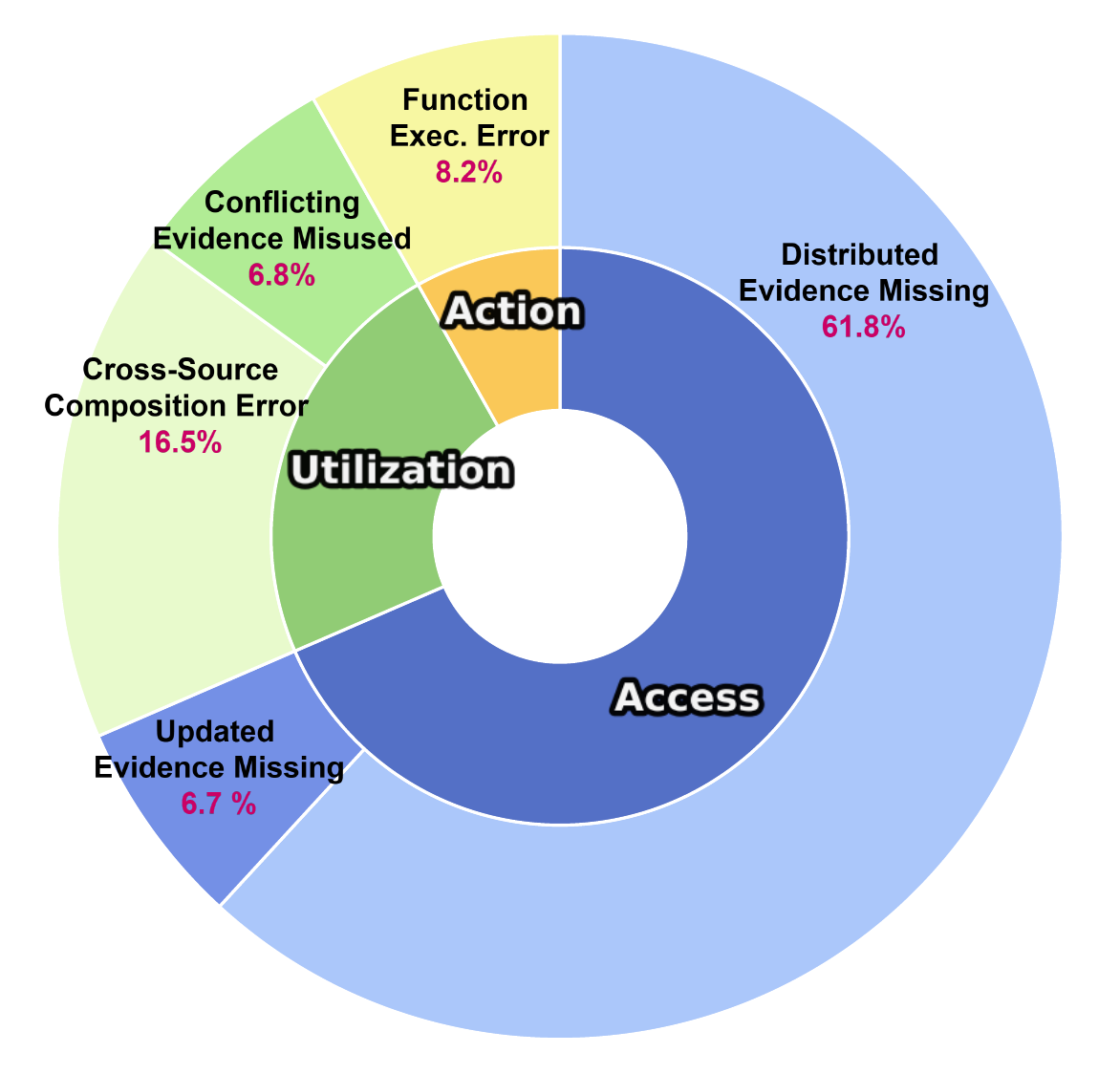}
  \caption{Error Diagnosis Categories}
  \label{fig:rq5_failure_analysis}
\end{minipage}
\hfill
\begin{minipage}[t]{0.56\textwidth}
To better understand where current systems fail on \name, we use gpt-4.1 as an LLM judger to diagnose 600 sampled error cases aggregated from results of the above baselines as shown in Figure~\ref{fig:rq5_failure_analysis}. The dominant pattern is still evidence access failure: most diagnosed errors arise because the system can not surface the needed evidence from distributed memory. Two additional patterns also stand out. First, \textbf{a non-trivial portion of errors comes from evidence utilization rather than evidence access alone}: even when some relevant information is available, agents still fail to correctly compose clues across sources or to prioritize updated evidence over stale or conflicting records. Second, \textbf{a smaller but clear portion of failures appears at the action stage, where models do not convert remembered information into correct executable outputs.}
\end{minipage}
\end{figure*}
Taken together, these diagnostics suggest that the main challenge of \name is layered: systems first struggle to recover the right distributed evidence, and then continue to fail on cross-source composition, temporal prioritization, and action grounding. Details of the LLM-judge setup are provided in Appendix~\ref{appendix:additional_experiment}. Cases of failure can be seen in Appendix~\ref{appendix:diagnosis_experiment}.

\section{Conclusion}
We introduced \emph{Source-distributed Multimodal Memory Bench} (\name), containing $1877$ samples and $5$ types of tasks, to address an under-evaluated challenge in multimodal agent memory: whether systems can retrieve, align, and compose multimodal evidence distributed across independently originated sources. Experiments on representative memory-style and retrieval-based baselines show that current systems remain far from effective, with substantial room for improvement even for the strongest methods in a source-distributed evaluation environment. Overall, our findings suggest that progress in multimodal agent memory should be evaluated not only by performance within coherent or pre-assembled contexts, but also by whether systems can retrieve and compose source-distributed evidence into reliable answers and actions.

\bibliographystyle{unsrtnat}
\bibliography{references}

\appendix
\clearpage
\startcontents[appendix]

\section*{Appendix Contents}
\printcontents[appendix]{}{1}{}

\clearpage
\section{Problem Formulation}
\label{appendix:problem_formulation}
We formulate \name as a memory-grounded question answering and action prediction benchmark over \emph{source-distributed} multimodal evidence. The key distinction from standard long-context evaluation is that the relevant evidence is not merely far apart within one long input, but dispersed across multiple \emph{independently originated sources} that must be identified and composed.

\paragraph{Source Objects.}
A \emph{source} in our benchmark is a coherent memory object with its own local context and organizational boundary, such as a group chat, a private thread, a profile page, a document, a table, an image, or another file-like artifact. Each source is internally meaningful on its own, but is not constructed to directly present the final answer for the benchmark query. In this sense, a source is not just a chunk of a larger context; it is an independently originated object that may differ from other sources in participants, purpose, format, temporal role, and informational scope.

Formally, each sample contains a set of sources
\begin{equation}
\mathcal{S} = \{S_1, S_2, \dots, S_m\},
\end{equation}
where each source $S_i$ consists of one or more evidence-bearing items:
\begin{equation}
S_i = \{o_{i,1}, o_{i,2}, \dots, o_{i,n_i}\}.
\end{equation}
Each item is represented as
\begin{equation}
o_{i,j} = \langle x_{i,j}, s_i, \tau_{i,j} \rangle,
\end{equation}
where $x_{i,j}$ is the content of the item, $s_i$ is the source identity shared by all items in source $S_i$, and $\tau_{i,j}$ is its timestamp or local temporal position. The content $x_{i,j}$ may include text, images, tables, document pages, or other forms of multimodal evidence.

\paragraph{What Counts as Source-Distributed.}
We say that a benchmark sample is \emph{source-distributed} if its answer-critical evidence cannot be recovered from a single source alone, and instead must be composed from evidence distributed across at least two independently originated sources. Let $\mathcal{E}(q)$ denote the set of minimal evidence items required to answer question $q$. A sample is source-distributed if
\begin{equation}
|\{\, s(o) \mid o \in \mathcal{E}(q) \,\}| \geq 2,
\end{equation}
that is, the required evidence spans at least two distinct source identities. This criterion distinguishes our setting from ordinary long-context reasoning within a single conversation or document, where evidence may be far apart but still belongs to one unified source.

Under this definition, long-range dependencies within one chat or one document are not sufficient by themselves to qualify as source-distributed. They remain challenging, but they are treated as \emph{within-source long-context} difficulty rather than \emph{cross-source memory composition}. By contrast, when one clue appears in a private chat, another in a project document, and a final update in a profile or image, the agent must reason across source boundaries, which is the target difficulty of \name.

\paragraph{Memory Construction and Query Answering.}
Given the source set $\mathcal{S}$, the agent incrementally observes items from these sources and maintains an external memory state $M_t$. For conversational sources, the observations follow their turn order; for non-conversational sources such as documents or images, the observations correspond to their associated source items. We denote the overall observation stream as
\begin{equation}
\mathcal{O} = \{o_1, o_2, \ldots, o_T\},
\end{equation}
where each observation retains its source identity. As each observation arrives, the memory is updated by
\begin{equation}
M_{t+1} = \Phi(M_t, o_t),
\end{equation}
where $\Phi$ is the memory update operator.

After the full source environment has been ingested, the agent receives a question $q$ and retrieves relevant memory units:
\begin{equation}
M_{\mathrm{ret}} = R(M_T, q).
\end{equation}
The final answer is then generated as
\begin{equation}
y = G(q, M_{\mathrm{ret}}).
\end{equation}

Under this formulation, success requires more than recalling isolated facts from a long input. A successful system must (1) preserve source-aware memory over heterogeneous source objects, (2) retrieve evidence spanning the right source boundaries, and (3) compose or reconcile these pieces into a coherent final answer or action. In this sense, \name evaluates \emph{source-distributed memory composition} rather than only long-context multimodal recall.

\section{Benchmark Construction Details}
\label{appendix:dataset_construction_details}

\subsection{Sources of Benchmark}
All basic QA supervision in \name is derived from public open-source datasets and benchmarks with reusable images, documents, or textual annotations. We group the sources by the target capability they primarily support.

\textbf{Single-Hop QA.} Single-hop samples are mainly collected from \textsc{ChartQA\_Pro}\footnote{\url{https://github.com/vis-nlp/ChartQAPro}}~\citep{Masry2025ChartQAProAM}, \textsc{SlideVQA} \footnote{\url{https://github.com/nttmdlab-nlp/SlideVQA}}~\citep{tanaka2023slidevqa}, and \textsc{MMDocRAG} \footnote{\url{https://github.com/MMDocRAG/MMDocRAG}}~\citep{dong2025benchmarking}. These sources provide strong multimodal supervision over charts, slides, and document pages, and naturally expose image- and document-centric evidence objects. For \textsc{SlideVQA} and \textsc{MMDocRAG}, we prefer samples with multiple gold pages or multiple gold evidence items so that the final benchmark instance remains compatible with our multi-source construction goal.

\textbf{Multi-Hop QA.} Multi-hop samples are mainly sourced from \textsc{MMCV} \footnote{\url{https://github.com/mmcv-dataset/MMCV}}~\citep{wang-etal-2025-piecing}. We focus on subsets whose reasoning chain already spans multiple supporting items, which makes them suitable for later conversion into cross-source memory questions. In practice, we prioritize longer-hop subsets because they better stress evidence aggregation after insertion into long-running conversations.

\textbf{Conflict Resolution.} Conflict-oriented samples are mainly sourced from \textsc{MLLMKC} \footnote{\url{https://github.com/MLLMKCBENCH/MLLMKC}}~\citep{jia2026benchmarking} and \textsc{MMKE} \footnote{\url{https://github.com/MMKE-Bench-ICLR/MMKE-Bench}}~\citep{du2025mmke}. These datasets are useful because they explicitly encode knowledge updates, contradictions, or stale-vs.-new evidence. We use them to construct settings in which a model must decide which memory item is valid after multiple partially conflicting mentions have been woven into the interaction history.

\textbf{Preference Reasoning.} Preference-focused samples are mainly sourced from \textsc{MMPB} \footnote{\url{https://github.com/AIDASLab/MMPB}}~\citep{kim2025mmpb} and \textsc{LifeSim} \footnote{\url{https://github.com/dfy37/lifesim}}~\citep{duan2026lifesim}. \textsc{MMPB} provides explicit preference statements together with question-relevant multimodal context, while \textsc{LifeSim} is particularly useful for implicit preference signals that unfold over dialogue. Their combination helps us cover both explicitly stored user constraints and preference information that must be inferred from recurring behavior.

\textbf{Function Calling.} Function-calling samples are mainly sourced from \textsc{M3Bench} \footnote{\url{https://github.com/EtaYang10th/Open-M3-Bench}}~\citep{zhou2025m}. From these examples we extract the candidate tool set, the multimodal or file-based operating context, and the gold function-call output. We only retain samples whose action target can be evaluated as a single grounded call rather than a long cascading workflow.

In addition to QA supervision, we construct lightweight agent profiles and conversational identities from public persona-style and profile-style resources. These materials are not used as evaluation targets themselves; instead, they provide the surrounding user and collaborator context required for long-horizon conversation simulation.

\subsection{Open-Source Dataset Preprocess}
\label{appendix:open_source_dataset_preprocess}

\begin{figure}[t]
    \centering
    \includegraphics[width=0.8\linewidth]{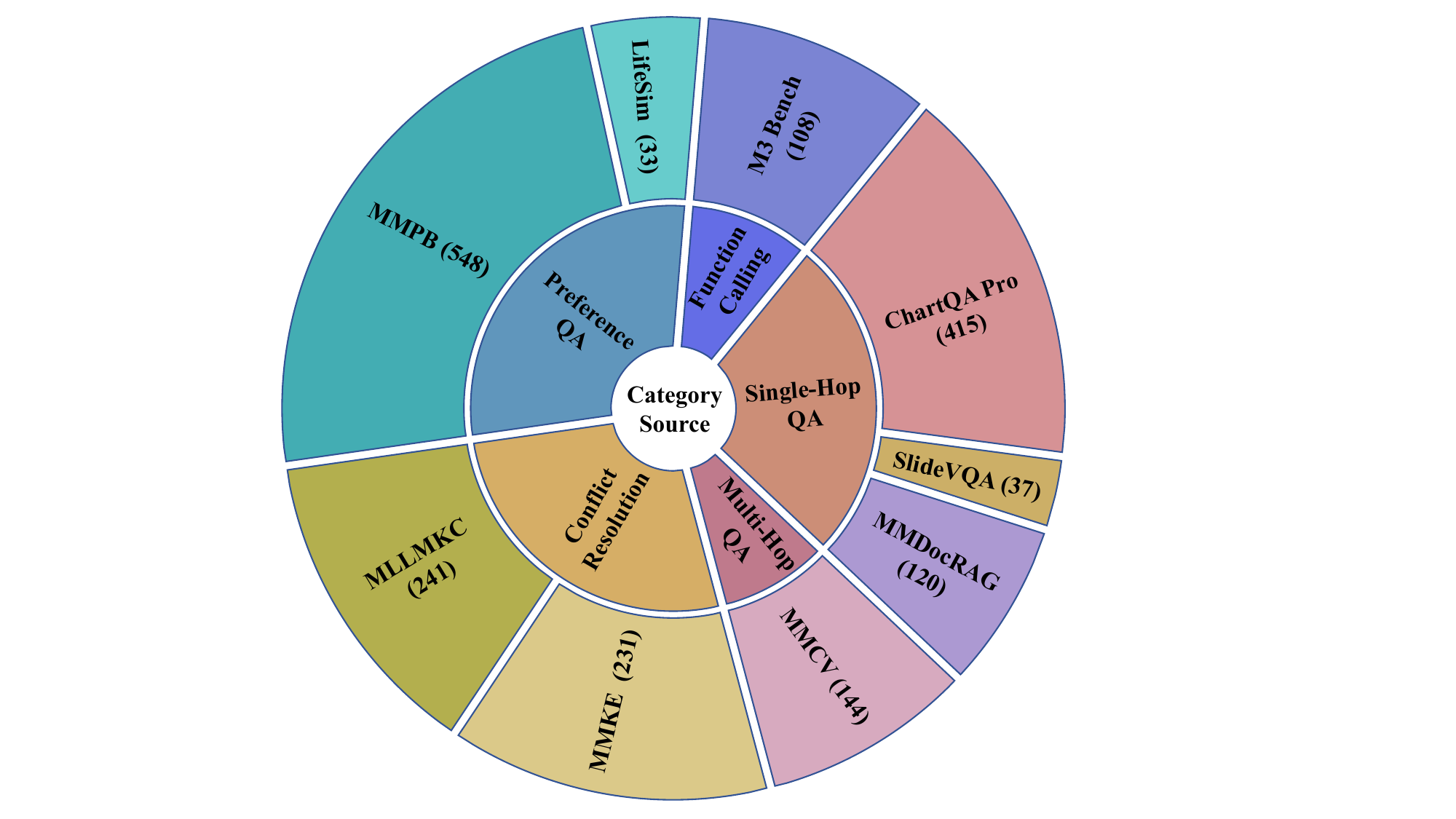}
    \caption{Open-Source Datasets of \name}
    \label{fig:appendix_open_source_datasets}
\end{figure}

Before entering the three-stage construction pipeline, all open-source samples are normalized into a shared schema consisting of a question, a gold answer, and multiple evidence units annotated by source type and modality. This preprocessing step is necessary because the raw sources expose supervision in heterogeneous forms: some use highlighted document spans, some point to images or pages, some provide structured preference statements, and others specify tool arguments or state variables.

For \textsc{ChartQA\_Pro}, we ask a strong LLM to derive a concise textual condition from the original question, answer, and chart image. The generated condition captures the intended interpretation angle or constraint without leaking the chart content verbatim. We then rewrite the original QA pair so that solving it requires jointly using the textual condition and the chart itself, rather than exploiting surface shortcuts from either side alone.

For \textsc{MMDocRAG}, we retain retrieved pages and golden evidence quotes, then partition long documents into smaller file-like units aligned with the gold supporting regions. This conversion makes the source more compatible with a memory benchmark in which evidence should appear as realistic files, pages, or attachments rather than as one monolithic context block.

For \textsc{LifeSim}, we keep preference instances whose supervision is expressed implicitly across dialogue. To make them compatible with multimodal evaluation, we replace some preference-bearing entities or activities with corresponding image evidence and lightly rewrite the surrounding turns to avoid directly leaking the image content through text. This turns implicit preference reasoning into a jointly text-and-image grounded task.

For \textsc{M3Bench}, we retain single grounded function-call instances and remove cases that require long cascaded execution chains. We additionally synthesize short procedural or contextual textual evidence so that the final benchmark sample contains both task-oriented action requirements and memory-relevant descriptive context.

For \textsc{SlideVQA}, \textsc{MMCV}, \textsc{MLLMKC}, \textsc{MMKE}, and \textsc{MMPB}, preprocessing is comparatively light. Beyond filtering for samples with sufficiently rich evidence and converting the raw inputs into the unified schema, we preserve their native supervision with minimal rewriting.

After task-specific normalization, all samples pass through three shared quality-control stages. First, we generate distractor options for the multiple-choice tasks; these distractors are designed to be semantically plausible and partially related to the evidence rather than trivial negatives. Second, we run answer verification, where a strong LLM re-solves the rewritten item under majority-vote prompting to detect malformed or ambiguous samples. Third, we perform evidence verification by masking one evidence unit at a time and retaining only samples whose answerability degrades as expected, which helps ensure that the retained evidence items contain genuinely necessary information. The surviving pool is then passed to manual spot checking before conversation generation.

We construct each benchmark sample through a staged pipeline that explicitly separates {evidence construction}, {conversation simulation}, and {evidence grounding}. This design is motivated by the goal of benchmarking source-distributed, multimodal agent memory: the evaluation sample should not merely contain long conversations, but should also require the model to integrate evidence distributed across heterogeneous sources such as chats, images, tables, documents, and file-like artifacts. Accordingly, our construction process starts from question-answer pairs and evidence collected from open-source datasets and benchmarks, then grows a profile-consistent long-horizon conversation environment around them, and finally inserts the processed evidence back into the conversation stream with locally coherent transitions.

\subsection{QA Preprocessing}
\label{section:dataset_construction_sourcing}
In this stage, we build the QA pool by collecting samples from a diverse set of public datasets and benchmarks spanning multi-hop reasoning, conflict resolution, preference inference, document QA, and workflow-oriented function calling. Preprocessing at this stage is aimed at deriving, from these open-source resources, a collection of controllable evaluation QA samples: each sample comprises a question-answer pair, and multiple multimodal evidence items, which allows us to preserve answerability while substantially enriching the surrounding memory context. Introductory descriptions of each constituent dataset and benchmark, together with the corresponding basic preprocessing steps, are provided in the appendix.

We first normalize each sample sourced from the open-source datasets and benchmarks. Concretely, we reorganize the raw supporting information for every sample into unified evidence units, each annotated with its modality and corresponding multimodal content. This preprocessing is essential because the original datasets expose supervision in very different forms: some provide textual snippets, some provide image references, some involve tables or schedules, some surface evidence through documents or file-like artifacts typical of office and collaboration workflows, and some require state variables for function-calling decisions. By converting them into a common schema of multimodal memory objects, we can control how they are woven into long-horizon conversation while still recording well-defined multi-source evaluation in later pipeline stages.

After normalization, we apply modality-aware verification and refinement. We first filter out samples that remain answerable from only a single evidence item or a single modality alone, which are too simple in multi-source, multimodal settings. For the remaining samples that contain multiple multimodal evidence, we run a modality-aware ablation test: we repeatedly mask all evidence from one certain modality, prompt a strong LLM to answer using only the unmasked modalities, and retain a sample in the high-quality pool only if the model fails under every such mask---equivalently, each modality contributes information that cannot be compensated by the others. Samples rejected by the initial screen are not necessarily discarded; we instead ask a strong LLM to rewrite the question and answer so that the pair does not trivially leak surface cues from the evidence, which recovers additional usable samples while reducing accidental shortcut solutions.

In addition, we prompt a strong LLM to generate lightweight structured metadata for each retained sample, including coarse topical descriptors and modality-aware captions that briefly describe non-textual evidence. These fields support later filtering, inspection, and conditioning during long-horizon dialogue synthesis without substituting for the underlying multimodal content. The output of this stage is a curated pool of QA samples with verified multimodal evidence that can later be woven into simulated agent conversational environments.

\subsection{Conversational Source Synthesis}
\label{section:dataset_construction_conversation_generation}

Once evidence-augmented QA samples are ready, we build the conversational sources in which their evidence will eventually reside. We assemble each pre-configured cast into multiple group channels and private one-to-one threads, then simulate several independent multi-turn dialogues in which those agents interact under their assigned profiles. This yields a pool of long-horizon chat trajectories that later serve as the sources for evidence insertion.

Agent profiles are assembled from public persona-style attributes, lightweight demographic or workplace-style metadata, and sampled preference statements. These profiles constrain later dialogue generation, but they are not intended to dominate the surface form of every utterance. Instead, they act as soft behavioral anchors that keep recurring participants reasonably consistent over time.

After the participants are fixed, we generate topic plans for each channel. These topic plans are derived from evidence-bearing QA samples selected for the current batch, but they are phrased as ordinary project, personal, scheduling, or collaboration discussions rather than as explicit question-answering prompts. Each topic usually unfolds over multiple turns and is connected to adjacent topics through short transition spans, which helps the final history read as an organic conversation rather than a sequence of isolated evidence dumps.

After the participating agents and conversation channels for a cluster have been finalized, we randomly sample a handful of the high-quality evidence-augmented QA instances from the Stage~1 pool. For each sample, we propose one or more high-level conversation themes derived only from its coarse topical metadata and non-evidence cues, explicitly avoiding any wording that would reveal the associated multimodal evidence or shortcut the eventual answer. This separation is essential: conditioning dialogue synthesis on gold evidence too early yields unnaturally goal-directed exchanges and erodes the meandering style typical of everyday chat.

For each channel--theme assignment, we instruct agents to simulate ordinary multi-turn interactions with diverse discourse intents---clarifying, questioning, supporting, disagreeing, elaborating, or pivoting the focus---and we modulate per-turn response length to discourage rhythmic templates. Throughout generation, the simulator may attend only to prior turns in the same channel, each agent's profile-based preferences and constraints, and the active theme; it is not given access to QA evidence. The resulting long-horizon trajectories therefore interleave task-adjacent material with unrelated small talk, yielding a realistic conversational memory environment whose raw text and images still contain no direct leakage of the curated evidence objects.

Conversation generation proceeds in batches. Each batch instantiates a shared pool of agents, their lightweight profiles, and a set of candidate QA samples whose evidence will later be inserted into one or more channels. We create multiple parallel channels per batch, including group-like discussions and private exchanges, so that some agents naturally reappear across several sources. This design lets a final evaluation item draw on evidence that is distributed over heterogeneous but partially overlapping histories.

Utterances within a topic are generated turn by turn. Before each turn, the selected speaker receives the recent local context, profile constraints, the current topic intent, and a discourse act sampled from a small inventory such as agreement, rebuttal, elaboration, asking for clarification, summarization, or topic shifting. We also vary the expected utterance length. This control mechanism improves diversity and reduces the risk that all speakers collapse into the same tone or interaction pattern.

Evidence insertion is performed after the base dialogue skeleton is available. Rather than letting agents freely choose when to reveal evidence, we first synthesize a short local scaffold around each evidence item, including its conversational motivation and a small amount of before/after bridging context. We then insert the evidence object together with the scaffold into a temporally plausible location in one of the channels. This post-hoc insertion strategy gives us fine-grained control over where evidence appears, how it is mentioned, and which other sources remain available as distractors or complementary memories.

At the close of this stage, we hold a collection of long multi-turn conversations produced by role-playing the personalized agents across the configured channels. None of these conversations yet contains QA evidence: all multimodal support objects remain outside the conversation stream until the following dedicated insertion.

\subsection{Source-Aware Evidence Insertion}
\label{section:dataset_construction_evidence_insertion}

In this final stage, we weave preprocessed multimodal QA evidence into the long multi-turn conversations so that the augmented sources read as organically continued conversations rather than stitched attachments. 

A core design choice is multi-source dispersion: we deliberately route different evidence units to distinct interaction sources, such as separate group or private channels, documents or files. For QA samples whose gold evidence encodes information updates (superseding or conflicting facts over time), we further constrain insertion timestamps so that each newer evidence item is scheduled at least a minimum number of conversational turns after the older material it revises, which keeps chronologically fresh content strictly downstream of stale records. For other evidence, we still stagger timestamps so that supporting material never collapses into a single tidy exposition block. Once an interaction source has been chosen, we resolve fine-grained assignment by prompting a strong LLM with chunked excerpts of the host material together with the candidate evidence; the model proposes the insertion offset (e.g., a turn index for chat logs or a paragraph anchor for documents and files).

Rather than committing each evidence unit to an arbitrary turn, we first synthesize a compact, evidence-centered textual micro-thread: a localized multi-turn aside that situates the forthcoming artifact and its pragmatic stakes while withholding answer-bearing surface cues. These micro-threads function as insertion scaffolds, supplying register-matched dialogue through which multimodal evidence can enter the stream without clashing with topical momentum or channel-specific speaking style. After candidate anchors are screened for temporal plausibility, discourse coherence, and source diversity, we write both the evidence object and its adjoining scaffold into the history, then lightly generate a short span of bridging turns immediately before and after the inserted block so that the augmented segment blends into the pre-existing trajectory rather than reading as a pasted fragment. Taken together, this post-hoc scaffolding affords fine-grained control over where and in what conversational guise each multimodal item appears, while sustaining the phenomenology of an ongoing interaction.

Through this three-stage pipeline, each final benchmark sample contains (i) a question derived from curated open-source supervision, (ii) evidence distributed across multiple sources and modalities, and (iii) a profile-consistent long-context interaction history into which that evidence has been smoothly woven. This construction protocol is well aligned with our benchmark objective: evaluating whether an agent can use heterogeneous multi-source multimodal memories that emerge from realistic long-running interactions rather than from a single explicitly organized context.

\subsection{Data Quality Control}

\begin{table}[t]
\centering
\small
\renewcommand{\arraystretch}{1.12}
\begin{tabular}{llccc}
\toprule
\textbf{Source} & \textbf{Stage} & \textbf{\#Input} & \textbf{\#Output} & \textbf{Main Reason} \\
\midrule
\multirow{2}{*}{ChartQA\_Pro}
  & Correctness & 1948 & 1048 & non-unique answer \\
  & Evidence    & 1048 & 415 & Redundant evidence \\
\midrule
\multirow{2}{*}{SlideVQA}
  & Correctness & 454 & 388 & Incorrect answer \\
  & Evidence    & 388 & 37 &  Redundant evidence \\
\midrule
\multirow{2}{*}{\makecell[l]{MMDocRAG}}
  & Correctness & 419 & 311 & Incorrect answer \\
  & Evidence    & 311 & 120 &  Redundant evidence \\
\midrule
\multirow{2}{*}{\makecell[l]{MMCV}}
  & Correctness & 1687 & 290 & Incorrect answer \\
  & Evidence    & 290 & 144 & Redundant evidence \\
\midrule
\multirow{2}{*}{\makecell[l]{MMKE}}
  & Correctness & 1910 & 545 &  non-derivable answer \\
  & Evidence    & 545 & 231 & Redundant evidence \\
\midrule
\multirow{2}{*}{\makecell[l]{MLLMKC}}
  & Correctness & 520 & 518 & Incorrect answer \\
  & Evidence    & 518 & 241 & Redundant evidence \\
\midrule
\multirow{2}{*}{\makecell[l]{MMPB}}
  & Correctness & 9827 & 6190 & Incorrect / non-derivable answer \\
  & Evidence    & 6190 & 548 & Redundant evidence \\
\midrule
\multirow{2}{*}{\makecell[l]{LifeSim}}
  & Correctness & 119 & 38 & non-derivable answer \\
  & Evidence    & 38 & 33 & Redundant evidence \\
\midrule
\multirow{2}{*}{\makecell[l]{M3\_Bench}}
  & Correctness & 231 & 126 & non-unique answer \\
  & Evidence    & 126 & 108 & Redundant evidence \\
\bottomrule
\end{tabular}
\caption{Dataset construction funnel for data quality control. Each source benchmark first undergoes correctness verification and evidence necessity verification before entering the final curated pool.}
\label{tab:construction_funnel}
\end{table}

To ensure correctness, answerability, evidence dependency, and environmental naturalness, we adopt a two-stage data quality control pipeline: (\emph{i}) open-source QA preprocessing and (\emph{ii}) post-insertion validation. Table~\ref{tab:construction_funnel} summarizes the corresponding construction funnel. Across the source benchmarks used in this work, we begin with 17,115 raw candidate QA instances, retain 9,424 after correctness verification, and finally keep 1,877 after evidence necessity verification. The first stage therefore removes incorrect, unanswerable, non-unique, or weakly grounded QA candidates, while the second stage verifies that inserted evidence and its local scaffold are naturally integrated into the target conversational or memory environment, without introducing obvious artifacts, semantic inconsistencies, or shortcut leakage. Together, these procedures are designed to ensure that the final retained samples are not only answer-correct and evidence-dependent, but also natural and evaluable in a multi-source multimodal memory setting.

In the preprocessing stage, we first perform \textbf{LLM-based correctness verification} on all candidate QA instances. For each sample, we provide the evidence, question, and gold answer to a strong verifier model (\textsc{gpt-4.1}) and ask whether the answer is \emph{correct and derivable} from the given evidence. We define a positive judgment as ``the answer is correct and derivable'' and a negative judgment as ``the answer is incorrect or not derivable.'' The verification prompt also explicitly asks the model to consider answer uniqueness and decisiveness, so as to exclude cases with obvious ambiguity, multiple plausible answers, or insufficiently constrained evidence. To reduce single-run variance, each sample is verified three times independently, and we retain it only if at least two runs return positive judgments under majority voting. As shown in Table~\ref{tab:construction_funnel}, the main reasons for removal at this stage are incorrect answers, non-derivable answers, and non-unique answers, depending on the source benchmark.

We then conduct \textbf{evidence necessity verification}. For each sample that passes correctness checking, we iteratively mask one evidence item at a time and submit the remaining evidence, together with the original question and answer, to \textsc{gpt-4.1} under the same three-run majority-vote protocol. We retain a sample only if masking any single evidence item makes the answer no longer stably verifiable, i.e., at least half of the verification runs become negative under that ablation. This step removes instances with clearly redundant evidence or instances that can be solved from only a subset of the annotated support, and preferentially keeps samples in which each evidence item contributes materially to the final answer. The dominant removal reason at this stage is redundant or weakly grounded evidence, as also summarized in Table~\ref{tab:construction_funnel}.

After automatic filtering, we further perform \textbf{human spot-check review} on the retained candidates. Two expert annotators independently inspect a 10\% sample from each source benchmark and verify correctness, derivability, evidence dependency, and answerability. Concretely, the review checks whether (\emph{i}) the question and gold answer are correctly matched, (\emph{ii}) the answer can be reasonably derived from the annotated evidence, and (\emph{iii}) removing a key evidence item materially weakens answerability. The reviewers also inspect whether the sample contains ambiguity, multiple plausible answers, or shortcut cues that make the answer recoverable without the intended evidence. A sample is counted as passing only if it satisfies these criteria. This first-stage human audit achieves a pass rate of 96.3\%, providing additional support that the automatically retained samples are generally well-formed and properly grounded.

After evidence insertion, we conduct a second round of \textbf{human post-insertion review} to assess whether the inserted evidence and its scaffold are naturally integrated into the target memory environment. Two expert annotators inspect 5\% of the post-insertion samples, focusing on potential evidence leakage, contextual coherence, speaker and temporal consistency, and naturalness of local tone and phrasing. In particular, the review checks whether inserted content introduces shortcut cues, abrupt factual insertions, template-like stitching, or anomalous content blocks unrelated to the surrounding context. We also explicitly inspect whether captions of multimodal evidence reveal the answer too directly, so that captionized inputs do not create artificial shortcut solutions unavailable in the original source object. The post-insertion audit yields a pass rate of 92.8\%, suggesting that most inserted evidence and scaffolds remain natural and contextually compatible after integration.

Our leakage and ambiguity checks are therefore carried out in two complementary ways. At the preprocessing stage, ambiguity is controlled through the correctness-verification prompt, which explicitly rejects non-unique or insufficiently constrained answers, and through human review, which checks for multiple plausible answers or missing evidence support. At the post-insertion stage, leakage is controlled by manually inspecting whether local scaffold text, nearby turns, or generated captions reveal the answer more directly than intended, or whether they introduce shortcut cues that bypass the intended memory composition process.

\subsection{Benchmark Statistics}
\label{appendix:benchmark_statistics}
We present additional statistics in this section.

\begin{figure}[t]
    \centering
    \includegraphics[width=0.6\linewidth]{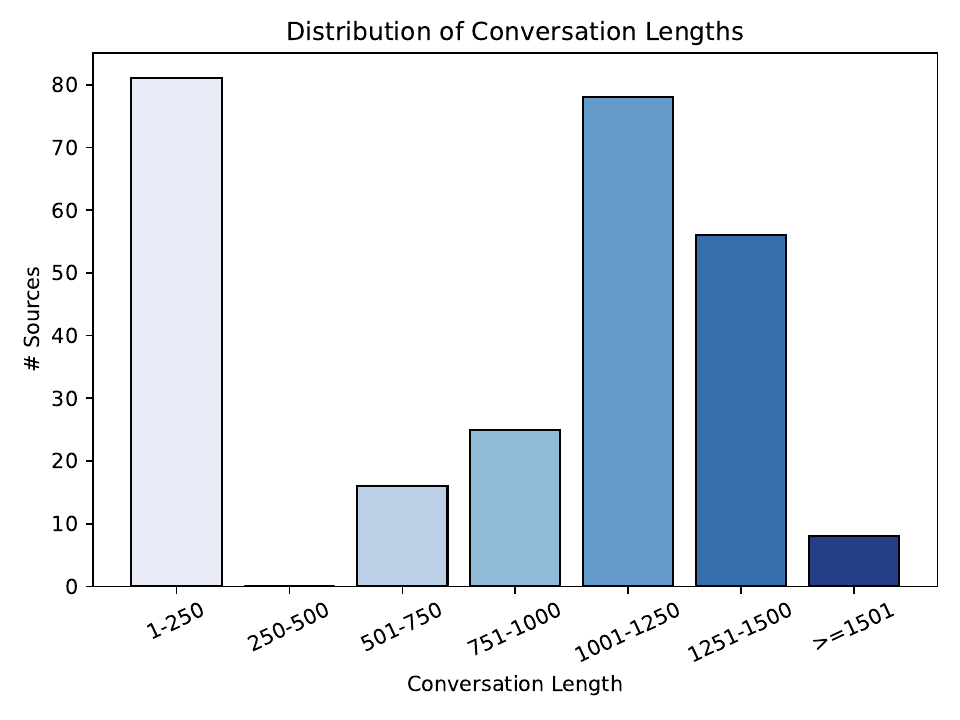}
    \caption{Distribution of conversation lengths across conversational sources in \name. Most conversation sources fall into either short (1--250) or long (1001--1500) ranges, indicating substantial variation in source length and context density.}
    \label{fig:appendix_distribution_conversation_lengths}
\end{figure}

The distribution of conversation lengths in Figure~\ref{fig:appendix_distribution_conversation_lengths} shows that conversational sources in the benchmark are not concentrated within a single length range, but instead cover both relatively short and relatively long contexts. In particular, the 1--250 and 1001--1500 ranges account for a large portion of the conversation sources, indicating that the dataset includes both compact local dialogues and long conversations with denser information and stronger cross-turn dependencies. This distribution has two implications. First, the benchmark does not reduce source-distributed memory to evidence retrieval in only short conversations. Second, even when an individual source is already long, the core challenge is still not merely long-context reading, but locating, relating, and integrating evidence across multiple independent sources.

\begin{figure}[ht]
    \centering
    \includegraphics[width=0.6\linewidth]{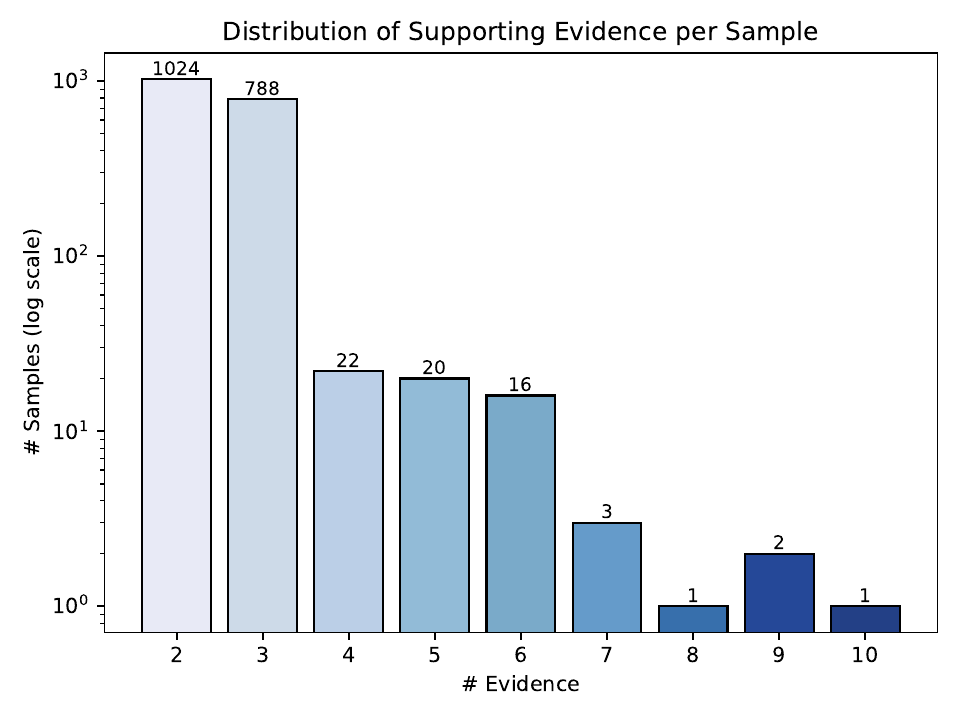}
    \caption{Distribution of the number of supporting evidence items per sample. Most samples require two or three evidence items, while a smaller but non-negligible subset requires four or more, forming a long-tailed composition difficulty.}
    \label{fig:appendix_evidence_distribution}
\end{figure}

\begin{figure}[ht]
    \centering
    \includegraphics[width=0.6\linewidth]{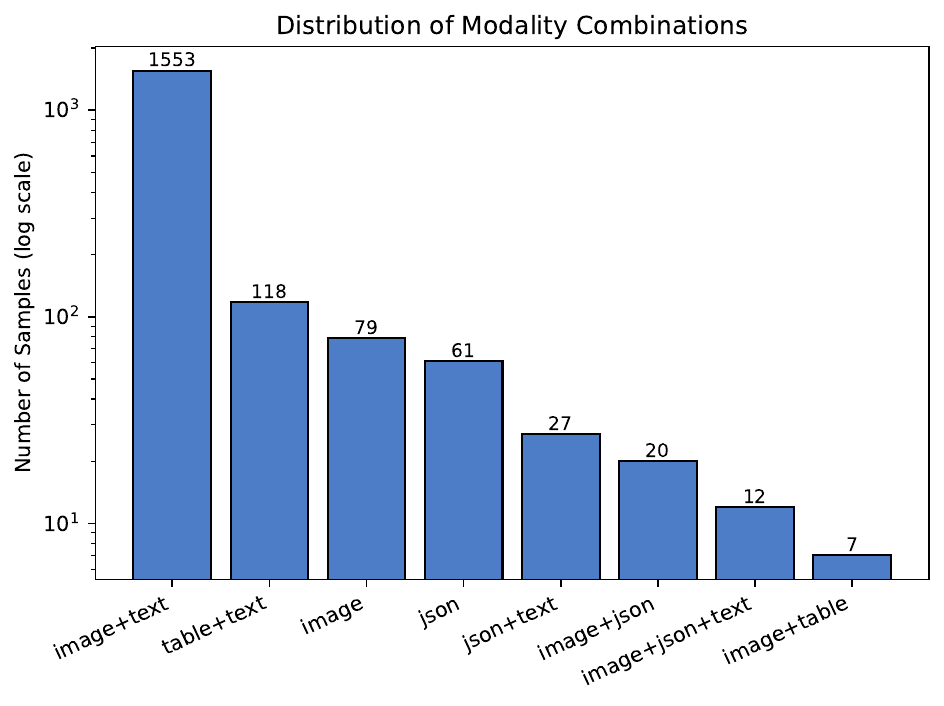}
    \caption{Distribution of modality combinations in \name. Image+text samples dominate the benchmark, while other combinations such as table+text, image-only, JSON-based, and mixed multimodal settings provide additional diversity. The y-axis is shown in log scale.}
    \label{fig:appendix_modality_distribution}
\end{figure}

The distribution of the number of supporting evidence items per sample in Figure~\ref{fig:appendix_evidence_distribution} shows that most samples require 2 to 3 pieces of supporting evidence to answer correctly, with samples requiring 2 or 3 evidence items making up the majority. At the same time, there is a clear long tail, where a smaller subset of samples requires 4+ or even more evidence items. This pattern suggests that the dominant difficulty in \name is not single-fact extraction, but multi-evidence composition. More importantly, when these pieces of evidence are further distributed across different sources, agent memory must do more than identify a single relevant clue: it must cover all necessary evidence and compose them at answer time. From a statistical perspective, this result supports the central design of the benchmark, namely that the main difficulty comes from the need to combine multiple local clues rather than solving the problem from a single source or a single evidence item.

The distribution of modality combinations in Figure~\ref{fig:appendix_modality_distribution} reveals a head-heavy but diverse pattern in the benchmark. image+text is by far the most dominant modality combination, indicating that joint visual-language understanding forms the main scenario in the dataset. At the same time, combinations such as table+text, image, json, json+text, and image+json are also explicitly represented. This distribution helps prevent the benchmark from being dominated by only one modality structure. On the one hand, the dominant combinations ensure that the benchmark remains grounded in common multimodal agent interaction settings. On the other hand, the tail of modality combinations introduces more heterogeneous source forms, requiring models to adapt to different information carriers and compose evidence across them. From the perspective of source-distributed memory, such modality diversity further increases the complexity of cross-source composition, because different sources are not only independent, but may also express complementary evidence in different modalities.

\section{Additional Experiments and Analysis}
\label{appendix:additional_experiment}

\subsection{Experiment Details}
\label{appendix:evaluation_pipeline_details}
In main experiments, all baselines are evaluated under a unified outer-loop protocol. Generally, for every sample in \name, we first prepare the memory environment in the format expected by the baseline, then issue the task query, collect the model output, and finally score it against the gold target.

For retrieval-oriented baselines, we serialize the memory environment into retrievable units. Depending on the baseline, these units may be chunked messages, document pages, image captions, or multimodal records. In our implementation, all retrieval experiments use the same embedding models, namely \texttt{bge-large-en-v1.5}\footnote{\url{https://huggingface.co/BAAI/bge-large-en-v1.5}}~\citep{bge_embedding} and \texttt{bge-m3}\footnote{\url{https://huggingface.co/BAAI/bge-m3}}~\citep{bge-m3}. Unless otherwise specified, the retriever selects the top-$K$ candidates with $K{=}20$, which is also the setting used for all main-table results. For text-only or Text+Caption settings, non-textual objects are replaced by captions generated in advance; for native multimodal settings, the original multimodal objects are preserved whenever the baseline supports them.

For memory-style baselines, we first replay the conversation history, files, and evidence objects into the model-specific memory interface. Some methods store memory as textual notes or summaries, some maintain structured records, and some combine persistent memory with later recall modules. For \textsc{MIRIX}, \textsc{MemGPT}, \textsc{Mem0}, and \textsc{MemVerse}, conversational memory is chunked into blocks of 12 dialogue turns with at most 2 images, and each chunk is further capped at 64{,}000 characters. Once the environment has been written, we ask the downstream question and record the model output under the baseline's native memory-access mechanism.

To reduce prompt-induced variance, we keep the evaluation instruction templates unified across methods at the answering stage. Whether a baseline answers from retrieved context concatenated into a reader prompt or returns an answer through its client-side memory interface, the final task prompt is kept the same in content and output-format constraints. The prompt specifies the task type, the answer format, and any restrictions such as returning only the multiple-choice option or only the final function call. Sample prompts are provided in Appendix~\ref{appendix:prompts}.

For function-calling evaluation, we parse model outputs by direct JSON decoding and compare the resulting structured call against the gold target under exact match. Outputs that fail JSON parsing are counted as incorrect. For captionized settings, all non-text evidence captions are generated uniformly with \textsc{gpt-4.1}; the caption prompts are listed later in Appendix~\ref{appendix:prompts}. The embedding-based components of our experiments were run on NVIDIA A100 80GB GPUs. Beyond the experiments reported in the paper, we did not use substantial additional compute resources for large-scale extra runs.

Unless otherwise noted, all reported experimental results are obtained with decoding temperature set to 0.01 for all experiments. Each evaluation is run three times independently, and we report the mean score across the three runs in order to reduce sampling noise from the backbone model and generation pipeline. The same scoring protocol is used for both open-book and retrieval-based settings.

\subsection{Main Results}
\label{appendix:main_results}
\begin{table*}[t]
  \centering
  \small
  \setlength{\tabcolsep}{5pt}
  \renewcommand{\arraystretch}{1.2}
  \caption{Main results on \name. \textbf{S.H.}, \textbf{M.H.}, \textbf{C.R.}, \textbf{P.R.}, and \textbf{F.C.} denote \textsc{Single-Hop QA}, \textsc{Multi-Hop QA}, \textsc{Conflict Resolution}, \textsc{Preference Reasoning}, and \textsc{Function Call}. All scores are averaged over 3 runs; standard deviations are shown for \textbf{S.H.}, \textbf{M.H.}, \textbf{C.R.}, and \textbf{P.R.}. The \textbf{Overall} reports the unweighted average score.}
  \label{tab:appendix_main_results}
  \resizebox{\textwidth}{!}{%
  \begin{tabular}{lcccccc}
  \hline
  \specialrule{1.1pt}{0pt}{0pt}
  \textbf{Baseline} & \textbf{S.H.} & \textbf{M.H.} & \textbf{C.R.} & \textbf{P.R.} & \textbf{F.C.} & \textbf{Overall} \\
  \hline
  \specialrule{1.1pt}{0pt}{0pt}
  Short-Term Mem.        & 0.2990 {\scriptsize$\pm$0.0139} & 0.2292 {\scriptsize$\pm$0.0199} & 0.2246 {\scriptsize$\pm$0.1648} & 0.2702 {\scriptsize$\pm$0.0097} & 0.0093 & 0.2064 \\
  Reflexion Mem.         & 0.5385 {\scriptsize$\pm$0.0108} & 0.4861 {\scriptsize$\pm$0.0096} & 0.3962 {\scriptsize$\pm$0.0046} & 0.3064 {\scriptsize$\pm$0.0028} & 0.0278 & 0.3510 \\
  Gen. Agent Mem.        & 0.3164 {\scriptsize$\pm$0.0107} & 0.2708 {\scriptsize$\pm$0.0170} & 0.2754 {\scriptsize$\pm$0.0209} & 0.2238 {\scriptsize$\pm$0.0106} & 0.0185 & 0.2210 \\
  Self Controlled Mem.   & 0.3129 {\scriptsize$\pm$0.0133} & 0.3403 {\scriptsize$\pm$0.0196} & 0.2881 {\scriptsize$\pm$0.0154} & 0.2203 {\scriptsize$\pm$0.0104} & 0.0185 & 0.2360 \\
  MIRIX (T+C)   & 0.3601 {\scriptsize$\pm$0.0100} & 0.2569 {\scriptsize$\pm$0.0284} & 0.3072 {\scriptsize$\pm$0.0017} & 0.2960 {\scriptsize$\pm$0.0134} & 0.0278 & 0.2496 \\
  MemGPT (T+C)  & 0.5734 {\scriptsize$\pm$0.0108} & 0.4722 {\scriptsize$\pm$0.0118} & 0.3326 {\scriptsize$\pm$0.0108} & 0.4389 {\scriptsize$\pm$0.0142} & 0.0370 & 0.3708 \\
  MemVerse               & 0.3549 {\scriptsize$\pm$0.0057} & 0.3264 {\scriptsize$\pm$0.0173} & 0.2691 {\scriptsize$\pm$0.0170} & 0.2754 {\scriptsize$\pm$0.0205} & 0.0093 & 0.2470 \\
  Mem0                   & 0.7430 {\scriptsize$\pm$0.0033} & 0.5903 {\scriptsize$\pm$0.0087} & 0.3665 {\scriptsize$\pm$0.0210} & 0.2909 {\scriptsize$\pm$0.0050} & 0.0398 & 0.4061 \\
  Native RAG             & 0.7797 {\scriptsize$\pm$0.0046} & 0.6667 {\scriptsize$\pm$0.0087} & 0.4068 {\scriptsize$\pm$0.0167} & 0.3683 {\scriptsize$\pm$0.0150} & 0.0741 & 0.4591 \\
  HMRAG                  & 0.8129 {\scriptsize$\pm$0.0014} & 0.7153 {\scriptsize$\pm$0.0000} & 0.5191 {\scriptsize$\pm$0.0069} & 0.3081 {\scriptsize$\pm$0.0008} & 0.1111 & 0.4933 \\
  MemGPT (MM)            & 0.3497 {\scriptsize$\pm$0.0021} & 0.5139 {\scriptsize$\pm$0.0100} & 0.2691 {\scriptsize$\pm$0.0115} & 0.1997 {\scriptsize$\pm$0.0123} & 0.0241 & 0.2713 \\
  MIRIX (MM)             & 0.5507 {\scriptsize$\pm$0.0079} & 0.3472 {\scriptsize$\pm$0.0147} & 0.4025 {\scriptsize$\pm$0.0020} & 0.2031 {\scriptsize$\pm$0.0061} & 0.0769 & 0.3161 \\
  OmniSimpleMem          & 0.2311 {\scriptsize$\pm$0.0014} & 0.0375 {\scriptsize$\pm$0.0065} & 0.2153 {\scriptsize$\pm$0.0036} & 0.1840 {\scriptsize$\pm$0.0125} & 0.0185 & 0.1373 \\
  UniversalRAG           & 0.3322 {\scriptsize$\pm$0.0109} & 0.3403 {\scriptsize$\pm$0.0017} & 0.2606 {\scriptsize$\pm$0.0134} & 0.1566 {\scriptsize$\pm$0.0064} & 0.0370 & 0.2253 \\
  VRAG                   & 0.4913 {\scriptsize$\pm$0.0141} & 0.4514 {\scriptsize$\pm$0.0118} & 0.3919 {\scriptsize$\pm$0.0185} & 0.1842 {\scriptsize$\pm$0.0077} & 0.0463 & 0.3130 \\
  \hline
  \specialrule{1.1pt}{0pt}{0pt}
  \end{tabular}
  }
  \renewcommand{\arraystretch}{1.0}
  \vspace{-0.2cm}
\end{table*}

\paragraph{Existing Methods Still Struggle on Source-Distributed, Multimodal Memory Tasks.}
The reference baselines again show that the benchmark is solvable but still far from effective. On the one hand, the \textsc{Golden Evidence Baseline} reaches 0.7473 Overall when the model is given the gold supporting evidence directly, confirming that the answer space itself is learnable once the required evidence is made available. On the other hand, realistic systems remain substantially below this reference condition. The strongest benchmarked baseline, \textsc{HMRAG}, reaches only 0.4933 Overall, leaving a gap of 0.2540 to the gold-evidence setting. This gap is large enough to suggest that the main difficulty does not lie in answer existence alone, but in whether a system can successfully retrieve, preserve, align, and compose the needed evidence under realistic source-distributed access.

The task-wise results further show that current methods are not uniformly weak in the same way, but instead exhibit clear capability imbalance across task families. In the Text+Caption setting, \textsc{Mem0} is quite strong on factual QA, reaching 0.7430 on \textsc{S.H.} and 0.5903 on \textsc{M.H.}, yet it drops to 0.2909 on \textsc{P.R.} and only 0.0398 on \textsc{F.C.}. \textsc{Native RAG} and \textsc{HMRAG} also perform relatively well on \textsc{S.H.} and \textsc{M.H.} (e.g., 0.7797/0.6667 for \textsc{Native RAG} and 0.8129/0.7153 for \textsc{HMRAG}), but remain much weaker on \textsc{F.C.} at 0.0741 and 0.1111. A similar pattern appears in the MM setting: \textsc{MIRIX} reaches 0.5507 on \textsc{S.H.} and 0.4025 on \textsc{C.R.}, but still only 0.0769 on \textsc{F.C.}. These differences indicate that no single system is consistently strong across all task families, and that source-distributed multimodal memory remains a broad, unsolved challenge rather than a narrow weakness confined to one task type.

\paragraph{Native Multimodal Access Alone Does Not Resolve the Source-Distributed Challenge.}
A clean comparison comes from paired results within the same memory system, where the primary change is whether the system has access to native multimodal memory or only Text+Caption representations. Under this paired view, the impact of native multimodal access is clearly not uniform. \textsc{MIRIX} improves from 0.2496 Overall in the Text+Caption setting to 0.3161 in the native MM setting, a gain of 0.0665. By contrast, \textsc{MemGPT} drops from 0.3708 to 0.2713 under the same switch, a decrease of 0.0995. These opposite trends suggest that native multimodal access is not by itself sufficient to overcome the benchmark difficulty.

This pattern is consistent with the interpretation that the core challenge is not merely whether a system can ``see'' native multimodal content, but whether it can use that content effectively once the required evidence is scattered across independent sources. Native multimodal inputs may preserve richer information than caption conversion in principle, but this advantage only becomes useful if the system can still retrieve the right sources, recover the right clues from them, and align those clues across source boundaries. In this sense, source-distributed memory places a stronger demand than modality access alone: it requires not only multimodal perception, but also robust cross-source memory access and evidence composition. The fact that native MM access helps \textsc{MIRIX} but hurts \textsc{MemGPT} further suggests that improvements in source-distributed settings depend on the interaction between modality access and the system's underlying retrieval-and-memory architecture, rather than on modality exposure in isolation.

\paragraph{Function Calling Exposes a Persistent Memory-to-Action Gap.}
Function calling remains the most challenging setting in the benchmark. Across all baselines, the best \textsc{F.C.} score is only 0.1111, achieved by \textsc{HMRAG}, and even the \textsc{Golden Evidence Baseline} reaches only 0.2778. This is far below its performance on the other task families, such as 0.8753 on \textsc{S.H.}, 0.7768 on \textsc{M.H.}, 0.8365 on \textsc{C.R.}, and 0.9699 on \textsc{P.R.}. The gap suggests that function calling is not difficult merely because it uses a structured output format, but because it requires models to recover precise arguments from source-distributed multimodal evidence and then convert that evidence into executable actions.

This difficulty is also clearly visible within individual systems rather than only across overall rankings. In the Text+Caption setting, \textsc{MemGPT} reaches 0.5734 and 0.4722 on \textsc{S.H.} and \textsc{M.H.}, but only 0.0370 on \textsc{F.C.}. \textsc{Mem0} reaches 0.7430 and 0.5903 on \textsc{S.H.} and \textsc{M.H.}, but only 0.0398 on \textsc{F.C.}. In the MM setting, \textsc{MIRIX} achieves 0.5507 on \textsc{S.H.}, 0.3472 on \textsc{M.H.}, and 0.4025 on \textsc{C.R.}, but only 0.0769 on \textsc{F.C.}. These row-wise contrasts show that relatively strong performance on recognition- or answer-oriented tasks does not readily translate into success on memory-grounded action prediction. Put differently, the source-distributed setting does not only make evidence harder to find; it also makes it harder to assemble fine-grained fields into an exact action specification. This is precisely why \textsc{F.C.} serves as a particularly strong stress test for memory systems that are expected to support downstream agent behavior.

\paragraph{Stronger Overall Performance Remains Highly Profile-Dependent.}
Beyond the overall ranking, the main results show that stronger systems succeed through noticeably different capability profiles rather than through a common recipe. \textsc{HMRAG} achieves the best Overall score, 0.4933, with comparatively balanced performance across \textsc{S.H.} (0.8129), \textsc{M.H.} (0.7153), and \textsc{C.R.} (0.5191), although it still remains weak on \textsc{P.R.} (0.3081) and especially \textsc{F.C.} (0.1111). \textsc{Mem0}, by contrast, reaches a lower Overall score of 0.4061 but is unusually strong on factual QA, especially \textsc{S.H.} and \textsc{M.H.}, before dropping sharply on \textsc{P.R.} and \textsc{F.C.}. \textsc{MemGPT} is not among the strongest factual systems overall, but in the Text+Caption setting it remains one of the relatively stronger baselines on \textsc{P.R.} at 0.4389, suggesting that preference-sensitive memory use may favor a different storage-and-recall profile than evidence-heavy factual aggregation.

These contrasts indicate that the benchmark is separating systems not only by average accuracy, but by the \emph{shape} of their capability profile under source-distributed multimodal memory demands. Some methods appear to benefit more from retrieval-oriented factual access, while others remain relatively more stable on personalized or user-sensitive reasoning. This is a useful property of the benchmark, because it suggests that source-distributed multimodal memory is not a single-axis capability. Instead, it contains multiple interacting sub-problems, including factual evidence composition, preference-sensitive recall, conflict handling, and action-oriented assembly.

\paragraph{Memory-Style and Retrieval-Style Methods Fail Differently.}
A second clear pattern is that retrieval-oriented baselines tend to dominate the more factual and conflict-heavy parts of the benchmark, while memory-style systems are only occasionally competitive in narrower regions. In the Text+Caption setting, both \textsc{Native RAG} and \textsc{HMRAG} outperform most memory-style baselines on \textsc{S.H.}, \textsc{M.H.}, and \textsc{C.R.}. \textsc{Native RAG} reaches 0.7797/0.6667/0.4068 on these three tasks, and \textsc{HMRAG} reaches 0.8129/0.7153/0.5191. By contrast, several memory baselines such as \textsc{Short-Term Mem.}, \textsc{Gen. Agent Mem.}, and \textsc{Self Controlled Mem.} remain near the floor across most task families, with Overall scores of 0.2064, 0.2210, and 0.2360 respectively.

This gap suggests that when the supporting evidence is distributed across multiple independent sources, explicit retrieval remains a strong practical bias because it increases the chance that the downstream model can at least access the necessary evidence. However, the results also make clear that retrieval alone is not sufficient. Even the strongest retrieval-style systems remain far below the \textsc{Golden Evidence Baseline}, and they still degrade sharply on \textsc{P.R.} and \textsc{F.C.}. For example, \textsc{HMRAG} reaches only 0.3081 on \textsc{P.R.} and 0.1111 on \textsc{F.C.}, despite being the strongest Overall baseline. This suggests that source-distributed difficulty includes at least two layers: first, retrieving the right sources; and second, aligning and composing their contents into the final answer or action. Retrieval-oriented methods help most clearly with the first layer, but the second remains a substantial bottleneck.



\subsection{Source-Distribution Experiment}
\label{appendix:source_distribution_experiment}

For the comparison between source-centralized and source-distributed in Section~\ref{section:rq2_source_distribution_experiment}, we construct a matched control from the same underlying QA instance rather than comparing different questions. Starting from the original multi-source sample, we identify the chunks containing the gold evidence and swap them with non-gold chunks from other sources so that all gold evidence is relocated into a single source in the concentrated condition. We then apply light scaffold editing to maintain local fluency and speaker/context consistency after the swap. Importantly, this procedure keeps the question, answer, gold evidence content, and the surrounding context budget as stable as possible across the two conditions, while changing only whether the relevant evidence is dispersed across sources or concentrated within one source. 

\begin{figure*}[h]
\centering
\begin{minipage}[t]{0.32\textwidth}
  \centering
  \includegraphics[width=\linewidth]{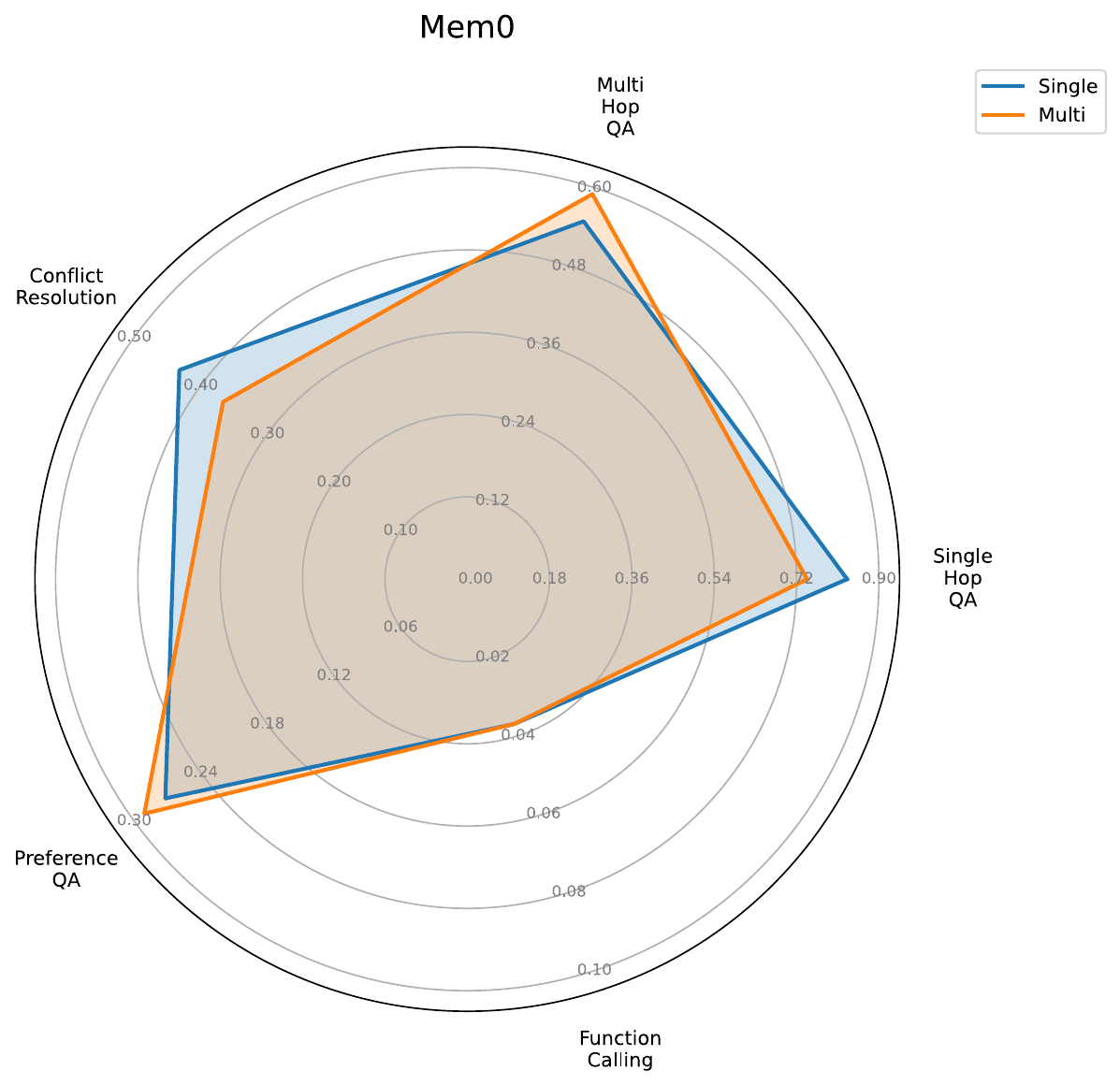}
  \caption{\textsc{Mem0}}
  \label{fig:appendix_rq2_mem0}
\end{minipage}\hfill
\begin{minipage}[t]{0.32\textwidth}
  \centering
  \includegraphics[width=\linewidth]{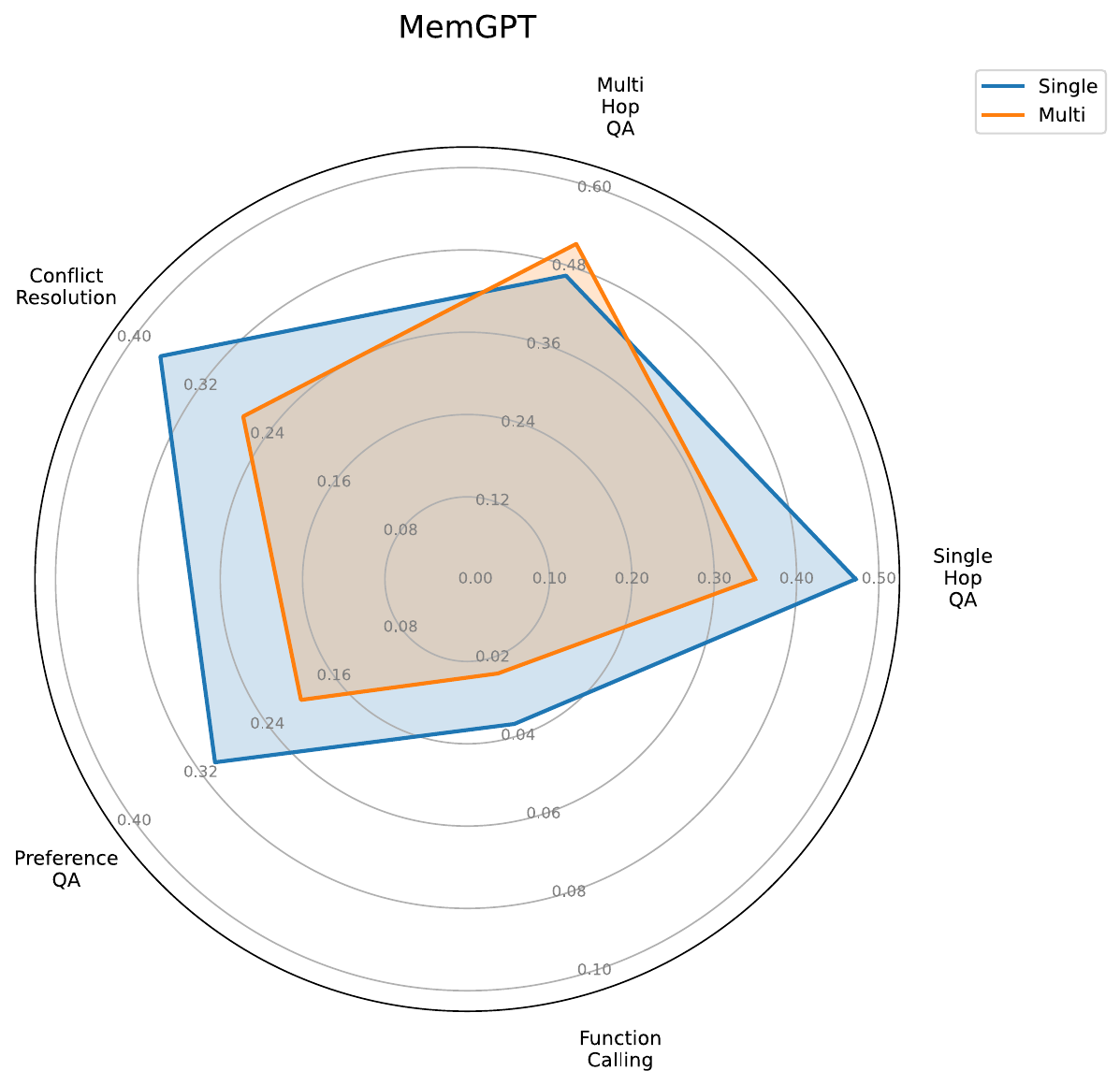}
  \caption{\textsc{MemGPT}}
  \label{fig:appendix_rq2_memgpt}
\end{minipage}\hfill
\begin{minipage}[t]{0.32\textwidth}
  \centering
  \includegraphics[width=\linewidth]{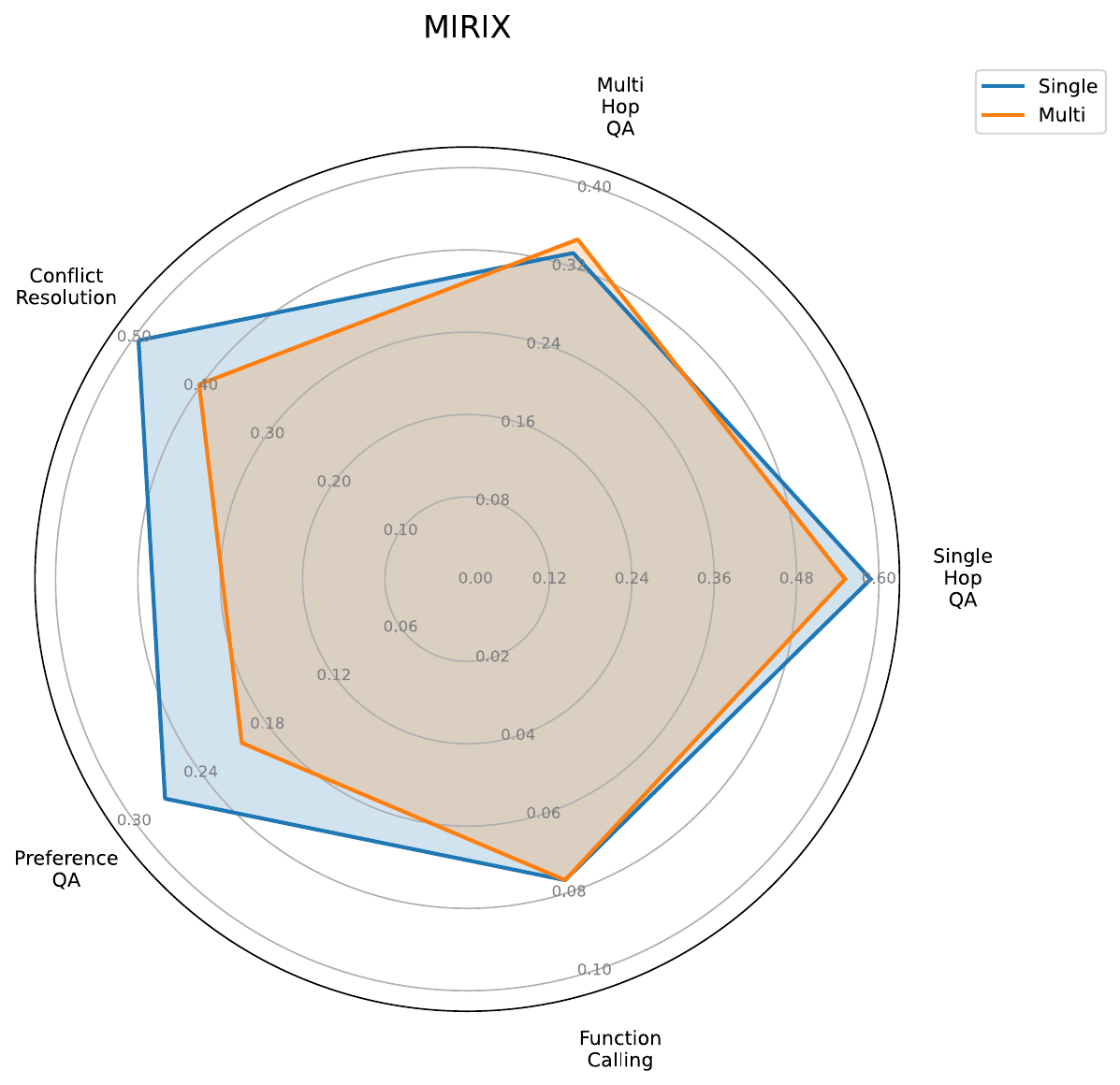}
  \caption{\textsc{MIRIX}}
  \label{fig:appendix_rq2_mirix}
\end{minipage}
\end{figure*}

This subsection reports the task-wise and model-wise breakdown for the single-source versus multi-source comparison summarized in the main text. In particular, we provide the per-task scores underlying Figure~\ref{fig:rq2_source_dispersion}.

The task-wise results show that the effect of source dispersion is not perfectly uniform across methods or task types, but several patterns are clear. First, all three representative systems suffer an Overall drop when moving from single-source to multi-source evidence, confirming that the degradation in the main paper is not driven by only one baseline. Second, the clearest and most consistent decreases appear on \textsc{Single-Hop QA} and \textsc{Conflict Resolution}: \textsc{Mem0}, \textsc{MemGPT}, and \textsc{MIRIX} all perform worse in the multi-source condition on these two tasks, suggesting that even apparently direct recall becomes less reliable once the supporting clue is no longer concentrated in one source, and that conflict-sensitive cases become harder when updated evidence must be located across sources.

Taken together, these task-wise comparisons refine the main-paper conclusion. Source dispersion does not degrade every task in exactly the same way, but it most reliably harms tasks that depend on locating one decisive clue or resolving updated evidence across sources, while its effect on multi-hop and preference reasoning is more interaction-dependent with the baseline's memory and composition strategy.

\subsection{Retrieval Experiment}
\label{appendix:retrieval_experiment}

\begin{table*}[t]
\centering
\small
\setlength{\tabcolsep}{5pt}
\renewcommand{\arraystretch}{1.15}
\begin{tabular}{llcccccc}
\toprule
\textbf{Baseline} & \textbf{$K$} & \textbf{S.H.} & \textbf{M.H.} & \textbf{C.R.} & \textbf{P.R.} & \textbf{F.C.} & \textbf{Overall} \\
\midrule
\multirow{4}{*}{Reflexion Mem.}
& 10  & 0.5297 & 0.4444 & 0.3983 & 0.3029 & 0.0463 & 0.3443 \\
& 20  & 0.5385 & 0.4861 & 0.3962 & 0.3064 & 0.0278 & 0.3510 \\
& 50  & 0.5647 & 0.4792 & 0.3644 & 0.3012 & 0.0463 & 0.3512 \\
& 100 & 0.5524 & 0.4931 & 0.3877 & 0.3029 & 0.0463 & 0.3565 \\
\midrule
\multirow{4}{*}{Native RAG}
& 10  & 0.7517 & 0.5833 & 0.4280 & 0.3528 & 0.0278 & 0.4287 \\
& 20  & 0.7797 & 0.6667 & 0.4068 & 0.3683 & 0.0741 & 0.4591 \\
& 50  & 0.8323 & 0.6585 & 0.4690 & 0.3341 & 0.1167 & 0.4821 \\
& 100 & 0.8636 & 0.7222 & 0.5106 & 0.3873 & 0.1481 & 0.5264 \\
\midrule
\multirow{4}{*}{HMRAG}
& 10  & 0.7955 & 0.6597 & 0.4513 & 0.2410 & 0.0278 & 0.4350 \\
& 20  & 0.8129 & 0.7153 & 0.5191 & 0.3081 & 0.1111 & 0.4933 \\
& 50  & 0.7972 & 0.7361 & 0.5763 & 0.3442 & 0.1296 & 0.5167 \\
& 100 & 0.7308 & 0.4028 & 0.5657 & 0.3005 & 0.0882 & 0.4176 \\
\midrule
\multirow{4}{*}{VRAG}
& 10  & 0.4336 & 0.4621 & 0.3470 & 0.2651 & 0.0278 & 0.3071 \\
& 20  & 0.4913 & 0.4514 & 0.3919 & 0.1842 & 0.0463 & 0.3130 \\
& 50  & 0.4983 & 0.5069 & 0.3898 & 0.2065 & 0.0000 & 0.3203 \\
& 100 & 0.4825 & 0.4444 & 0.3962 & 0.2203 & 0.0000 & 0.3087 \\
\bottomrule
\end{tabular}
\caption{Detailed retrieval-budget results for the representative baselines used in RQ3.}
\label{tab:appendix_retrieval_budget}
\end{table*}

This subsection provides the complete retrieval-budget results for all evaluated values of $K$, including the overall scores summarized in the main text and the corresponding Recall@$K$/Precision@$K$ statistics. We also include per-task breakdowns where available. 

\begin{figure*}[t]
\centering
\begin{minipage}[t]{0.48\textwidth}
  \centering
  \includegraphics[width=\linewidth]{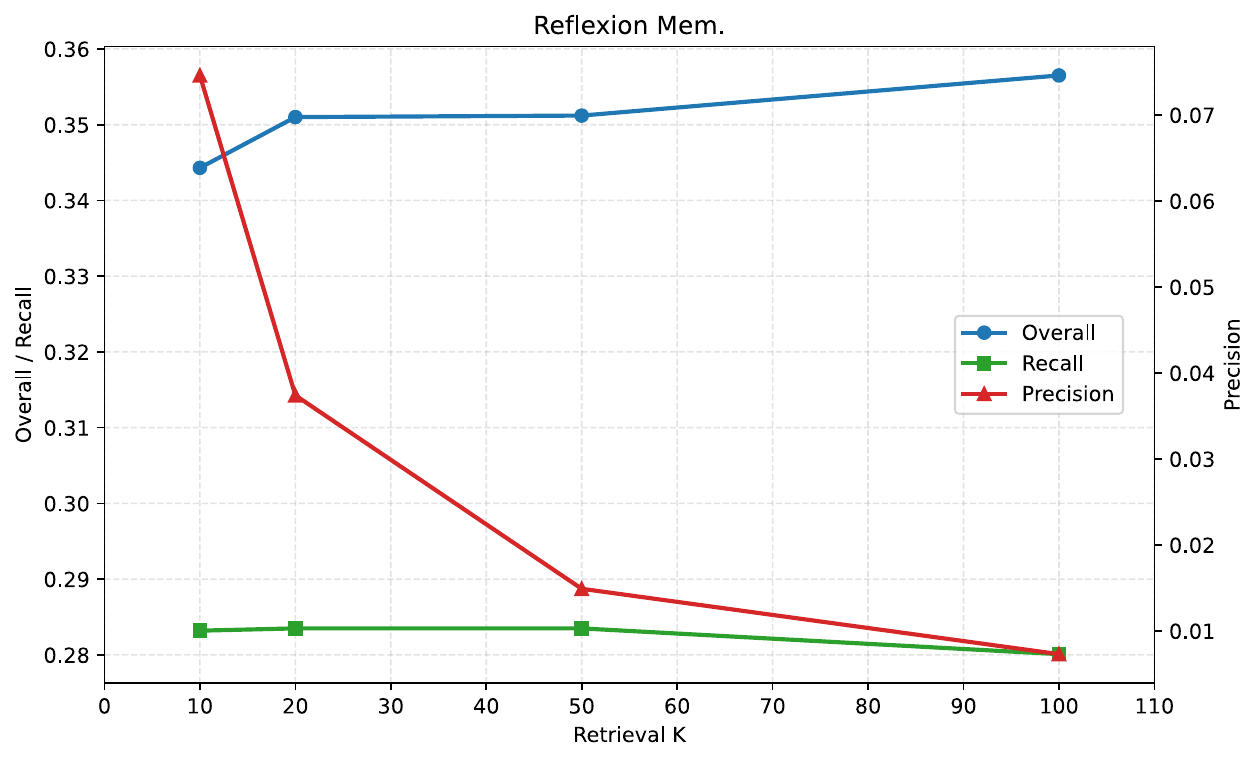}
  \caption{\textsc{Reflexion Mem.}}
  \label{fig:appendix_rq3_reflexion_mem}
\end{minipage}\hfill
\begin{minipage}[t]{0.48\textwidth}
  \centering
  \includegraphics[width=\linewidth]{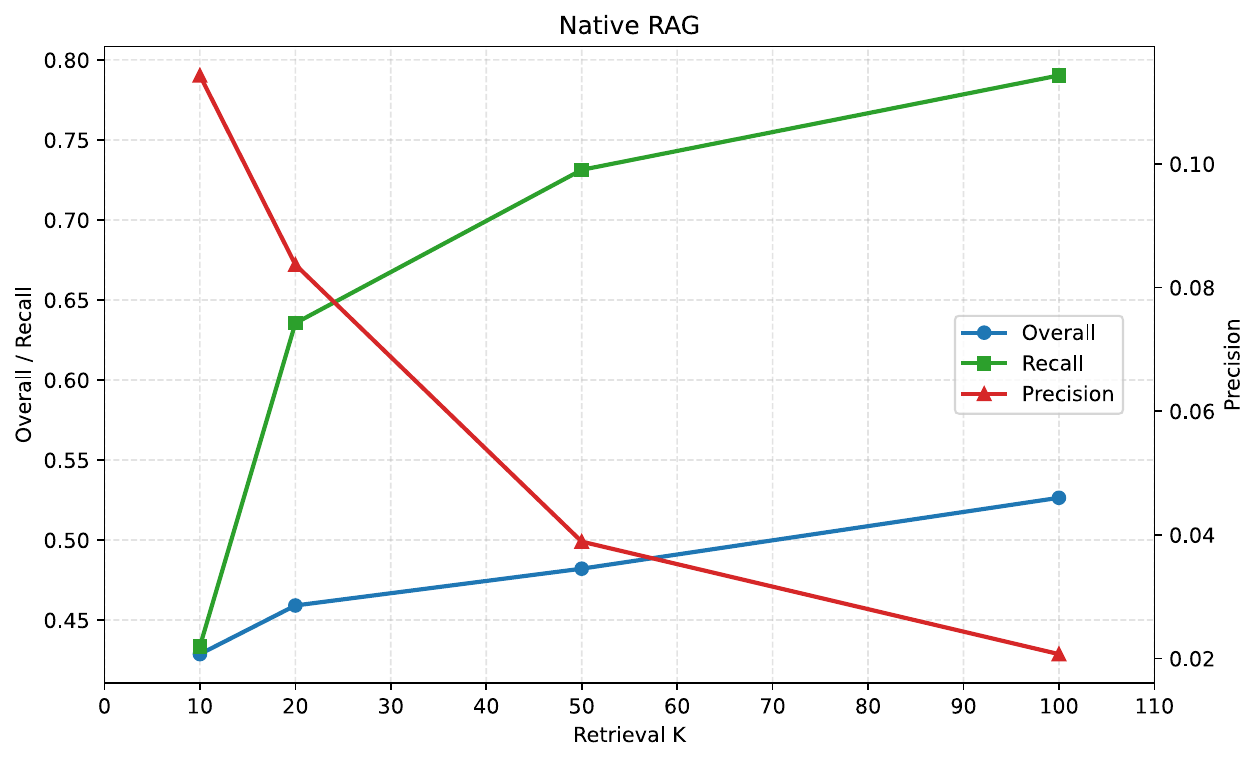}
  \caption{\textsc{Native RAG}}
  \label{fig:appendix_rq3_native_rag}
\end{minipage}

\vspace{0.8em}

\begin{minipage}[t]{0.48\textwidth}
  \centering
  \includegraphics[width=\linewidth]{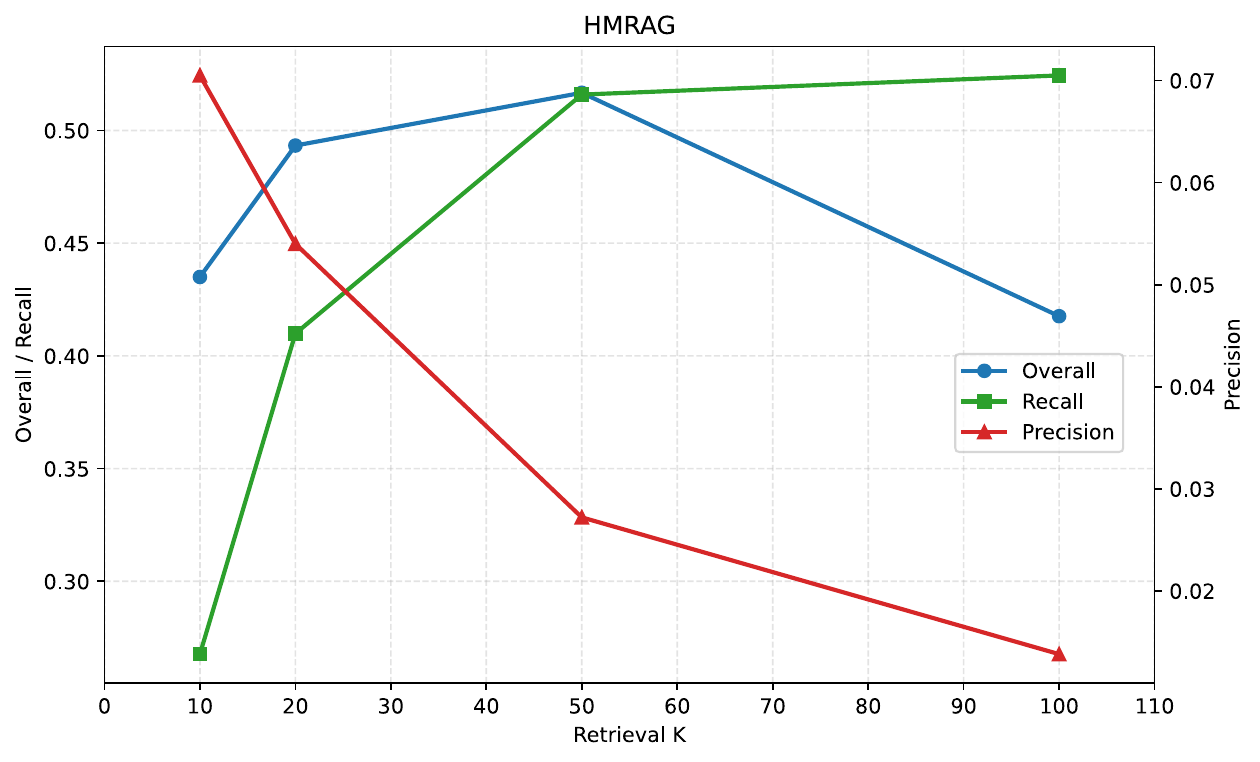}
  \caption{\textsc{HMRAG}}
  \label{fig:appendix_rq3_hmrag}
\end{minipage}\hfill
\begin{minipage}[t]{0.48\textwidth}
  \centering
  \includegraphics[width=\linewidth]{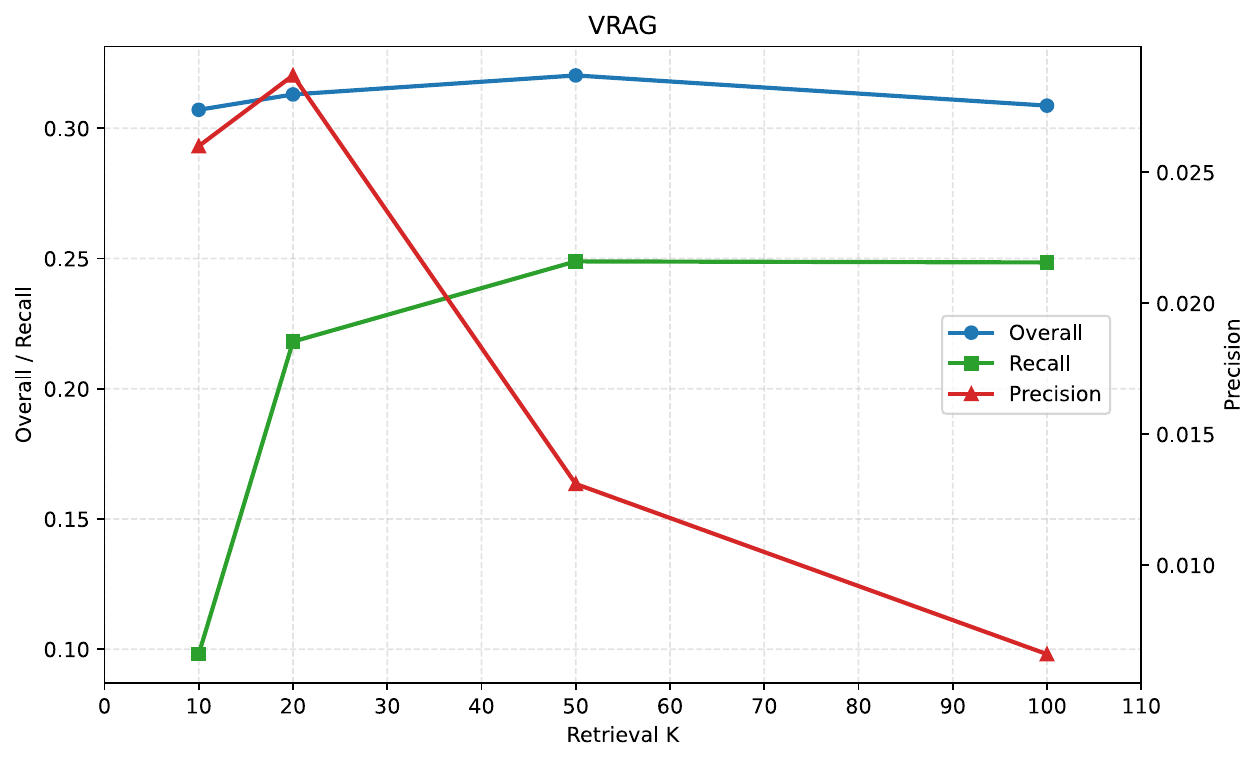}
  \caption{\textsc{VRAG}}
  \label{fig:appendix_rq3_vrag}
\end{minipage}
\end{figure*}

The appendix results further clarify the trade-off discussed in the main text. Across the retrieval-based methods, increasing $K$ generally improves Recall@$K$ while lowering Precision@$K$, but the end-task benefit depends strongly on the method. \textsc{Native RAG} shows the clearest monotonic gain: as $K$ increases, recall rises substantially and the Overall score continues to improve, indicating that this pipeline is able to convert the additional retrieved evidence into useful downstream gains despite the accompanying precision drop.

\textsc{HMRAG} follows a different pattern. Its recall continues to increase with larger $K$, and task-wise scores improve from $K{=}10$ to $K{=}50$ on several categories, especially \textsc{Multi-Hop QA}, \textsc{Conflict Resolution}, and \textsc{Function Calling}. However, performance drops sharply at $K{=}100$, even though recall remains high. This case makes the coverage--noise trade-off especially visible: beyond a moderate retrieval budget, additional evidence appears to introduce enough distractors to outweigh the benefit of higher coverage.

\textsc{VRAG} and \textsc{Reflexion Mem.} show more limited sensitivity to retrieval budget. For \textsc{VRAG}, recall improves from $K{=}10$ to $K{=}50$, but the Overall gain remains modest and reverses slightly at $K{=}100$, suggesting that larger candidate pools do not translate into proportional downstream improvements. \textsc{Reflexion Mem.} changes the least: its recall is nearly flat across budgets, precision steadily decreases, and the Overall score moves only slightly. Taken together, these detailed results reinforce that larger retrieval pools help only when the downstream pipeline can effectively filter and use the extra evidence, rather than merely accumulate more candidates.

\subsection{Modality Experiment}
\label{appendix:modality_experiment}

\begin{figure}[t]
    \centering
    \includegraphics[width=0.5\linewidth]{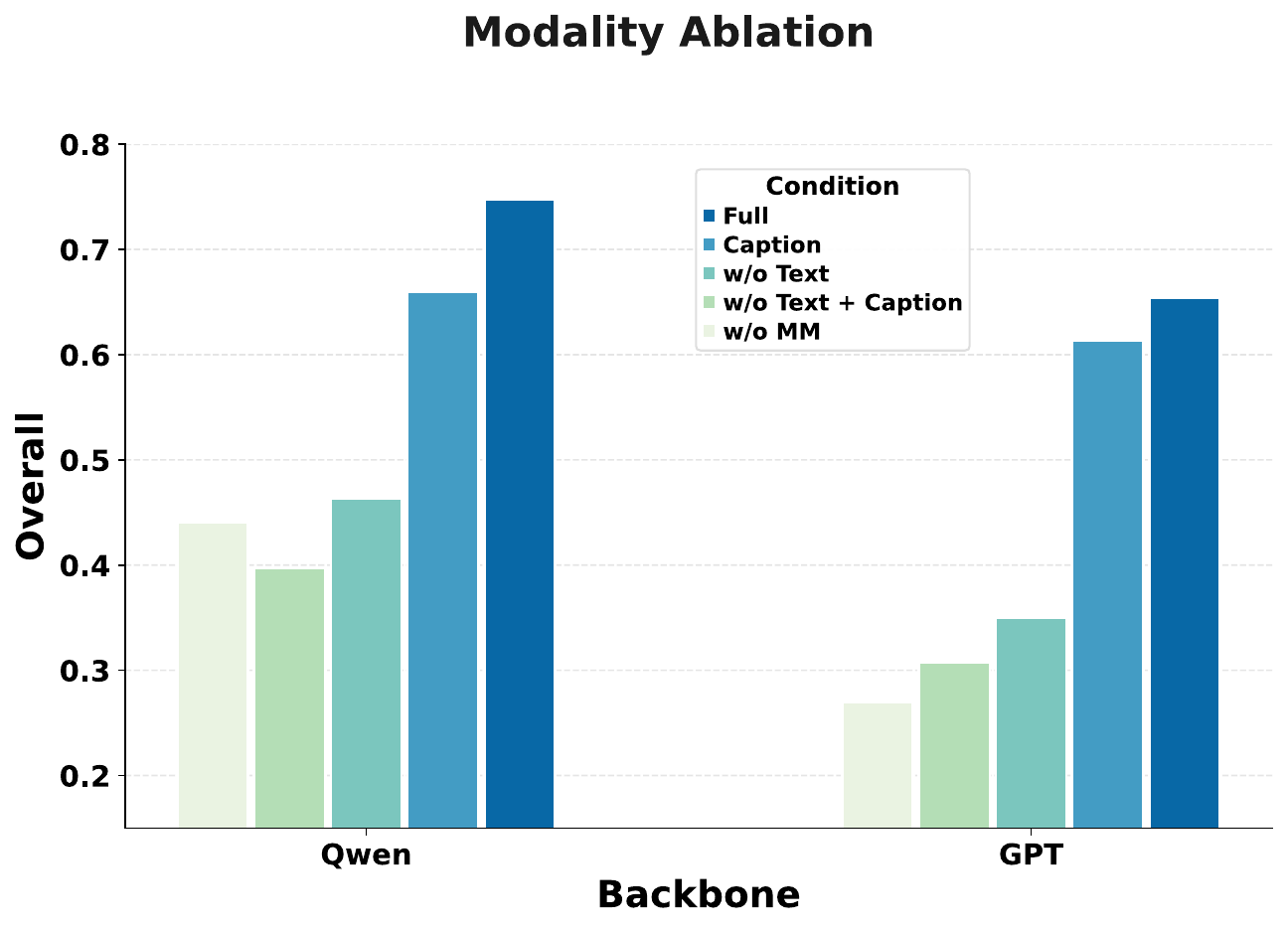}
    \caption{Experiments on different modality ablation settings. `Qwen' for qwen3-vl-235b-instruct, `GPT' for `gpt-4.1'.}
    \label{fig:appendix_modality_experiment}
\end{figure}

We conduct a modality ablation study by directly providing different evidence subsets to the backbone LLM in Figure~\ref{fig:appendix_modality_experiment}. Full uses all available evidence; Caption replaces multimodal evidence with captions while keeping the textual evidence unchanged; w/o text removes textual evidence while preserving native multimodal (non-textual) evidence inputs; w/o MM removes native multimodal evidence while keeping the textual evidence; and w/o Text + Caption removes textual evidence and replaces multimodal evidence with captions.

The clearest result is that textual evidence contributes a large share of benchmark performance. Across both backbones, removing textual evidence leads to the largest drop relative to Full, indicating that many answer-critical cues are carried by conversations, documents, and other text-bearing memory objects. At the same time, multimodal evidence also contributes non-trivially: both w/o MM and Caption remain below Full, showing that replacing or removing raw multimodal content incurs a measurable loss. The comparison between Full and Caption is particularly informative. Caption-based inputs remain much closer to Full than to the text-removed settings, suggesting that the captioning pipeline preserves a large portion of the answer-relevant visual information. However, the remaining gap between Caption and Full shows that captioning is not lossless, and that native multimodal evidence still contains useful information beyond textualized summaries. Additional task-wise breakdowns are provided in the appendix

\subsection{Diagnosis Experiment}
\label{appendix:diagnosis_experiment}
To better understand where current systems fail on \name, we conduct an error-diagnosis experiment on failed predictions from Table~\ref{tab:main_results}. We randomly sample 600 failed cases aggregated across all benchmarked baselines and use GPT-4.1 as an LLM judge to assign each case a single primary failure label. The judge is given the question, model raw response, gold answer, available context, and optional conflict-resolution reference when applicable.

We use a six-way fine-grained taxonomy. \textit{Distributed Evidence Missing} denotes cases where the provided context lacks one or more key facts needed for the gold answer. \textit{Updated Evidence Missing} denotes cases where the gold answer depends on newer, corrected, or authoritative evidence that is absent from the context. \textit{Outdated or Conflicting Evidence Misused} denotes cases where both outdated/conflicting and updated/authoritative evidence are present, but the model follows the wrong one. \textit{Preference Inference Error} denotes cases where the context contains enough cues to infer a preference or convention, but the model fails to infer it correctly. \textit{Function Execution Error} denotes cases where the context is sufficient, but the model fails at the executable output layer, such as using the wrong function, wrong parameters, or malformed output. \textit{Cross-Source Composition Error} denotes cases where the relevant clues are present, but the model fails to combine evidence across sources, modalities, or artifacts.

For higher-level analysis, we further organize these six labels into a two-level hierarchy with three coarse stages: \textit{Access}, \textit{Utilization}, and \textit{Action}. The \textit{Access} stage includes \textit{Distributed Evidence Missing} and \textit{Updated Evidence Missing}, capturing failures to recover the necessary evidence from distributed memory. The \textit{Utilization} stage includes \textit{Cross-Source Composition Error}, \textit{Outdated or Conflicting Evidence Misused}, and \textit{Preference Inference Error}, capturing failures to correctly integrate, prioritize, or interpret already available evidence. The \textit{Action} stage includes \textit{Function Execution Error}, capturing failures in converting remembered information into executable outputs.

Figure~\ref{fig:rq5_failure_analysis} summarizes the diagnosis results. The dominant failure type is \textit{Distributed Evidence Missing}, which accounts for 61.8\% of the diagnosed errors, showing that the main bottleneck still lies in recovering the required evidence from source-distributed memory. The next largest category is \textit{Cross-Source Composition Error} at 16.5\%, indicating that even when some relevant clues are available, systems often fail to combine them correctly across sources. Smaller but still meaningful portions come from \textit{Function Execution Error} (8.2\%), \textit{Outdated or Conflicting Evidence Misused} (6.8\%), and \textit{Updated Evidence Missing} (6.7\%).

Overall, these results suggest that the difficulty of \name is layered. Most failures occur first at the \textit{Access} stage, where systems do not surface the needed distributed evidence at all. A second group of failures then appears at the \textit{Utilization} stage, where models fail to compose clues across sources or prioritize updated evidence over stale or conflicting records. Finally, a smaller but clear portion of failures arises at the \textit{Action} stage, where remembered information is not converted into correct executable outputs. Together, these patterns reinforce that source-distributed multimodal memory is challenging not only because evidence is hard to retrieve, but also because it must be correctly integrated and grounded after retrieval.

\newpage

\setlength{\LTleft}{0pt}
\setlength{\LTright}{0pt}
\begin{longtable}{@{} >{\raggedright\arraybackslash}p{2.6cm} @{\quad} p{\dimexpr\linewidth - 2.6cm - 1em\relax} @{}}
\caption{Illustrative `Conflicting Evidence Misused' case. \color{red}{Red content means misleading or conflicting evidence content}. \color{yellow!75!black}{Golden content means golden evidence or ground truths}. \color{black}In this case, HMRAG recalled both misleading and true evidence, but used misleading one to answer the question, and output wrong answer.}
\label{tab:appendix_conflicting_evidence_misused_case}\\
\hline
\rowcolor{gray!20}%
\textit{Case \& ID} & MMKE\_cc30 from HMRAG \\
\hline
\endfirsthead

\multicolumn{2}{@{}l@{}}{\small\textbf{Table \thetable\ (continued)}}\\
\hline
\rowcolor{gray!20}%
\textit{Case ID} & MMKE\_4723 from HMRAG \\
\hline
\endhead

\hline
\multicolumn{2}{r@{}}{\small\textit{Continued on next page}} \\
\endfoot

\hline
\endlastfoot

  \textbf{Retrieved} &
    \begingroup
    \small
    \setlength{\parskip}{0.35em}%
    \raggedright

    {    
    \noindent {\textbf{Retrieved Items}}\par
    }
    \noindent{$\cdots\cdots\cdots$}\par

    {    
    \noindent \color{yellow!75!black}{Cardamine bulbosa, commonly called bulbous bittercress or fall cress, is a perennial plant in the rose family. It is native to a widespread area of western South America, in both Chile and Argentina. Its natural habitat is dry soils of highland forests and tundras, often in acidic areas.In late summer and early fall, flowers are produced well above the foliage. Its leaves are edible and have a sweet taste.}\par
    }

    \noindent{$\cdots\cdots\cdots$}\par

    {    
    \noindent \color{yellow!75!black}{.../MMKE\_449df6d90ec74289.png}\par
    }

    \noindent{$\cdots\cdots\cdots$}\par

    {    
    \noindent \color{red}{Cardamine bulbosa, commonly called bulbous bittercress or spring cress, is a perennial plant in the mustard family. It is native to a widespread area of eastern North America, in both Canada and the United States. Its natural habitat is moist soils of bottomland forests and swamps, often in calcareous areas. In late spring and early summer, white flowers are produced well above the foliage. Its leaves are edible, and have a peppery taste.}\par
    }
    
    \endgroup \\
  \hline
  \textbf{QA} &
    \small
    Based on Fig. 9e33599e, Flowers of the species shown in the image are produced during which seasons?
    \par                
    \textbf{(A)}\quad Mid autumn and early winter
    \par
    \textbf{(B)}\quad Winter and early spring
    \par
    {\color{yellow!75!black}{\textbf{(C)}\quad Late summer and early fall}}
    \par
    \textbf{(D)}\quad \color{red}{Late spring and early summer}  \\
  \hline
\end{longtable}

\setlength{\LTleft}{0pt}
\setlength{\LTright}{0pt}
\begin{longtable}{@{} >{\raggedright\arraybackslash}p{2.6cm} @{\quad} p{\dimexpr\linewidth - 2.6cm - 1em\relax} @{}}
\caption{Illustrative `Updated Evidence Missing' case. \color{red}{Red content means misleading or conflicting evidence content}. \color{yellow!75!black}{Golden content means ground truths}. \color{black}In this case, HMRAG only recalled misleading evidence, causing wrong answer.}
\label{tab:appendix_updated_evidence_missing_case}\\
\hline
\rowcolor{gray!20}%
\textit{Case \& ID} & MMKE\_4723 from HMRAG \\
\hline
\endfirsthead

\multicolumn{2}{@{}l@{}}{\small\textbf{Table \thetable\ (continued)}}\\
\hline
\rowcolor{gray!20}%
\textit{Case ID} & MMKE\_4723 from HMRAG \\
\hline
\endhead

\hline
\multicolumn{2}{r@{}}{\small\textit{Continued on next page}} \\
\endfoot

\hline
\endlastfoot

  \textbf{Retrieved} &
    \begingroup
    \small
    \setlength{\parskip}{0.35em}%
    \raggedright

    {    
    \noindent {\textbf{Retrieved Items}}\par
    }
    \noindent{$\cdots\cdots\cdots$}\par

    {    
    \noindent \color{yellow!75!black}{.../MMKE\_b9faf19096e44ffc.png}\par
    }

    \noindent{$\cdots\cdots\cdots$}\par

    {    
    \noindent \color{red}{Oxythyrea funesta, known as the \"White spotted rose beetle,\" is a phytophagous beetle from the Cetoniidae family, Cetoniinae subfamily. This beetle is found throughout most of Europe, the eastern Palearctic realm, and the Near East. Larvae feed on plant roots and can stay in the soil until the next spring, growing up to 30 mm long. Adults emerge in early spring, mostly seen from May to July. They are considered pests as they damage floral organs, especially targeting light-colored buds and flowers.These beetles are black, sometimes bronzed, with typically six white spots on the pronotum and several on the elytra. They are covered in white pubescence, but older beetles often lose these hairs over time.}\par
    }
    
    \endgroup \\
  \hline
  \textbf{QA} &
    \small
    Based on Fig. 0ee4a299, What specific plant is associated with the common name of the species shown in the image?
    \par                
    \textbf{(A)}\quad Rose
    \par
    \textbf{(B)}\quad Oak tree
    \par
    {\color{yellow!75!black}{\textbf{(C)}\quad Lavender }}
    \par
    \textbf{(D)}\quad \color{red}{Daisy}  \\
  \hline
\end{longtable}

\newpage
\section{Cases}
\label{appendix:evaluation_samples}
\subsection{Single-Hop QA}
This subsection presents representative \textsc{Single-Hop QA} samples, including the reconstructed memory environment, the gold evidence items, the multiple-choice options, and the gold answer. The examples are selected to illustrate cases where one decisive clue is embedded in a broader multi-source context.

\subsection{Multi-Hop QA}
This subsection provides example \textsc{Multi-Hop QA} instances whose answers require combining evidence distributed across multiple memory objects. We include both the final evaluation form and the source-level evidence annotations to show how reasoning chains are preserved after conversation insertion.

\subsection{Conflict Resolution}
This subsection contains conflict-resolution examples in which stale, contradictory, or superseded memory items coexist in the environment. The examples highlight how the benchmark requires not only retrieval of relevant facts, but also correct prioritization of the updated or authoritative source.
\subsection{Preference Reasoning}
This subsection shows examples for user- and project-preference reasoning. We include both explicit preference cases and implicit preference cases whose signals are distributed across several turns or sources, demonstrating the range of preference representations covered by the benchmark.
\subsection{Function Call}
This subsection provides representative function-calling cases, including the memory context, candidate tool schema, and gold executable output. These examples illustrate why the task is more stringent than multiple-choice QA: the model must recover and assemble the exact action arguments rather than merely identify a correct option.

\newpage
\setlength{\LTleft}{0pt}
\setlength{\LTright}{0pt}
\begin{longtable}{@{} >{\raggedright\arraybackslash}p{2.6cm} @{\quad} p{\dimexpr\linewidth - 2.6cm - 1em\relax} @{}}
\caption{Illustrative \textsc{Single-Hop QA} case. \color{violet}{Violet text indicates scaffold content surrounding the evidence}. \color{yellow!75!black}{Golden content means golden evidence or ground truths}.}
\label{tab:appendix_single_hop_case_example}\\
\hline
\rowcolor{gray!20}%
\textit{Case Source \& ID} & QA\_sample\_5f42a925 from ChartQA\_Pro \\
\hline
\endfirsthead

\multicolumn{2}{@{}l@{}}{\small\textbf{Table \thetable\ (continued)}}\\
\hline
\rowcolor{gray!20}%
\textit{Case Source \& ID} & QA\_sample\_5f42a925 from ChartQA\_Pro \\
\hline
\endhead

\hline
\multicolumn{2}{r@{}}{\small\textit{Continued on next page}} \\
\endfoot

\hline
\endlastfoot

  \textbf{Sources} &
    \begingroup
    \small
    \setlength{\parskip}{0.35em}%
    \raggedright

    \textbf{Group chat:}\quad\texttt{group\_chat\_food\_environment\_lifestyle\_6d738a64}\par\medskip

    \noindent{$\cdots\cdots\cdots$}\par

    {    
    \noindent\textbf{Miya Cruz}\quad\texttt{2023-04-13 11:29:27}\par
    \noindent \color{violet}{Speaking of tourism, it's wild how some cities get totally transformed by visitors....}\par
    }

    {
    \noindent\textbf{Briley Hanson}\quad\texttt{2023-04-13 11:30:25}\par
    \noindent \color{yellow!75!black}{Look at this figure: Fig. 04b3f139}\par
    }
    
    {
    \noindent\textbf{Briley Hanson}\quad\texttt{2023-04-13 11:40:07}\par
    \noindent \includegraphics[width=0.45\linewidth]{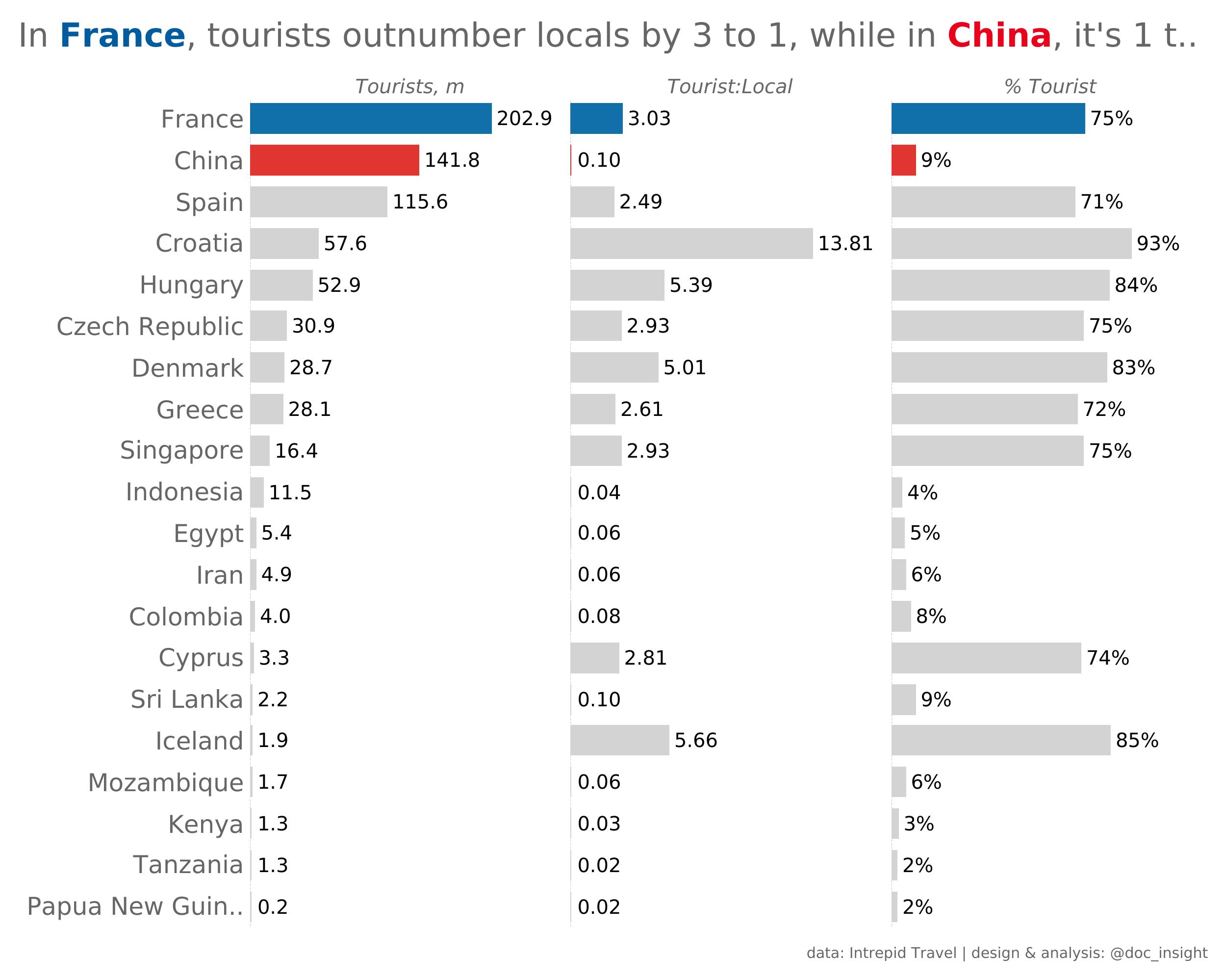}
    }
    
    {
    \noindent\textbf{Ricardo Bruce}\quad\texttt{2023-04-13 11:46:57}\par
    \noindent \color{violet}{Wow, I had no idea Croatia had so many more tourists than locals ....}\par
    }

    \noindent{$\cdots\cdots\cdots$}\par

    \bigskip
    \textbf{Group chat:}\quad\texttt{group\_chat\_films\_transportation\_others\_3bc60e19}\par\medskip

    \noindent $\cdots\cdots$
    
    \noindent\textbf{Axel Hart}\quad\texttt{2023-04-11 09:14:33}\par
    \noindent I think they started with soap, right? I remember reading somewhere that Lever was all about making affordable soap for everyone, and then Unilever just kept expanding into other stuff like food and personal care.\par

    {%
    \noindent\textbf{Miya Cruz}\quad\texttt{2023-04-11 09:25:23}\par
    \noindent \color{violet}Speaking of things that have been around forever, it's kind of like how some tourist destinations have changed so much over time, too.\par
    }%

    {%
    \noindent\textbf{Bridget Deleon}\quad\texttt{2023-04-11 09:25:37}\par
    \noindent \color{violet}Yeah, whether it's brands or places, it's interesting to see how they adapt to stay popular.\par
    }%

    \noindent\textcolor{violet}{$\cdots\cdots\cdots$}\par

    {%
    \noindent\textbf{Guillermo Lynn}\quad\texttt{2023-04-11 09:58:36}\par
    \noindent \color{yellow!75!black}A country is considered to have a high tourism density if the ratio of tourists to locals exceeds 2.8. Additionally, a high tourism influence is defined as having more than 70\% of the population being tourists in the given data.\par
    }%
    {
    \noindent\textbf{Miya Cruz}\quad\texttt{2023-04-11 09:59:06}\par
    \noindent \color{violet}So basically, if a country has way more tourists than locals, it's considered high density, and if most people there are tourists, that's high influence? That's wild---I wonder which countries actually hit those numbers.\par
    }

    \noindent{$\cdots\cdots\cdots$}\par

    \endgroup \\
  \hline
  \textbf{QA} &
    \small
    Based on Fig. 04b3f139, the definition, which country has "high tourism density" but does not have "high tourism influence"?
    \par                
    \textbf{(A)}\quad Czech Republic
    \par
    \textbf{(B)}\quad Denmark
    \par
    {\color{yellow!75!black}{\textbf{(C)}\quad Spain}}
    \par
    \textbf{(D)}\quad Croatia  \\
  \hline
\end{longtable}

\newpage
\setlength{\LTleft}{0pt}
\setlength{\LTright}{0pt}
\begin{longtable}{@{} >{\raggedright\arraybackslash}p{2.6cm} @{\quad} p{\dimexpr\linewidth - 2.6cm - 1em\relax} @{}}
\caption{Illustrative \textsc{Multi-Hop QA} case.\color{violet}{Violet text indicates scaffold content surrounding the evidence}. \color{yellow!75!black}{Golden content means golden evidence or ground truths}.}
\label{tab:appendix_multi_hop_case}\\
\hline
\rowcolor{gray!20}%
\textit{Case Source \& ID} & QA\_sample\_9c56672c from MMCV \\
\hline
\endfirsthead

\multicolumn{2}{@{}l@{}}{\small\textbf{Table \thetable\ (continued)}}\\
\hline
\rowcolor{gray!20}%
\textit{Case Source \& ID} & QA\_sample\_9c56672c from MMCV \\
\hline
\endhead

\hline
\multicolumn{2}{r@{}}{\small\textit{Continued on next page}} \\
\endfoot

\hline
\endlastfoot

  \textbf{Sources} &
    \begingroup
    \small
    \setlength{\parskip}{0.35em}%
    \raggedright

    \textbf{Group chat:}\quad\texttt{group\_chat\_nature\_films\_fashion\_f3c49653}\par\medskip

    \noindent\textbf{Kenya Decker}\quad\texttt{2023-04-05 18:04:00}\par
    \noindent{Yeah, it's kind of like how Coresoft jumped from ...}\par

    \noindent\textbf{Colt Kemp}\quad\texttt{2023-04-05 18:04:03}\par
    \noindent\textcolor{violet}{It's funny, because whether it's travel or game design, being open to surprises seems to make things more memorable.}\par

    \noindent\textbf{Selina Gonzalez}\quad\texttt{2023-04-05 18:22:14}\par
    \noindent\textcolor{yellow!75!black}{Look at this table: Table.~e45b6ac3}\par

    {
    \noindent\textbf{Selina Gonzalez}\quad\texttt{2023-04-05 18:22:14}\par
    \noindent \includegraphics[width=0.6\linewidth]{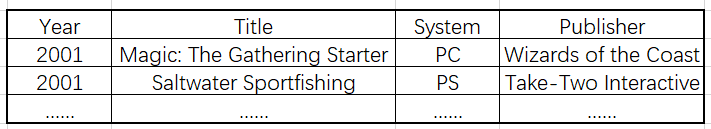}
    }

    \noindent\textbf{Harper Clark}\quad\texttt{2023-04-05 18:27:50}\par
    \noindent\textcolor{violet}{Wow, I had no idea Coresoft made so many different types of games---everything from Magic: The Gathering to fishing and even Cake Mania. That's a pretty wild range.}\par

    \noindent\textcolor{violet}{$\cdots\cdots\cdots$}\par
    
    \bigskip

    \textbf{Group chat:}\quad\texttt{group\_chat\_lifestyle\_food\_nature\_eace8e98}\par\medskip

    \noindent\textbf{Andy Stewart}\quad\texttt{2023-04-03 23:09:41}\par
    \noindent Absolutely---what we're calling `regional trends' are often just the fallout...\par

    \noindent\textbf{Joselyn Moss}\quad\texttt{2023-04-03 23:10:22}\par
    \noindent\textcolor{violet}{Speaking of platforms and what people actually want, it's kind of like how board games get reinvented to fit what's fun for the group, not just what the rules say.}\par

    \noindent\textbf{Andy Stewart}\quad\texttt{2023-04-03 23:18:11}\par
    \noindent\textcolor{yellow!75!black}{Squander (written as ``\$QUANDER'' on the box and in the rules) is an Avalon Hill board game published in 1965. It is based loosely on the game Monopoly, but in reverse. ...}\par

    \noindent\textbf{Harper Clark}\quad\texttt{2023-04-04 00:06:24}\par
    \noindent\textcolor{violet}{That actually sounds hilarious---I love the idea of trying ....}\par

    \noindent{$\cdots\cdots\cdots$}\par

    \bigskip
    \textbf{Group chat:}\quad\texttt{group\_chat\_environment\_animals\_economy\_4b965250}\par\medskip

    \noindent\textbf{Briley Hanson}\quad\texttt{2023-04-06 07:26:54}\par
    \noindent\textcolor{violet}{True, and just like mixing Lego sets, maybe brands need to mix up their strategies for each festival instead of sticking to one formula.}\par

    \noindent\textbf{Maggie Rachael}\quad\texttt{2023-04-06 07:36:05}\par
    \noindent\textcolor{yellow!75!black}{Avalon Hill Games Inc. is a game company that specializes in wargames and strategic board games. ...}\par

    \noindent\textbf{Joselyn Moss}\quad\texttt{2023-04-06 07:39:51}\par
    \noindent\textcolor{violet}{I remember playing some Avalon Hill games with my dad when I was younger---those strategy ones could go on for hours, but they were always a blast.}\par

    \endgroup \\
  \hline
  \textbf{QA} &
    \small
    Based on Table. e45b6ac3, Which company that owns a subsidiary focused on strategic games was the publisher of a video game series in 2004 and created a board game with reversed Monopoly rules in 1965?
    \par
    \textbf{(A)}\quad Parker Brothers
    \par
    {\color{yellow!75!black}{\textbf{(B)}\quad Avalon Hill}}
    \par
    \textbf{(C)}\quad Wizards of the Coast
    \par
    \textbf{(D)}\quad Hasbro \\
  \hline
\end{longtable}

\newpage
\setlength{\LTleft}{0pt}
\setlength{\LTright}{0pt}
\begin{longtable}{@{} >{\raggedright\arraybackslash}p{2.6cm} @{\quad} p{\dimexpr\linewidth - 2.6cm - 1em\relax} @{}}
\caption{Illustrative \textsc{Conflict Resolution} case.\color{violet}{Violet text indicates scaffold content surrounding the evidence}. \color{yellow!75!black}{Golden content means golden evidence or ground truths}.}
\label{tab:appendix_conflict_resolution_case}\\
\hline
\rowcolor{gray!20}%
\textit{Case Source \& ID} & QA\_sample\_1ce15338 from MLLMKC \\
\hline
\endfirsthead

\multicolumn{2}{@{}l@{}}{\small\textbf{Table \thetable\ (continued)}}\\
\hline
\rowcolor{gray!20}%
\textit{Case Source \& ID} & QA\_sample\_1ce15338 from MLLMKC \\
\hline
\endhead

\hline
\multicolumn{2}{r@{}}{\small\textit{Continued on next page}} \\
\endfoot

\hline
\endlastfoot

  \textbf{Sources} &
    \begingroup
    \small
    \setlength{\parskip}{0.35em}%
    \raggedright

    \textbf{Group chat:}\quad\texttt{group\_chat\_films\_art\_and\_design\_music\_dccdfbfa}\par\medskip

    \noindent\textbf{Immanuel Goodwin}\quad\texttt{2023-04-04 19:03:41}\par
    \noindent\textcolor{violet}{That reminds me---have you noticed how some athletes transition into acting and bring a whole new energy to the screen?....}\par

    \noindent\textbf{Kayden Soto}\quad\textcolor{yellow!75!black}{\texttt{2023-04-04 19:09:59}}\par
    \noindent\textcolor{yellow!75!black}{John Cena was born on April 23, 1977. John Cena is Canadian. John Cena is a professional wrestler and actor.}\par

    \noindent\textbf{George Villegas}\quad\texttt{2023-04-04 19:28:09}\par
    \noindent\textcolor{violet}{I always forget that John Cena is Canadian! It's wild how he's managed to balance wrestling and acting since he was born in 1977.}\par

    \noindent{$\cdots\cdots\cdots$}\par

    \bigskip
    \textbf{Group chat:}\quad\texttt{group\_chat\_religion\_sports\_animals\_e78b964a}\par\medskip

    \noindent\textbf{George Villegas}\quad\texttt{2023-04-04 14:42:03}\par
    \noindent\textcolor{violet}{Speaking of merch and visual identity, have you all noticed how wrestling belts and shirts have become iconic symbols too?}\par

    \noindent\textbf{Kane Owen}\quad\texttt{2023-04-04 14:46:31}\par
    \noindent\textcolor{yellow!75!black}{Look at this figure: Fig. 0fbc5062}\par

    \noindent\textbf{Kane Owen}\quad\texttt{2023-04-04 14:47:11}\par
    \noindent\includegraphics[width=0.45\linewidth]{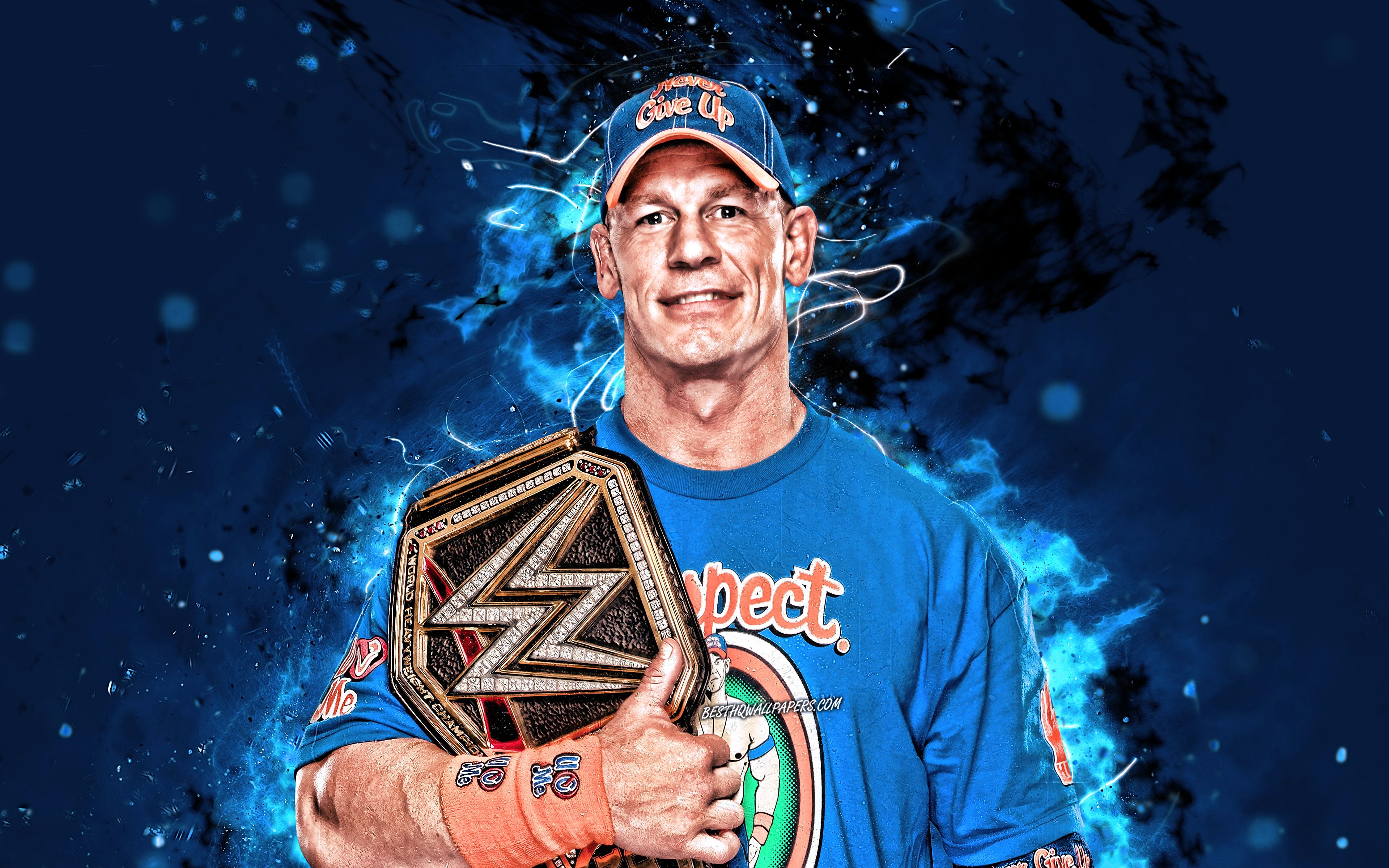}\par

    \noindent\textbf{Maggie Rachael}\quad\texttt{2023-04-04 14:48:20}\par
    \noindent\textcolor{violet}{That championship belt looks awesome! I love how intense the background is---it really makes the whole image pop.}\par

    \noindent{$\cdots\cdots\cdots$}\par

    \bigskip
    \textbf{Group chat:}\quad\texttt{group\_chat\_architecture\_history\_technology\_4856238e}\par\medskip

    \noindent\textbf{Alan Woods}\quad\texttt{2023-04-05 12:37:49}\par
    \noindent\textcolor{violet}{Yeah, it's wild how some people start in one world---like sports or activism---and then totally pivot to something unexpected, but still keep their influence.}\par

    \noindent{$\cdots\cdots\cdots$}\par

    \noindent\textbf{Zander Aguilar}\quad\textcolor{yellow!75!black}{\texttt{2023-04-05 12:59:37}}\par
    \noindent\textcolor{yellow!75!black}{John Cena was born on April 23, 1977. John Cena is American. John Cena is a professional wrestler and actor.}\par

    \noindent{$\cdots\cdots\cdots$}\par

    \endgroup \\
  \hline
  \textbf{QA} &
    \small
    Based on Fig. 0fbc5062, what is the nationality of the person in the picture?
    \par
    {\color{yellow!75!black}{\textbf{(A)}\quad American}}
    \par
    \textbf{(B)}\quad Canadian
    \par
    \textbf{(C)}\quad Panamanian
    \par
    \textbf{(D)}\quad Mexican \\
  \hline
\end{longtable}

\newpage
\setlength{\LTleft}{0pt}
\setlength{\LTright}{0pt}
\begin{longtable}{@{} >{\raggedright\arraybackslash}p{2.6cm} @{\quad} p{\dimexpr\linewidth - 2.6cm - 1em\relax} @{}}
\caption{Illustrative \textsc{Preference Reasoning} case.\color{violet}{Violet text indicates scaffold content surrounding the evidence}. \color{yellow!75!black}{Golden content means golden evidence or ground truths}.}
\label{tab:appendix_preference_disney_case}\\
\hline
\rowcolor{gray!20}%
\textit{Case Source \& ID} & QA\_sample\_fbaa3802\_1 from MMPB \\
\hline
\endfirsthead

\multicolumn{2}{@{}l@{}}{\small\textbf{Table \thetable\ (continued)}}\\
\hline
\rowcolor{gray!20}%
\textit{Case Source \& ID} & QA\_sample\_fbaa3802\_1 from MMPB \\
\hline
\endhead

\hline
\multicolumn{2}{r@{}}{\small\textit{Continued on next page}} \\
\endfoot

\hline
\endlastfoot

  \textbf{Sources} &
    \begingroup
    \small
    \setlength{\parskip}{0.35em}%
    \raggedright

    \textbf{Group chat:}\quad\texttt{group\_chat\_sports\_music\_literature\_dc339f8b}\par\medskip

    \noindent\textbf{Bridget Deleon}\quad\texttt{2023-04-04 08:04:32}\par
    \noindent\textcolor{violet}{That's a good point, Kayden. I've noticed a lot of crime dramas lately really get into the psychology behind systems and decisions---kind of like what we're talking about here.}\par

    \noindent\textcolor{yellow!75!black}{\textbf{Joselyn Moss}}\quad\texttt{2023-04-04 08:06:18}\par
    \noindent\textcolor{yellow!75!black}{In terms of entertainment, I enjoys crime thrillers, historical dramas, but dislikes sports TV, romantic dramas}\par

    \noindent\textbf{Bridget Deleon}\quad\texttt{2023-04-04 08:06:19}\par
    \noindent\textcolor{violet}{Have you seen ``Mindhunter''? It's a crime thriller with a bit of history mixed in, might be right up your alley.}\par
    
    \noindent{$\cdots\cdots\cdots$}\par

    \bigskip
    \textbf{Group chat:}\quad\texttt{group\_chat\_music\_politics\_animals\_74b7030e}\par\medskip

    \noindent\textbf{Bridget Deleon}\quad\texttt{2023-04-04 06:09:59}\par
    \noindent\textcolor{violet}{Speaking of surprising roles, I feel like streaming platforms have really changed how we discover actors in new lights.}\par

    \noindent{$\cdots\cdots\cdots$}\par

    \noindent\textbf{Bridget Deleon}\quad\texttt{2023-04-04 06:20:21}\par
    \noindent\textcolor{yellow!75!black}{Look at this figure: Fig. d760cff8}\par

    \noindent\textbf{Bridget Deleon}\quad\texttt{2023-04-04 06:21:23}\par
    \noindent\includegraphics[width=0.45\linewidth]{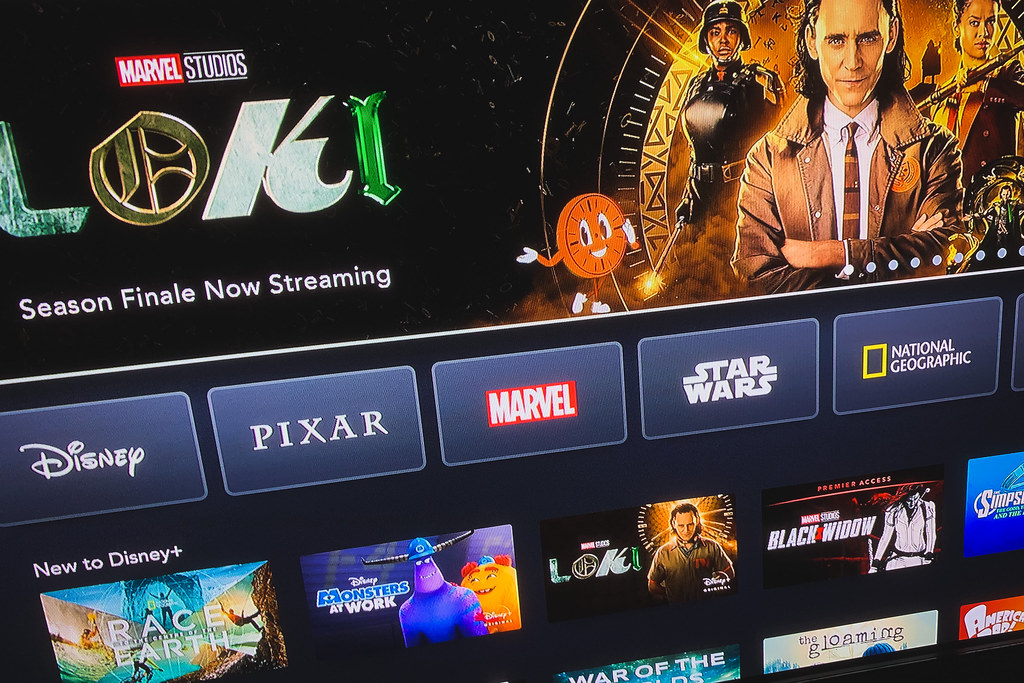}\par

    \noindent\textbf{Miles Roberts}\quad\texttt{2023-04-04 06:21:44}\par
    \noindent\textcolor{violet}{I just finished watching the Loki season finale and it was wild! Disney+ has so much good stuff lately, I might check out Black Widow next.}\par

    \noindent{$\cdots\cdots\cdots$}\par

    \endgroup \\
  \hline
  \textbf{QA} &
    \small
    Based on Fig. d760cff8, among the activities that could reasonably occur in the given image, which one is Joselyn Moss least likely to be doing?
    \par
    \textbf{(A)}\quad Watching movie marathons
    \par
    {\color{yellow!75!black}{\textbf{(B)}\quad Watching sports TV}}
    \par
    \textbf{(C)}\quad Practicing meditation
    \par
    \textbf{(D)}\quad Streaming the latest series \\
  \hline
\end{longtable}

\newpage
\setlength{\LTleft}{0pt}
\setlength{\LTright}{0pt}
\begin{longtable}{@{} >{\raggedright\arraybackslash}p{2.6cm} @{\quad} p{\dimexpr\linewidth - 2.6cm - 1em\relax} @{}}
\caption{Illustrative \textsc{Function Call} case.\color{violet}{Violet text indicates scaffold content surrounding the evidence}. \color{yellow!75!black}{Golden content means golden evidence or ground truths}.}
\label{tab:appendix_function_call_hydrocortisone_case}\\
\hline
\rowcolor{gray!20}%
\textit{Case Source \& ID} & FC\_sample\_c29ba5bd from M3\_Bench \\
\hline
\endfirsthead

\multicolumn{2}{@{}l@{}}{\small\textbf{Table \thetable\ (continued)}}\\
\hline
\rowcolor{gray!20}%
\textit{Case Source \& ID} & FC\_sample\_c29ba5bd from M3\_Bench \\
\hline
\endhead

\hline
\multicolumn{2}{r@{}}{\small\textit{Continued on next page}} \\
\endfoot

\hline
\endlastfoot

  \textbf{Sources} &
    \begingroup
    \small
    \setlength{\parskip}{0.35em}%
    \raggedright

    \textbf{Group chat:}\quad\texttt{group\_chat\_business\_nature\_health\_9981760b}\par\medskip

    \noindent\textbf{Kara Yates}\quad\texttt{2023-04-21 19:32:23}\par
    \noindent\textcolor{violet}{You know, all this talk about shelf placement and visual cues got me thinking---what if we had a digital tool that did the same thing? Like, before you even walk into the store?}\par

    \noindent\textbf{Linda Anderson}\quad\texttt{2023-04-21 19:48:59}\par
    \noindent\textcolor{violet}{I've used those kinds of apps before for supplements, but never for actual meds. If it pulled straight from FDA data, I'd trust it way more than random Google results.}\par

    \noindent\textbf{Kayden Soto}\quad\texttt{2023-04-21 19:49:06}\par
    \noindent\textcolor{yellow!75!black}{FDA drug lookup and comparison. This workflow begins by retrieving official drug information for a specific medication....}\par
    
    \noindent\textbf{Guillermo Lynn}\quad\texttt{2023-04-21 19:49:13}\par
    \noindent\textcolor{violet}{That sounds super useful, especially if you're trying...}\par
    
    \noindent{$\cdots\cdots\cdots$}\par
    
    \bigskip
    \textbf{Group chat:}\quad\texttt{group\_chat\_religion\_sports\_television\_0e949769}\par\medskip

    \noindent\textbf{Amiah Sweeney}\quad\texttt{2023-04-22 23:52:27}\par
    \noindent\textcolor{violet}{Alright, so we're set on ``The Usual Suspects'' for tonight. Can't wait to see if it holds up!}\par

    \noindent\textbf{Justice Clark}\quad\texttt{2023-04-22 23:56:07}\par
    \noindent\textcolor{violet}{Yeah, I'm looking forward to it. But before we get too comfy, anyone else always end up bringing random essentials to movie nights?}\par

    \noindent\textbf{Colt Kemp}\quad\texttt{2023-04-23 00:02:55}\par
    \noindent\textcolor{violet}{Haha, I do! I usually have snacks and a mini first aid kit in my bag, just in case. You never know what'll happen.}\par

    \noindent\textbf{Asia Rivers}\quad\texttt{2023-04-23 00:15:59}\par
    \noindent\textcolor{yellow!75!black}{(Image; see Fig. 8dc8ab97.)}\par
    \noindent\includegraphics[width=0.45\linewidth]{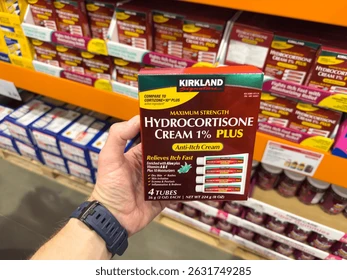}\par

    \noindent\textbf{Immanuel Goodwin}\quad\texttt{2023-04-23 00:34:02}\par
    \noindent\textcolor{violet}{I've actually used that cream before and it works pretty well for bug bites and rashes. It's nice that you get multiple tubes in one pack too.}\par
    
    \noindent{$\cdots\cdots\cdots$}\par

    \bigskip
    \textbf{Candidate Tools}\par
    \noindent\textcolor{yellow!75!black}{
    \{"name": "HEALTHCARE\_MCP\_drug\_lookup", "description": "Look up FDA drug information by name.", "args": \{"drug\_name": "str"\}\}
    }\par
    \noindent{$\cdots\cdots\cdots$}\par
    \endgroup \\
  \hline
  \textbf{QA} &
    \small
    Based on Fig. 8dc8ab97, what information is available about the medication shown in this image? Invoke functions to solve this questions.
    \par
    \textcolor{yellow!75!black}{
    \texttt{HEALTHCARE\_MCP\_drug\_lookup("drug\_name"="hydrocortisone cream 1\%")}}
    \\
  \hline
\end{longtable}


\newpage
\section{Prompts}
\label{appendix:prompts}
This section collects the main prompt templates used in construction and evaluation. It includes prompts for evidence normalization, question rewriting, distractor generation, evidence verification, conversation generation, memory insertion, and final evaluation. We separate prompts by stage so that future users of the benchmark can reproduce or adapt individual components without reusing the full pipeline verbatim.

\newpage
\subsection{Correctness Verification}
\label{appendix:prompt_correctness_verification}
\begingroup
\setlength{\fboxsep}{9pt}%
\setlength{\fboxrule}{0.65pt}%
\begin{figure}[h]
\noindent\begin{minipage}{\linewidth}
\noindent\fcolorbox{gray!65}{gray!65}{\parbox{\dimexpr\linewidth-2\fboxsep-2\fboxrule}{\centering\textbf{\textcolor{white}{Prompt for Correctness Verification}}}}\\[-0.5pt]
\noindent\fcolorbox{gray!50}{gray!10}{\parbox{\dimexpr\linewidth-2\fboxsep-2\fboxrule}{%
\footnotesize\raggedright
\textbf{\#\#\# Task}\par\vspace{0.35em}
You are a multimodal answer verifier. You will be given images, pieces of text, a question, and a proposed answer. Carefully check whether the proposed answer is correct given all the information from both the images and the texts.\par\vspace{0.35em}
If the answer is completely correct, output YES. If the answer is partially correct, incorrect, incomplete, or contradicted by the text or the image, output NO.\par\vspace{0.35em}
\textbf{\#\#\# Output Format}\par\vspace{0.35em}
Output exactly one word: YES or NO.\par
\vspace{0.2em}}}
\end{minipage}\par\vspace{0.65em}
\caption{Prompt template for correctness verification.}
\label{fig:prompt-correctness-verification}
\end{figure}
\endgroup

\subsection{Evidence Necessity Verification}
\label{appendix:prompt_evidence_necessity_verification}
\begin{figure}[h]
\begingroup
\setlength{\fboxsep}{9pt}%
\setlength{\fboxrule}{0.65pt}%
\noindent\begin{minipage}{\linewidth}
\noindent\fcolorbox{gray!65}{gray!65}{\parbox{\dimexpr\linewidth-2\fboxsep-2\fboxrule}{\centering\textbf{\textcolor{white}{Prompt for Evidence Necessity Verification}}}}\\[-0.5pt]
\noindent\fcolorbox{gray!50}{gray!10}{\parbox{\dimexpr\linewidth-2\fboxsep-2\fboxrule}{%
\footnotesize\raggedright
\textbf{\#\#\# Task}\par\vspace{0.35em}
You are an assistant tasked with verifying whether a given question can be correctly answered using the provided evidence. You will be given a question-answer pair and a set of evidence items, which may include text, images, and tables. One potentially relevant piece of evidence is missing.\par\vspace{0.35em}
\textbf{\#\#\# Important Rules}\par\vspace{0.35em}
1. Judge only based on the provided evidence.\par
2. Do not rely on background knowledge or external reasoning.\par
3. If the correct answer cannot be uniquely determined, output No.\par
4. If the evidence is sufficient to correctly and unambiguously support the answer, output Yes.\par\vspace{0.35em}
\textbf{\#\#\# Output Format}\par\vspace{0.35em}
Output only one word: Yes or No.\par
\vspace{0.2em}}}
\end{minipage}\par\vspace{0.65em}
\captionof{figure}{Prompt template for evidence necessity verification.}
\label{fig:prompt-evidence-necessity-verification}
\endgroup
\end{figure}

\subsection{Distractor Generation}
\label{appendix:prompt_distractor_generation}
\begin{figure}[h]
\begingroup
\setlength{\fboxsep}{9pt}%
\setlength{\fboxrule}{0.65pt}%
\noindent\begin{minipage}{\linewidth}
\noindent\fcolorbox{gray!65}{gray!65}{\parbox{\dimexpr\linewidth-2\fboxsep-2\fboxrule}{\centering\textbf{\textcolor{white}{Prompt for Distractor Generation}}}}\\[-0.5pt]
\noindent\fcolorbox{gray!50}{gray!10}{\parbox{\dimexpr\linewidth-2\fboxsep-2\fboxrule}{%
\footnotesize\raggedright
\textbf{\#\#\# Task}\par\vspace{0.35em}
You are an assistant tasked with creating distractor options for a multiple-choice question. You will be given images, pieces of text, a question, and its correct answer. Generate three distractor options that are related to the question but incorrect.\par\vspace{0.35em}
\textbf{\#\#\# Additional Instructions}\par\vspace{0.35em}
1. The distractors should be plausible enough to be considered potential answers, but not the correct one.\par
2. The distractors should have a similar format and length to the original answer, but be clearly different.\par
3. The distractors should not be trivial or too far-fetched.\par\vspace{0.35em}
\textbf{\#\#\# Output Format}\par\vspace{0.35em}
Output one distractor option per line. Do not provide explanations or extra text.\par
\vspace{0.2em}}}
\end{minipage}\par\vspace{0.65em}
\captionof{figure}{Prompt template for distractor generation.}
\label{fig:prompt-distractor-generation}
\endgroup
\end{figure}

\newpage
\subsection{Metadata and Caption Generation}
\label{appendix:prompt_metadata_caption_generation}
\begin{figure}[h]
\begingroup
\setlength{\fboxsep}{9pt}%
\setlength{\fboxrule}{0.65pt}%
\noindent\begin{minipage}{\linewidth}
\noindent\fcolorbox{gray!65}{gray!65}{\parbox{\dimexpr\linewidth-2\fboxsep-2\fboxrule}{\centering\textbf{\textcolor{white}{Prompt for Metadata and Caption Generation}}}}\\[-0.5pt]
\noindent\fcolorbox{gray!50}{gray!10}{\parbox{\dimexpr\linewidth-2\fboxsep-2\fboxrule}{%
\footnotesize\raggedright
\textbf{\#\#\# Task}\par\vspace{0.35em}
You are an assistant tasked with topic classification for multimodal content. You will be given a question-answer pair, evidence items including text, images, and tables, along with a list of candidate topics.\par\vspace{0.35em}
Select one or more topics from the candidate list that best match the overall content and intent. If none of the candidate topics are suitable, output \texttt{others} and additionally provide a short topic description.\par\vspace{0.35em}
\textbf{\#\#\# Candidate Topics}\par\vspace{0.35em}
films,science,politics,sports,video games,transportation,television,music,animals,history,literature,architecture,art and design,fashion,food,health,lifestyle,nature,religion,travel,business,education,environment,government,economy,technology.\par\vspace{0.35em}
\textbf{\#\#\# Output Rules}\par\vspace{0.35em}
1. Output all selected topics on a single line, separated by commas.\par
2. If using \texttt{others}, it must appear at the beginning of the output.\par
3. Do not output explanations or extra text.\par
\vspace{0.2em}}}
\end{minipage}\par\vspace{0.65em}
\caption{Prompt template for metadata and caption generation.}
\label{fig:prompt-metadata-caption-generation}
\endgroup
\end{figure}

\subsection{Conversation Theme Planning}
\label{appendix:prompt_conversation_theme_planning}
\begin{figure}[h]
\begingroup
\setlength{\fboxsep}{9pt}%
\setlength{\fboxrule}{0.65pt}%
\noindent\begin{minipage}{\linewidth}
\noindent\fcolorbox{gray!65}{gray!65}{\parbox{\dimexpr\linewidth-2\fboxsep-2\fboxrule}{\centering\textbf{\textcolor{white}{Prompt for Conversation Theme Planning}}}}\\[-0.5pt]
\noindent\fcolorbox{gray!50}{gray!10}{\parbox{\dimexpr\linewidth-2\fboxsep-2\fboxrule}{%
\footnotesize\raggedright
\textbf{\#\#\# Task}\par\vspace{0.35em}
You are a group chat topic control agent. Your job is to help guide and control the flow of the group chat conversation based on the provided inputs. You will be given the potential QA including multimodal content such as text, images, and tables.\par\vspace{0.35em}
\textbf{\#\#\# Important Instructions}\par\vspace{0.35em}
1. Encourage natural use of the multimodal content, but do not explicitly mention it as a gold clue.\par
2. Phrase sub-topics mainly as declarative sentences.\par
3. Encourage productive discussion, decision making, and problem solving.\par
4. Keep the group on track and avoid derailment.\par
5. Do not mention the question-answer pair directly.\par\vspace{0.35em}
\textbf{\#\#\# Output Format}\par\vspace{0.35em}
Generate 10--20 sub-topics, one per line, mainly as declarative sentences.\par
\vspace{0.2em}}}
\end{minipage}\par\vspace{0.65em}
\captionof{figure}{Prompt template for conversation theme planning.}
\label{fig:prompt-conversation-theme-planning}
\endgroup
\end{figure}

\newpage
\subsection{Turn-Level Conversation Generation}
\label{appendix:prompt_turn_level_conversation_generation}
\begin{figure}[h]
\begingroup
\setlength{\fboxsep}{9pt}%
\setlength{\fboxrule}{0.65pt}%
\noindent\begin{minipage}{\linewidth}
\noindent\fcolorbox{gray!65}{gray!65}{\parbox{\dimexpr\linewidth-2\fboxsep-2\fboxrule}{\centering\textbf{\textcolor{white}{Prompt for Turn-Level Conversation Generation}}}}\\[-0.5pt]
\noindent\fcolorbox{gray!50}{gray!10}{\parbox{\dimexpr\linewidth-2\fboxsep-2\fboxrule}{%
\footnotesize\raggedright
\textbf{\#\#\# Task}\par\vspace{0.35em}
You are an assistant tasked with replying to a group chat.\par
You will be given:\par
1. A series of messages from the group chat.\par
2. A specific topic to be aware of when replying.\par
3. Your personal preferences.\par\vspace{0.35em}
\textbf{\#\#\# Step 1}\par\vspace{0.35em}
Internally decide whether the reply should be SHORT (1--2 sentences) or LONG (5--8 sentences) based on the conversation length.\par\vspace{0.35em}
\textbf{\#\#\# Step 2}\par\vspace{0.35em}
Generate the reply accordingly.\par\vspace{0.35em}
\textbf{\#\#\# Requirements}\par\vspace{0.35em}
- Respond relevantly to the ongoing conversation.\par
- Break echo chambers by introducing new perspectives when needed.\par
- Avoid repetition and mere agreement.\par
- Keep the reply conversational and casual.\par
- Stay consistent with the provided preferences.\par\vspace{0.35em}
\textbf{\#\#\# Output Format}\par\vspace{0.35em}
Output a single, natural group-chat message and nothing else.\par
\vspace{0.2em}}}
\end{minipage}\par\vspace{0.65em}
\captionof{figure}{Prompt template for turn-level conversation generation.}
\label{fig:prompt-turn-level-conversation-generation}
\endgroup
\end{figure}

\subsection{Final Evaluation Prompt}
\label{appendix:prompt_final_evaluation}
\begin{figure}
\begingroup
\setlength{\fboxsep}{9pt}%
\setlength{\fboxrule}{0.65pt}%
\noindent\begin{minipage}{\linewidth}
\noindent\fcolorbox{gray!65}{gray!65}{\parbox{\dimexpr\linewidth-2\fboxsep-2\fboxrule}{\centering\textbf{\textcolor{white}{Prompt for Final Evaluation}}}}\\[-0.5pt]
\noindent\fcolorbox{gray!50}{gray!10}{\parbox{\dimexpr\linewidth-2\fboxsep-2\fboxrule}{%
\footnotesize\raggedright
\textbf{\#\#\# Task}\par\vspace{0.35em}
You are a long-context multimodal assistant. You will be given a conversation history that may include a large amount of text and images. Your task is to carefully read the entire provided conversation, understand the user's question, and answer it as accurately as possible.\par\vspace{0.35em}
\textbf{\#\#\# Specific Instructions}\par\vspace{0.35em}
1. Use only the information available in the conversation and images.\par
2. Pay attention to earlier parts of the conversation if they contain necessary definitions, assumptions, or details.\par
3. If the question cannot be answered from the given context and images alone, do not guess.\par
4. Some runs may provide only captions or descriptions instead of the original multimodal evidence. If that information is insufficient, answer cautiously.\par\vspace{0.35em}
\textbf{\#\#\# Output Format}\par\vspace{0.35em}
Output exactly one choice from \texttt{(A)}, \texttt{(B)}, \texttt{(C)}, or \texttt{(D)}, and nothing else.\par
If the question cannot be answered from the available information, output:\par
\texttt{No, I can not answer this question based on the available information}\par
\vspace{0.2em}}}
\end{minipage}\par\vspace{0.65em}
\captionof{figure}{Prompt template for final evaluation.}
\label{fig:prompt-final-evaluation}
\endgroup
\end{figure}

\newpage
\subsection{Insertion Anchor Selection}
\label{appendix:prompt_insertion_anchor_selection}
\begin{figure}[h]
\begingroup
\setlength{\fboxsep}{9pt}%
\setlength{\fboxrule}{0.65pt}%
\noindent\begin{minipage}{\linewidth}
\noindent\fcolorbox{gray!65}{gray!65}{\parbox{\dimexpr\linewidth-2\fboxsep-2\fboxrule}{\centering\textbf{\textcolor{white}{Prompt for Insertion Anchor Selection}}}}\\[-0.5pt]
\noindent\fcolorbox{gray!50}{gray!10}{\parbox{\dimexpr\linewidth-2\fboxsep-2\fboxrule}{%
\footnotesize\raggedright
\textbf{\#\#\# Task}\par\vspace{0.35em}
You are an expert at analyzing group chat conversations. Your task is to determine the best turn to insert a given scaffold conversation segment into an existing conversation.\par\vspace{0.35em}
You will see:\par
- A conversation window with turns labeled by index.\par
- A scaffold conversation segment that needs to be inserted.\par\vspace{0.35em}
Determine which turn index is the most suitable insertion point.\par\vspace{0.35em}
\textbf{\#\#\# Consider}\par\vspace{0.35em}
1. Topic relevance.\par
2. Natural flow.\par
3. Timing.\par\vspace{0.35em}
\textbf{\#\#\# Output Format}\par\vspace{0.35em}
Output only a single integer, or \texttt{NONE} if no position is suitable.\par
\vspace{0.2em}}}
\end{minipage}\par\vspace{0.65em}
\captionof{figure}{Prompt template for insertion anchor selection.}
\label{fig:prompt-insertion-anchor-selection}
\endgroup
\end{figure}

\subsection{Scaffold and Smoothing}
\label{appendix:prompt_scaffold_smoothing}

\begin{figure}[h]
\begingroup
\setlength{\fboxsep}{9pt}%
\setlength{\fboxrule}{0.65pt}%
\noindent\begin{minipage}{\linewidth}
\noindent\fcolorbox{gray!65}{gray!65}{\parbox{\dimexpr\linewidth-2\fboxsep-2\fboxrule}{\centering\textbf{\textcolor{white}{Prompt for Scaffold and Smoothing}}}}\\[-0.5pt]
\noindent\fcolorbox{gray!50}{gray!10}{\parbox{\dimexpr\linewidth-2\fboxsep-2\fboxrule}{%
\footnotesize\raggedright
You are a conversational conversation generator for group chats.\par\vspace{0.35em}

The user will provide:\par\vspace{0.35em}
1. A central topic, which may be a piece of text, an image, or a table.\par\vspace{0.35em}
2. A conversation history in "SpeakerName: message" format, involving multiple speakers.\par\vspace{0.35em}
3. The next speaker's name - you must generate a message as if this person is speaking.\par\vspace{0.35em}

Your task is to generate a single message from the specified speaker's perspective:\par\vspace{0.35em}
- The message should naturally continue the discussion based on the existing turns and the given topic.\par\vspace{0.35em}
- Write as if you are that specific speaker: use first-person perspective (I, my, etc.) or respond naturally as that person would in a group chat.\par\vspace{0.35em}
- The message should be substantial and relevant to the given topic.\par\vspace{0.35em}
- If the theme involves a particular person's preferences, you can shape the conversation around the things they like.\par\vspace{0.35em}

\textbf{\#\#\# Constraints}\par\vspace{0.35em}
1. The generated message should be casual, realistic, and human-like. Do not output emojis.\par\vspace{0.35em}
2. Reply in a relaxed, natural tone—like how a real person would talk.\par\vspace{0.35em}
3. Keep it concise: one or two sentences is appropriate, not too long.\par\vspace{0.35em}
4. Direct descriptions of images and tables are forbidden; you can extend the discussion around their content.\par\vspace{0.35em}
5. Do NOT include the speaker's name in your output—output only the message content, as if the speaker is typing it.\par\vspace{0.35em}

\textbf{\#\#\# Output format}\par\vspace{0.35em}
Output ONLY the message text that the specified speaker would say. No prefix, no "Name:", no quotes. Just the raw message.

\vspace{0.2em}}}
\end{minipage}\par\vspace{0.65em}
\captionof{figure}{Prompt template for scaffold and smoothing.}
\label{fig:prompt-scaffold-smoothing}
\endgroup
\end{figure}

\newpage
\subsection{Non-Text Captioning Prompt}
\label{appendix:prompt_nontext_captioning}
\begin{figure}[h]
\begingroup
\setlength{\fboxsep}{9pt}%
\setlength{\fboxrule}{0.65pt}%
\noindent\begin{minipage}{\linewidth}
\noindent\fcolorbox{gray!65}{gray!65}{\parbox{\dimexpr\linewidth-2\fboxsep-2\fboxrule}{\centering\textbf{\textcolor{white}{Prompt for Non-Text Captioning}}}}\\[-0.5pt]
\noindent\fcolorbox{gray!50}{gray!10}{\parbox{\dimexpr\linewidth-2\fboxsep-2\fboxrule}{%
\footnotesize\raggedright
\textbf{\#\#\# Task}\par\vspace{0.35em}
You are an expert at describing images for retrieval and QA.\par
Provide a single detailed caption in English: cover main subjects, layout, colors, text, charts/tables/axes if any, and fine-grained spatial relations. Be factual and exhaustive without speculation.\par
\vspace{0.2em}}}
\end{minipage}\par\vspace{0.65em}
\captionof{figure}{Prompt template for non-text captioning.}
\label{fig:prompt-nontext-captioning}
\endgroup
\end{figure}

\subsection{LLM-Judge Error Diagnosis}
\label{appendix:prompt_llm_judge_error_diagnosis}
\begin{figure}[h]
\begingroup
\setlength{\fboxsep}{9pt}%
\setlength{\fboxrule}{0.65pt}%
\noindent\begin{minipage}{\linewidth}
\noindent\fcolorbox{gray!65}{gray!65}{\parbox{\dimexpr\linewidth-2\fboxsep-2\fboxrule}{\centering\textbf{\textcolor{white}{Prompt for LLM-Judge Error Diagnosis}}}}\\[-0.5pt]
\noindent\fcolorbox{gray!50}{gray!10}{\parbox{\dimexpr\linewidth-2\fboxsep-2\fboxrule}{%
\footnotesize\raggedright
\textbf{\#\#\# Task}\par\vspace{0.35em}
You are an expert error-analysis judge for a multimodal, multi-source memory benchmark. Given a question, model response, gold answer, available context, and optional conflict-resolution reference, choose the single most likely primary error type.\par\vspace{0.35em}

\textbf{\#\#\# Taxonomy}\par\vspace{0.35em}
1. \textbf{Distributed Evidence Missing}: The context is missing one or more key facts needed for the gold answer.\par
2. \textbf{Updated Evidence Missing}: The gold answer depends on newer, corrected, or authoritative evidence that is absent from the context.\par
3. \textbf{Outdated or Conflicting Evidence Misused}: Both outdated/conflicting and updated/authoritative evidence are present, but the model follows the wrong one.\par
4. \textbf{Preference Inference Error}: The context contains enough cues to infer a preference or convention, but the model fails to infer it correctly.\par
5. \textbf{Function Execution Error}: The context is sufficient, but the model fails at the action layer, such as wrong function, wrong parameters, or malformed output.\par
6. \textbf{Cross-Source Composition Error}: The clues are present, but the model fails to combine evidence across sources, modalities, or artifacts.\par\vspace{0.35em}

\textbf{\#\#\# Decision Principles}\par\vspace{0.35em}
- Prefer missing-evidence labels when key support is absent.\par
- Prefer \textbf{Updated Evidence Missing} when the missing fact is specifically the latest or authoritative one.\par
- Prefer \textbf{Outdated or Conflicting Evidence Misused} when both sides are present and the model follows the wrong one.\par
- Prefer \textbf{Preference Inference Error} for implicit preference questions.\par
- Prefer \textbf{Function Execution Error} for function-call or action-prediction tasks when evidence is sufficient.\par
- Prefer \textbf{Cross-Source Composition Error} when integration is the main failure.\par\vspace{0.35em}

\textbf{\#\#\# Output Format}\par\vspace{0.2em}
\texttt{\{}\par
\texttt{\ \ "label": "<one of the six labels exactly as written>",}\par
\texttt{\ \ "reason": "<brief explanation, 1-3 sentences>",}\par
\texttt{\ \ "confidence": <a float between 0 and 1>}\par
\texttt{\}}\par
\vspace{0.2em}}}
\end{minipage}\par\vspace{0.65em}
\captionof{figure}{Prompt template for LLM-judge error diagnosis.}
\label{fig:prompt-llm-judge-error-diagnosis}
\endgroup
\end{figure}

\end{document}